\documentclass[11pt,a4paper]{article}
\usepackage{times,latexsym}
\usepackage{url}
\usepackage[T1]{fontenc}

\usepackage[acceptedWithA]{tacl2021v1}
\usepackage{xspace,mfirstuc,tabulary}

\newif\iftaclinstructions
\taclinstructionsfalse % AUTHORS: do NOT set this to true
\iftaclinstructions
\renewcommand{\confidential}{}
\renewcommand{\anonsubtext}{(No author info supplied here, for consistency with
TACL-submission anonymization requirements)}
\newcommand{\instr}
\fi

\iftaclpubformat % this "if" is set by the choice of options

\else

\fi

\usepackage{amsmath}
\usepackage{times}
\usepackage{latexsym}
\usepackage[T1]{fontenc}
\usepackage[utf8]{inputenc}
\usepackage{booktabs}
\usepackage{microtype}
\usepackage{inconsolata}
\usepackage{float}
\usepackage{afterpage}
\usepackage{graphicx}
\usepackage{enumitem}
\usepackage[capitalize]{cleveref}
\usepackage{tikz}
\usepackage{multirow}
\usepackage{twemojis}
\usepackage{rotating}
\usepackage{tabularray}
\usepackage{wasysym}
\usepackage{multicol}
\usepackage{caption}
\usepackage{hyperref}
\usepackage{subfigure}
\usepackage[most]{tcolorbox}
\usepackage{inconsolata} 
\usepackage{pgfplots,pgfplotstable}
\tikzset{
  font={\fontsize{8pt}{10}\selectfont}}
\usepackage{adjustbox}
\tcbset{
  colframe=gray!30,
  colback=gray!4,
  boxrule=0.4pt,
  arc=2pt,
  outer arc=1pt,
  left=4pt,
  right=4pt,
  top=4pt,
  bottom=4pt,
  fonttitle=\bfseries,
  coltitle=black,
  enhanced jigsaw,
  breakable,
}

\newtcolorbox{promptbox}[2][]{
  title={#2},
  coltitle=black,
  fontupper=\ttfamily\small,
  #1
}

\usepackage[table,x11names]{xcolor}

\setlist{left=0mm,noitemsep}

\newcommand{\tikzlabel}[2]{%
    \tikz[baseline=(T.base)]\node(T)[
        fill=#1,
        rounded corners=2pt,
        inner sep=2pt,
        text=black, % You can change text color here
        font=\bfseries,
    ]{#2};%
}

\definecolor{GenericColor}{HTML}{B2A4FF}
\definecolor{GenderColor}{HTML}{FEE440}
\definecolor{AccentColor}{HTML}{71C792}
\definecolor{CSColor}{HTML}{EBAF66}
\definecolor{DisfluentColor}{HTML}{70d6ff}
\definecolor{NEColor}{HTML}{FFC6FF}
\definecolor{NoiseColor}{HTML}{A0C4FF}
\definecolor{EmotionColor}{HTML}{EB8B8B}
\definecolor{LongColor}{HTML}{bee1e6}
\definecolor{LangColor}{HTML}{edede9}
\definecolor{LLMcolor}{HTML}{EE9B51}
\definecolor{SpeechLLMcolor}{HTML}{7AB368}
\definecolor{SFMcolor}{HTML}{417595}

\definecolor{deltagcolor}{HTML}{FEE440}

\definecolor{darkyellow}{RGB}{251,188,4}
\definecolor{darkgreen}{RGB}{52,168,83}
\definecolor{lightblue}{RGB}{66,133,244}
\definecolor{acqua}{RGB}{70,189,198}

\newcommand{\GenericCat}{\tikzlabel{GenericColor}{\texttt{\textcolor{black}{GENERIC}}}}

\newcommand{\GenderCat}{\tikzlabel{GenderColor}{\texttt{\textcolor{black}{GENDER BIAS}}}}
\newcommand{\AccentCat}{\tikzlabel{AccentColor}{\texttt{\textcolor{black}{ACCENTS}}}}
\newcommand{\CSCat}{\tikzlabel{CSColor}{\texttt{\textcolor{black}{CODE SWITCHING}}}}
\newcommand{\DisfluentCat}{\tikzlabel{DisfluentColor}{\texttt{\textcolor{black}{DISFLUENCIES}}}}
\newcommand{\NECat}{\tikzlabel{NEColor}{\texttt{\textcolor{black}{NAMED ENTITIES}}}}
\newcommand{\NoiseCat}{\tikzlabel{NoiseColor}{\texttt{\textcolor{black}{NOISE}}}}
\newcommand{\EmotionCat}{\tikzlabel{EmotionColor}{\texttt{\textcolor{black}{EMOTION}}}}
\newcommand{\LongCat}{\tikzlabel{LongColor}{\texttt{\textcolor{black}{LONG-FORM}}}}
\newcommand{\enlang}{\tikzlabel{LangColor}{\texttt{\textcolor{black}{en}}}}
\newcommand{\delang}{\tikzlabel{LangColor}{\texttt{\textcolor{black}{de}}}}
\newcommand{\eslang}{\tikzlabel{LangColor}{\texttt{\textcolor{black}{es}}}}
\newcommand{\frlang}{\tikzlabel{LangColor}{\texttt{\textcolor{black}{fr}}}}
\newcommand{\itlang}{\tikzlabel{LangColor}{\texttt{\textcolor{black}{it}}}}
\newcommand{\ptlang}{\tikzlabel{LangColor}{\texttt{\textcolor{black}{pt}}}}

\newcommand{\zhlang}{\tikzlabel{LangColor}{\texttt{\textcolor{black}{zh}}}}

\definecolor{NewColor}{HTML}{aaf683}
\newcommand{\newtag}{\tikzlabel{NewColor}{\textsc{\textcolor{black}{NEW!}}}}

\newcommand{\corepi}{\,\raisebox{0.6ex}{\tikz\draw[fill=LLMcolor,draw=LLMcolor] (0,0) circle (0.55ex);}}
\newcommand{\coreauth}{\,\raisebox{0.6ex}{\tikz\draw[fill=SpeechLLMcolor,draw=SpeechLLMcolor] (0,0) circle (0.55ex);}}

\newcommand{\coloredsquare}[1]{\textcolor{#1}{$\blacksquare$}}

\newcommand{\crossmark}{\scalebox{1.25}{\twemoji{cross mark}}}

\newcommand{\diagonalsquare}[2]{%

  \tikz[baseline=0ex, x=1.5ex, y=1.5ex]{
    \useasboundingbox (1.48,0) rectangle (-1.6,1);
    \fill[#1] (0.4,0) -- (1.4,0) -- (1.4,1) -- cycle;
    \fill[#2] (0.4,0) -- (1.4,1) -- (0.4,1) -- cycle;
  }%
}

\newcommand{\xcomet}{\textbf{x{\small \textsc{COMET}}}}
\newcommand{\metricx}{\textbf{{\small \textsc{MetricX}}}}
\newcommand{\cometstrict}{\textbf{x{\small \textsc{COMET$^\text{QE}_\text{\textit{S}}$}}}}
\newcommand{\metricxstrict}{\textbf{{\small \textsc{MetricX$^\text{QE}_\text{\textit{S}}$}}}}

\newcommand{\linguapy}{\textsc{LinguaPY}}
\newcommand{\gendergap}{$\Delta_\text{\female\male}$}
\newcommand{\genderfone}{$\Delta \text{\textbf{F1}}_\text{\female\male}$}

\newcommand{\genderstereotypical}{$\Delta \text{\textbf{S}}_\text{\female\male}$}

\newcommand{\difluentgap}{$\Delta_\text{disfluency}$}
\newcommand{\neacc}{\textbf{\%}$_\text{\textbf{NE}}$}
\newcommand{\termacc}{\textbf{\%}$_\text{\textbf{term}}$}
\newcommand{\noisegap}{$\Delta_\text{noise}$}
\newcommand{\lengthgap}{$\Delta_\text{length}$}
\newcommand{\accentgap}{$\Delta_\text{accent}$}

\newcommand{\whisper}{Whisper}
\newcommand{\seamless}{Seamless}
\newcommand{\owsm}{OWSM}
\newcommand{\canary}{Canary}

\newcommand{\nonefixed}{\hbox to 8mm{}}
\newcommand{\whisperfixed}{\hbox to 8mm{\whisper}}
\newcommand{\seamlessfixed}{\hbox to 8mm{\seamless}}
\newcommand{\owsmfixed}{\hbox to 8mm{\owsm}}
\newcommand{\canaryfixed}{\hbox to 8mm{\canary}}

\newcommand{\aya}{Aya}
\newcommand{\gemma}{Gemma3}
\newcommand{\tower}{Tower+}
\newcommand{\desta}{DeSTA2}
\newcommand{\qwenaudio}{Qwen2-Audio}
\newcommand{\qwenomni}{Qwen3-Omni}
\newcommand{\phimultimodal}{Phi-4-Multimodal}
\newcommand{\voxtral}{Voxtral}
\newcommand{\spire}{Spire}
\newcommand{\qwenthreeomni}{Qwen3-Omni}

\makeatletter
\newcommand\blfootnote[1]{%
  \begingroup
    \renewcommand\thefootnote{}%
    \let\orig@makefntext\@makefntext
    \def\@makefntext##1{\noindent##1}%
    \footnotetext{#1}%
    \addtocounter{footnote}{0}%
    \let\@makefntext\orig@makefntext
  \endgroup
}
\makeatother

\colorlet{sfmcolor}{SFMcolor!20} 
\colorlet{cascadecolor}{LLMcolor!20}
\colorlet{speechllmcolor}{SpeechLLMcolor!20}

\colorlet{sfmcolor_v2}{SFMcolor!80} 
\colorlet{cascadecolor_v2}{LLMcolor!80}
\colorlet{speechllmcolor_v2}{SpeechLLMcolor!80}

\newcommand{\sfmbox}[1]{%
  \colorbox{sfmcolor}{\makebox[2.1cm][l]{#1}}%
}
\newcommand{\cascadebox}[1]{%
  \colorbox{cascadecolor}{\makebox[2.1cm][l]{#1}}%
}
\newcommand{\speechllmbox}[1]{%
  \colorbox{speechllmcolor}{\makebox[2.1cm][l]{#1}}%
}

\newcommand{\sfmboxl}[1]{%
  \colorbox{sfmcolor}{\makebox[3.5cm][l]{#1}}%
}
\newcommand{\cascadeboxl}[1]{%
  \colorbox{cascadecolor}{\makebox[3.5cm][l]{#1}}%
}
\newcommand{\speechllmboxl}[1]{%
  \colorbox{speechllmcolor}{\makebox[3.5cm][l]{#1}}%
}

\definecolor{ClosedLLMcolor}{HTML}{E57373}
\colorlet{closedllmcolor}{ClosedLLMcolor!20}

\newcommand{\closedllmbox}[1]{%
  \colorbox{closedllmcolor}{\makebox[2.1cm][l]{#1}}%
}

\newcommand{\cascadeboxss}[1]{%
  \colorbox{cascadecolor}{\makebox[1.7cm][l]{#1}}%
}
\newcommand{\speechllmboxss}[1]{%
  \colorbox{speechllmcolor}{\makebox[1.7cm][l]{#1}}%
}

\usepackage{amssymb}
\usepackage{arydshln}
\makeatletter
\def\adl@drawiv#1#2#3{%
        \hskip.5\tabcolsep
        \xleaders#3{#2.5\@tempdimb #1{1}#2.5\@tempdimb}%
                #2\z@ plus1fil minus1fil\relax
        \hskip.5\tabcolsep}
\newcommand{\cdashlinelr}[1]{%
  \noalign{\vskip 1.3pt
           \global\let\@dashdrawstore\adl@draw
           \global\let\adl@draw\adl@drawiv}
  \cdashline{#1}[.4pt/2pt]
  \noalign{\global\let\adl@draw\@dashdrawstore
           \vskip 3pt}}
\makeatother

\title{\makebox[0pt]{\color{white} \null\hspace{5cm}Hearing to Translate:}\includegraphics[width=190px]{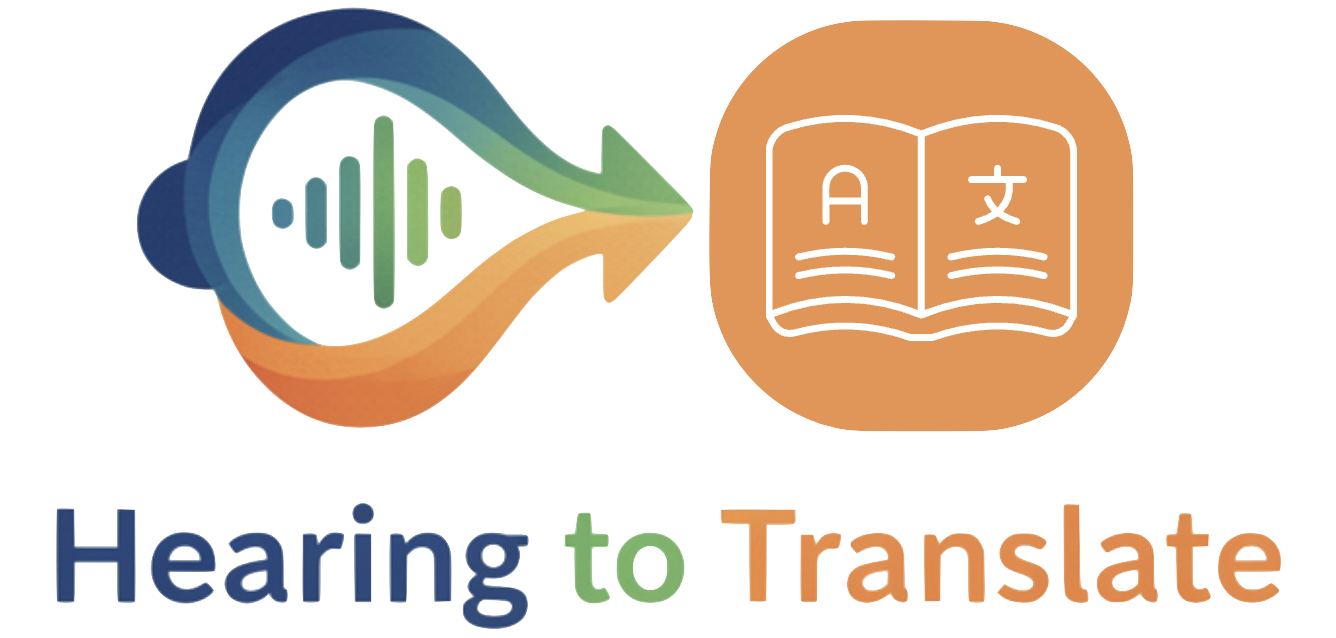}\\The Effectiveness of Speech Modality Integration into LLMs}

\author{
 \textbf{Sara Papi\corepi{}\textsuperscript{1}},
 \textbf{Javier Garcia Gilabert\coreauth{}\textsuperscript{2}},
 \textbf{Zachary Hopton\coreauth{}\textsuperscript{3}},
\textbf{Vilém Zouhar\coreauth{}\textsuperscript{4}},
\\
\textbf{Carlos Escolano\textsuperscript{5}},
\textbf{Gerard I. Gállego\textsuperscript{2,5}},
\textbf{Jorge Iranzo-Sánchez\textsuperscript{6}},
\textbf{Ahrii Kim\textsuperscript{7}},
\\
\textbf{Dominik Mach{\'a}{\v c}ek\textsuperscript{8}},
 \textbf{Patricia Schmidtova\textsuperscript{8}},
 \textbf{Maike Züfle\textsuperscript{9}}
\\
 \textsuperscript{1}Fondazione Bruno Kessler,
 \textsuperscript{2}Barcelona Supercomputing Center,
 \textsuperscript{3}University of Zurich,
 \\
 \textsuperscript{4}ETH Zurich,
 \textsuperscript{5}Universitat Politècnica de Catalunya,
 \textsuperscript{6}Universitat Politècnica de València, 
 \\
 \textsuperscript{7}Soongsil University,
 \textsuperscript{8}Charles University,
  \textsuperscript{9}KIT
\\
 \small{
   \textbf{Correspondence:} \href{mailto:spapi@fbk.eu}{\texttt{spapi@fbk.eu}}
 }
}

\date{}

\begin{document}
\maketitle

\blfootnote{\corepi{} core contributor and PI of the project, \coreauth{}  core contributors, in order of contribution; other authors are ordered alphabetically.}

\begin{abstract}
As Large Language Models (LLMs) expand beyond text, integrating speech as a native modality has given rise to SpeechLLMs, 
% which aim to translate spoken language directly, thereby bypassing traditional transcription-based pipelines. 
which directly process spoken language and enable speech-to-text translation (ST) and other downstream tasks, bypassing traditional transcription-based pipelines.
Whether this integration improves 
% speech-to-text translation 
ST quality over established cascaded architectures, however, remains an open question.
We present Hearing to Translate,%%%%% PREPRINT
\footnote{The \textit{Hearing-to-Translate Suite} is released at \url{https://github.com/sarapapi/hearing2translate}.}
%%%%% SUBMISSION
% \footnote{The test suite
% % , including data, models' outputs, and codebase, 
% will be released upon paper acceptance.}
% % under Apache 2.0 (code) and CC-BY 4.0 (outputs, data) license.} 
the first comprehensive test suite rigorously benchmarking 6 state-of-the-art SpeechLLMs against 16 strong direct and cascade systems that couple leading speech foundation models (SFM), with multilingual LLMs. Our analysis spans 16 benchmarks, 13 language pairs, and 9 challenging conditions, including disfluent, noisy, and long-form speech. Across this extensive evaluation, we find that cascaded systems remain the most reliable solution overall, but most recent SpeechLLMs can match or even outperform cascades in various settings while SFMs lag behind both, highlighting that integrating an LLM, either within the model or in a pipeline, is essential for high-quality speech translation.
\end{abstract}

\section{Introduction}

Large Language Models (LLMs) have transformed natural language processing, enabling unprecedented generalization and reasoning capabilities across a wide range of text-based tasks \citep{achiam2023gpt,touvron2023llama}. Recently, these models have been extended beyond text to encompass multimodal inputs, including vision and audio. Among these modalities, speech holds a particularly central role, as it is the most natural and information-rich form of human communication, conveying not only linguistic content but also prosodic, emotional, and paralinguistic cues \citep{10.1145/3129340}. Integrating this modality into LLMs promises a new generation of language technologies that can process and understand spoken language in a more human-like and contextually grounded manner \citep{latif2023sparks}.

This motivated the emergence of \textbf{SpeechLLMs}: models that extend text-based LLMs with the ability to process spoken language directly. A SpeechLLM typically integrates an audio encoder, often derived from powerful 
% self-supervised 
Speech Foundation Models (SFMs) such as Whisper \citep{whisper-paper} or SeamlessM4T \citep{barrault2023seamlessm4t}, with one or more adapters that bridge the gap between 
% continuous 
acoustic representations and the 
% discrete token space 
embedding space
of an LLM such as Gemma \citep{team2025gemma} or Tower+ \citep{rei2025tower+}. 
%%%%%%
% In the simplest design,  SpeechLLMs employ a single adapter that projects the encoder's speech representations directly into the LLM embedding space \citep{tang2024salmonn,gong2024listen}, but more sophisticated architectures decompose this process into two distinct adapters \citep{zhang2023tuning,10389705}: a length adapter, which compresses the temporal dimension of the speech representation to match the granularity expected by the LLM (i.e., typically reducing thousands of speech frames into a manageable sequence length), and a modality adapter, which performs a semantic projection that aligns the compressed audio features in the LLM's internal representation space. 
% Through this architectural integration, SpeechLLMs can directly process spoken utterances while leveraging the linguistic and world knowledge of large-scale text-pretrained models.
%
This paradigm challenges the traditional architectures that have long dominated speech-to-text translation (ST). Conventional ST systems are typically either \textit{cascade} or \textit{direct} \citep{bentivogli-etal-2021-cascade}. In cascaded setups, a dedicated Automatic Speech Recognition (ASR) model first transcribes the input speech into text, which is then translated by a separate Machine Translation (MT) or, more recently, LLM-based module. This modular design remains highly effective, as it allows each component to be trained on large available corpora and fine-tuned independently for new languages or domains, but it also introduces limitations: translation quality is tightly coupled to ASR accuracy \citep{758176}, potentially leading to error propagation issues \citep{sperber-paulik-2020-speech}, increased latency and computational costs, as two models have to be sequentially executed \citep{papi-etal-2025-real}, and the intermediate transcription step discards prosodic and paralinguistic information that may enrich meaning \citep{tsiamas-etal-2024-speech}. Direct ST models, in contrast, attempt to bypass these issues by mapping speech directly to translated text end-to-end \citep{berard2016listen,weiss17_interspeech}. However, these models are often data-hungry \citep{9054130,jia22b_interspeech,10.24963/ijcai.2023/761}, limited by the scarcity of large-scale parallel speech-translation corpora, and less flexible at test time, lacking the in-context reasoning and adaptability of LLMs.

SpeechLLMs offer a novel alternative to these monolithic ST models. By integrating the speech modality within a general-purpose LLM, they combine 
% the ability of directly processing speech representations as in end-to-end systems with the vast linguistic knowledge and contextual flexibility of LLMs in a unified model, which, in principle, could not only translate spoken language but also adapt their outputs to the user's communicative intent,
% correct errors through reasoning, 
end-to-end speech processing with the linguistic knowledge and contextual flexibility of LLMs, enabling translation, adaptation to user intent, and handling of cross-lingual contexts \citep{rubenstein2023audiopalm}.
These properties make SpeechLLMs an appealing framework for massively multilingual translation systems that can seamlessly operate across text and speech \citep{bapna2022mslam,nguyen-etal-2025-spirit}. However, the practical benefits of this integration remain an open question. It is unclear whether SpeechLLMs can match (or surpass) the performance of translation-specialized direct or cascaded systems that combine powerful SFMs with high-performing LLMs. Furthermore, existing works rarely compare these paradigms systematically \citep{gaido-etal-2024-speech} or consider complex real-world speech phenomena such as disfluencies, background noise, and code-switching.

In this paper, we present \textbf{Hearing to Translate}, the first comprehensive 
% empirical study 
test suite evaluating the effectiveness of speech modality integration into LLMs for translation. We systematically compare 6 state-of-the-art SpeechLLMs against 16 strong systems (4 direct and 12 cascade) built on top of leading SFMs
% (e.g., Whisper, SeamlessM4T) 
and multilingual and translation-oriented LLMs.
% (e.g., Gemma, Tower+).
% , with XXX systems among both direct (e.g., Whisper and SeamlessM4T) and cascaded architectures built upon strong SFMs paired with leading multilingual and translation-oriented LLMs (e.g., Gemma and Tower+). 
% Our evaluation spans 13 language pairs, and 16 benchmarks representing 9 diverse conditions, which reflect specific linguistic or acoustic phenomena, to assess translation quality and robustness under realistic scenarios. 
Our evaluation encompasses 13 language pairs and 16 benchmarks, covering 9 diverse conditions that capture a range of linguistic and acoustic phenomena, 
%thereby 
enabling a comprehensive assessment of translation quality
 and robustness 
in realistic settings.
Through this 
%large-scale 
analysis, we address a fundamental question for the SpeechLLM era: \textit{Does integrating the speech modality directly into LLMs truly enhance 
% spoken language 
speech
translation, or do cascaded architectures or traditional direct models remain the most effective solutions?}

% Through this analysis, we address a core question for the SpeechLLM era, providing crucial insights into how future LLMs should evolve to process spoken language effectively and robustly:

% \epigraph{\itshape Does integrating speech directly into LLMs yield better spoken language translation, or do modular architectures still offer the most effective solution?}

\section{Related Works}

% Cascade vs direct comparison from 2021, IWSLT recent eval campaigns showing that cascade systems surpass again direct models. Also recent work \citep{min2025end}. On the MT side, LLMs recently shown to be on par or superior to standard MT models at WMT..

\paragraph{Cascaded vs. Direct ST: A Historical Comparison.} The comparison between cascaded and direct architectures has long been a central topic in ST research. Though early works highlighted the potential of end-to-end models to reduce error propagation and latency while achieving comparable or superior results to pipeline approaches 
% \citep{pino-etal-2019-harnessing,9054759}, 
\citep{9054759},
recent evidence paints a more nuanced picture. Most recent IWSLT evaluation campaigns \citep{ahmad-etal-2024-findings,agostinelli-etal-2025-findings} consistently report that cascades, especially those combining strong SFMs with high-performing LLMs \citep{koneru-etal-2025-kits,wang-etal-2025-nyas}, again outperform direct approaches across multiple language pairs and acoustic conditions. Similarly, \citet{min2025end} show that despite architectural advances, direct systems still struggle to generalize
% robustly 
in realistic multilingual or low-resource scenarios. While these studies have clarified the strengths and weaknesses of each paradigm, systematic comparisons in the era of LLM-enhanced models remain limited \citep{gaido-etal-2024-speech}. Our work revisits this long-standing debate under a new lens: evaluating how 
% speech modality integration into LLMs 
SpeechLLMs
reshapes the traditional balance between cascaded and direct ST.

% \XXX{Dominik: \citet{zaitova-etal-2025-walk} can be mentioned: idioms are better translated with cascades than with direct models.}

\paragraph{The LLM Era is Here, for MT.} 
% LLMs have recently reshaped the MT landscape, achieving results comparable to or surpassing specialized translation models in recent WMT campaigns \citep{kocmi-etal-2024-findings,kocmi-etal-2025-findings}. 
LLMs have recently transformed the MT landscape, achieving performance comparable to or surpassing specialized translation models in recent WMT campaigns \citep{kocmi-etal-2024-findings,kocmi-etal-2025-findings}. 
% Studies such as \citet{garcia2023unreasonable,stap-etal-2024-fine,deutsch-etal-2025-wmt24} attribute these gains to the broad multilingual coverage, contextual reasoning, and in-context learning capabilities of LLMs, which enable high-quality translation even without task-specific fine-tuning. 
Their broad multilingual coverage, contextual reasoning, and in-context learning enable high-quality translation without task-specific fine-tuning \citep{garcia2023unreasonable,stap-etal-2024-fine,deutsch-etal-2025-wmt24}.
Beyond raw accuracy, LLMs excel in adaptation to user intent \citep{sarti-etal-2023-ramp}, style and formality control \citep{rippeth-etal-2022-controlling}, and explaining and correcting their own translations \citep{treviso-etal-2024-xtower}---dimensions traditionally outside the scope of standard MT models. This paradigm shift has sparked growing interest in extending LLMs beyond text to speech, motivating the development of SpeechLLMs for ST. However, while the superiority of LLMs over traditional MT systems has been established in text translation, this assumption has not yet been verified for SpeechLLMs in ST. Our work directly addresses this gap, providing the first study testing whether the advantages of LLM-based translation extend to the speech modality.

\section{The Hearing-to-Translate Suite}

In this section, we describe the main ingredients of the test suite: the analyzed phenomena (\cref{subsec:cat}), the selected benchmarks (\cref{subsec:benchmarks}), and the metrics used for evaluation (\cref{subsec:metrics}).

\subsection{Categorization of Analyzed Phenomena}
\label{subsec:cat}

To evaluate the robustness and generalization ability of SpeechLLMs across realistic scenarios, we introduce a diverse set of conditions collectively referred to as the Hearing-to-Translate Suite. Each condition targets a specific linguistic, acoustic, or sociolinguistic phenomenon known to challenge speech and translation systems \citep{shah2024speech}. The suite enables a controlled and comprehensive analysis of model behavior across nine categories:
% under a wide range of speech-related factors. The suite includes nine categories:

\begin{itemize}
    \item \GenericCat{} Clean, well-segmented speech from standard benchmarks, used as a reference for model performance under ideal conditions.
    \item \GenderCat{} Utterances balanced across male and female speakers to examine whether translation outputs preserve or distort gendered information and pronoun use.
    \item \AccentCat{} Speech from different geographic varieties of a given language, assessing the ability of models to generalize beyond the accent or dialect distribution seen during training.
    \item \CSCat{} Segments containing intra-sentential language alternation, which require models to dynamically adapt to mixed-language input and maintain coherence in translation.
    \item \DisfluentCat{} Spontaneous speech containing hesitations, repetitions, and self-corrections, used to evaluate how well models handle natural, non-scripted communication.
    \item \NECat{} Speech including person names, locations, and organizations, testing the preservation and accuracy of proper nouns.
    %\item \ToxicCat{} Detection of hallucinated, offensive, or harmful language in the translation that was not present in the source speech. 
    \item \NoiseCat{} Audio with added environmental or background noise, evaluating the robustness of models to unclean acoustic conditions.
    \item \EmotionCat{} Emotionally expressive speech, assessing whether prosodic and affective cues influence translation fidelity and tone.
    \item \LongCat{} Extended audio segments containing multiple sentences, often of several minutes, used to evaluate contextual consistency and memory handling in translation models.
\end{itemize}

\begin{table*}[t]
    \footnotesize
    %\scriptsize
    % \fontsize{10pt}{10pt}\selectfont
    \centering
    \setlength{\tabcolsep}{2pt}
    \begin{tabular}{llccc}
    \toprule
    \bf Benchmark & \bf License & \bf Phenomena & \bf Src Lang  \\
    \midrule
    FLEURS \citep{fleurs2022arxiv} & CC-BY 4.0 & \GenericCat{} \GenderCat{} & \enlang{} \delang{} \eslang{} \frlang{} \itlang{} \ptlang{} \zhlang{} \\
    \cmidrule{1-1}
    CoVoST2 \citep{wang2020covost} & CC-0 & \multirow{3}{*}{\GenericCat{}} & \enlang{} \delang{} \eslang{} \itlang{} \ptlang{} \zhlang{} \\
    EuroParlST \citep{jairsan2020a} & CC-BY-NC 4.0 & & \enlang{} \delang{} \eslang{} \frlang{} \itlang{} \ptlang{} \\
    WMT \citep{kocmi-etal-2024-findings,kocmi-etal-2025-findings}  & CC-BY 3.0 & & \enlang{} \\
    \cmidrule{1-1}
    WinoST \citep{costa-jussa-etal-2022-evaluating} & Custom & \GenderCat{} & \enlang{} \\
    \cmidrule{1-1}
    CommonAccent \citep{Zuluaga-GomezAV23} & CC-0 & \multirow{2}{*}{\AccentCat{}} & \enlang{} \delang{} \eslang{} \itlang{} \\
    ManDi \citep{zhao-chodroff-2022-mandi} & CC-BY-NC 3.0 & & \zhlang{} \\
    \cmidrule{1-1}
    CS-Dialogue \citep{zhou2025csdialogue} & CC-BY-NC-SA 4.0 & \multirow{2}{*}{\CSCat{}} & \zhlang{} \\
    CS-FLEURS \citep{yan25c_interspeech} & CC-BY-NC 4.0 & & \delang{} \eslang{} \frlang{} \zhlang{} \\
    \cmidrule{1-1}
    LibriStutter \citep{7178964} & CC-BY-NC 4.0 & \DisfluentCat{} & \enlang{} \\
    \cmidrule{1-1}
    NEuRoparlST \citep{gaido-etal-2021-moby} & CC-BY-NC 4.0 & \NECat{} & \enlang{} \\
    \cmidrule{1-1}
    NoisyFLEURS \newtag{} & CC-BY-NC 4.0 & \NoiseCat{} & \enlang{} \delang{} \eslang{} \frlang{} \itlang{} \ptlang{} \zhlang{} \\
    \cmidrule{1-1}
    EmotionTalk \citep{sun2025emotiontalk} & CC-BY-NC-SA 4.0 & \multirow{2}{*}{\EmotionCat{}} & \zhlang{} \\
    mExpresso \citep{seamless2023} & CC-BY-NC 4.0 & & \enlang{}  \\
    \cmidrule{1-1}
    ACL 60/60 \citep{salesky-etal-2023-evaluating} & \multirow{2}{*}{CC-BY 4.0} & \multirow{2}{*}{\LongCat{}} & \multirow{2}{*}{\enlang{}}\\
    MCIF \citep{papi2025mcifmultimodalcrosslingualinstructionfollowing} & & & \\
    \bottomrule
    \end{tabular}
    % \caption{Overview of benchmarks, their covered phenomena, and source language (ISO 639 two-letter language code is used).  $\dag$WinoST is available under the MIT license with the limitation that recordings cannot be used for speech synthesis, voice conversion, or other applications where the speaker's voice is imitated or reproduced.}    
    \caption{Benchmarks list with their covered phenomena, and source language (in ISO 639 two-letter language code).}   
    \label{tab:bench-sum}
\end{table*}

\subsection{Benchmarks}
\label{subsec:benchmarks}

To ground the analysis of the phenomena introduced in \cref{subsec:cat}, we select and create a set of 
% speech 
benchmarks that collectively cover the nine categories. 
% For each benchmark, we provide a brief description, including 
% %information about 
% its license, and specify the analyzed phenomenon. A summary is provided in \cref{tab:bench-sum}.
A summary, with license and covered languages,
% , including license and the analyzed phenomena, 
is presented in \cref{tab:bench-sum}. For each of them, we provide a brief description below:

\begin{itemize}
    \item \textbf{FLEURS}: It is an n-way parallel speech-text benchmark covering 102 languages, built on the FLoRes-101 MT dataset \cite{goyal-etal-2022-flores}. It provides roughly 12 hours of speech per language and supports evaluation of ASR, ST, language identification, and retrieval. Data collection enforced a balanced speaker sex ratio where possible, enabling analyses of gender bias \cite{attanasio-etal-2024-twists}.
    \item \textbf{CoVoST2}: It is a ST benchmark created for 15 English-to-many and 21 many-to-English language pairs. The source segments (audio and transcripts) are derived from validated segments in version 4 of Common Voice \citep{ardila-etal-2020-common}, and translated by professionals and verified using embedding-based approaches and length heuristics to ensure quality. 
    \item \textbf{EuroParlST}: It is a many-to-many speech translation dataset covering 9 European languages, built from European Parliament debates held between 2008 and 2012. It provides full speech recordings of parliamentary interventions, along with transcripts, translations, speaker metadata, and gold sentence segmentation. In this work, we leverage the en-de audios for deriving the en-zh, not originally supported by the benchmark. 
    \item \textbf{WMT}: The General MT Shared Task annually tracks progress in MT. Since 2024, it has included a \textit{speech domain} built from publicly available one-minute YouTube videos, with randomly sampled 30-50s segments containing at least 30\% speech. The benchmark is challenging due to 
    % long segments and 
    the presence of
    background noise. It covers 10-15 language pairs per edition, mostly out of English, with human reference translations.
    \item \textbf{WinoST}: WinoST is a dataset designed to evaluate gender bias in ST systems. It is the speech version of the WinoMT dataset \cite{stanovsky-etal-2019-evaluating}, and is used to assess inaccuracies in translations that arise from gender stereotypes, focusing on the gender information present in the sentence content rather than the speaker's voice.
    \item \textbf{CommonAccent}: Designed for accent-robust ASR, CommonAccent includes validated speech segments with accent or dialect annotations from Common Voice v7/v11 \citep{ardila-etal-2020-common}. Languages have 4-16 varieties, and test sets are balanced by capping each variety at 100 segments. 
    \item \textbf{ManDi}: It targets Mandarin dialect variation, with 9.6 hours of speech from 36 speakers across six regional dialects plus Standard Mandarin. Speakers read the same materials in both Standard Mandarin and their native dialect. We use only the poem and short-passage recordings, discarding single word recordings.
    \item \textbf{CS-Dialogue}: It is a 104-hour dataset of spontaneous Mandarin-English dialogues with 200 speakers, covering seven topics. We use only the code-switching portion of the test split, which consists primarily of Mandarin utterances containing embedded English.
    \item \textbf{CS-FLEURS}: Derived from FLEURS, it spans 52 languages and provides real and synthetic code-switched data for ASR and ST. For this work, we evaluate a subset of into-English pairs with read human speech.
    \item \textbf{LibriStutter}: It is derived from LibriSpeech \citep{7178964} by automatically inserting disfluencies such as interjections, sound repetitions, word/phrase repetitions, and prolongations, to evaluate their impact on ST quality. 
    \item \textbf{NEuRoparlST}: It is a derivative of EuroparlST with manually annotated Named Entities (NEs) and domain terminology for both transcripts and translated texts. 
    \item \textbf{NoisyFLEURS}: Derived from FLEURS \citep{fleurs2022arxiv}, it is created for this work to evaluate noise robustness. We add two types of 
    realistic 
    noise---babble (B) and ambient (A) from the MUSAN corpus \citep{musan2015}---following \citet{AnwarSGH0W23} to simulate challenging acoustic conditions.%%% PREPRINT VERSION
    \footnote{NoisyFLEURS is released under the CC-BY-NC 4.0 license at \url{https://huggingface.co/datasets/maikezu/noisy-fleurs}}
    %%% SUBMISSION VERSION
    % We will release it under CC-BY-NC 4.0 license upon paper acceptance.
    \item \textbf{EmotionTalk}: This dataset contains Chinese dyadic conversations recorded with 19 professional actors, annotated for seven emotions (happy, surprise, sad, disgust, anger, fear, neutral), their intensity, and speaking-style captions.
    \item \textbf{mExpresso}: This benchmark is based on an expanded subset of the Expresso dataset~\citep{nguyen23_interspeech}, containing seven read speech with different emotions/styles (default, happy, sad, confused, enunciated, whisper, laughing).
    \item \textbf{ACL 60/60}: Based on ACL 2022 presentations, it captures realistic conditions such as long-form audio and domain-specific terminology. It contains English audio with transcripts and translations into 10 languages. Audio was segmented and transcribed with ASR and manually post-edited, while MT outputs were post-edited to ensure alignment and correct handling of technical terminology.
    \item \textbf{MCIF}: It assesses crosslingual instruction-following in multimodal LLMs, offering 3 fully parallel modalities (text, speech, video) across 4 languages, with both short- and long-form inputs. Specifically for ST, it comprises 2 hours of human-annotated scientific talks from English into three languages (German, Italian, Chinese).
\end{itemize}

\subsection{Metrics}
\label{subsec:metrics}

Most speech benchmarks lack reference translations, and recent work has raised concerns about the reliability of reference-based automatic metrics \citep{freitag-etal-2023-results,zouhar-bojar-2024-quality}. Accordingly, we rely on quality estimation (QE) metrics for evaluation.
To this end, we employ \cometstrict\ and \metricxstrict: modified versions of \xcomet\ \cite{guerreiro-etal-2024-xcomet} and \metricx\ \cite{juraska-etal-2024-metricx} designed to penalize off-target outputs. This strict evaluation follows the recommendation of \citet{zouhar-etal-2024-pitfalls} and applies the maximal penalty to any translation identified by \linguapy\footnote{\url{https://github.com/pemistahl/lingua-py}} as being in the wrong language. Specific 
% evaluation 
settings are reported in Appendix \ref{app:eval-set}.
Besides pure quality-based scores, we also report tailored metrics, which are presented below:

\paragraph{Performance Gap.} For several phenomena, we quantify performance variation through a unified \emph{gap} formulation, which measures the relative difference between two quantities, $Q_{A}$ and $Q_{B}$:
\begin{equation*}
    \Delta = 100 \cdot (Q_{A} - Q_{B}) \; / \; Q_{A}
\end{equation*}
\noindent
where $Q_{A}$ and $Q_{B}$ denote evaluation scores computed on two contrasting subsets of the same benchmark, using either \cometstrict\ or task-specific metrics. A value close to zero indicates comparable performance across conditions; positive values indicate better performance on subset $A$ than on $B$, while negative values indicate better performance on $B$ than on $A$. The gap is computed for the following phenomena:
\begin{itemize}
    \item \textbf{Gender Speaker Gap (\gendergap)}: Following \citet{attanasio-etal-2024-twists}, we instantiate the gap by comparing the translation quality (either \cometstrict\ or \metricxstrict) of male ($A=\male$) and female ($B=\female$) speakers, capturing relative performance disparities across speaker gender. 
    \item \textbf{Gender Coreference Gap (\genderfone)}: For WinoST, we compute the relative difference in coreference resolution accuracy by applying the gap formulation to \textbf{F1} scores obtained on male ($A=\male$) and female ($B=\female$) subsets, using the official evaluation script.\footnote{\url{https://github.com/gabrielStanovsky/mt_gender}} 
    \item \textbf{Accent Gap (\accentgap)}: Accent robustness is evaluated by contrasting translation quality on standard varieties ($A=\text{STD}$) with non-standard or regional varieties ($B=\neg\text{STD}$).\footnote{This metric is applied only to ManDi, as CommonAccent does not define a single standard variety.} 
    \item \textbf{Disfluency Gap (\difluentgap)}: To assess robustness to speech disfluencies, we compare translation quality on fluent ($A=\text{fl}$) and disfluent ($B=\text{disfl}$) speech subsets.
    \item \textbf{Noise Gap (\noisegap)}: Noise robustness is quantified by instantiating the gap between translation quality obtained on clean ($A=\text{clean}$) and noisy ($B=\text{noisy}$) speech conditions.
    \item \textbf{Length Gap (\lengthgap)}: We measure sensitivity to long-form speech by contrasting short-form ($A=\text{short}$) and long-form ($B=\text{long}$) inputs. A large positive \lengthgap\ indicates substantial degradation when processing entire talks rather than sentence-level segments. Since short-form segments are not paired with references, we resegment system outputs and align them to references using SentencePiece \citep{kudo2018sentencepiecesimplelanguageindependent} and \textsc{mwerSegmenter} \citep{matusov-etal-2005-evaluating}, following standard ST evaluation practice \citep{ansari-etal-2020-findings}.
\end{itemize}

\paragraph{Accuracy.}
For named entities and domain-specific terminology, we report \textbf{case-sensitive accuracy} (\neacc, \termacc) using the official NEuroParl-ST evaluation script.\footnote{\url{https://github.com/mgaido91/FBK-fairseq-ST/blob/emnlp2021/scripts/eval/ne_terms_accuracy.py}}
Specifically:
\begin{align*}
\text{\neacc} &= M_{NE}/\lvert NE\rvert \\
\text{\termacc} &= M_{Term}/\lvert Term\rvert
\end{align*}
where $M_{NE}$ and $M_{Term}$ denote exact string matches in system outputs, and $NE$ and $Term$ are the corresponding reference sets.

\section{Experimental Settings}

\subsection{Models}

% We describe the models used in our analysis, including SpeechLLMs, SFMs used for end-to-end translation, and the pipeline composed of SFMs--used for transcription only--and LLMs. Due to budget constraints, models with more than 32B parameters are not considered in our study. Namely, we selected Whisper, SeamlessM4T, Canary, and OWSM as SFMs, Aya Expanse, Gemma, and Tower+ as LLMs, and Phi-4-Multimodal, Qwen2-Audio, DeSTA2, Voxtral, and Spire as SpeechLLMs. The specific model descriptions, weights, parameters, and library versions are reported in Appendix \ref{app:model-v}.

To allow for wider accessibility and easier reproduction of our results, we consider models with less than 32 billion parameters. 
Our analysis focuses on the three paradigms: SFMs \coloredsquare{sfmcolor_v2} (used either as ASR or directly for ST), cascades composed of SFMs, and LLMs \coloredsquare{cascadecolor_v2} and SpeechLLMs \coloredsquare{speechllmcolor_v2}. 
Specifically, we selected: 
% \whisper\ \citep{whisper-paper}, \seamless\ \citep{barrault2023seamlessm4t}, \canary\ \citep{canaryv2}, and \owsm\ \citep{peng25c_interspeech} as SFMs; \aya\ \citep{dang2024ayaexpanse}, \gemma\ \citep{team2025gemma}, and \tower\ \citep{rei2025tower+} as LLMs; and \desta\ \citep{lu2024desta2}, \phimultimodal\ \citep{abouelenin2025phi}, \qwenaudio\ \citep{chu2024qwen2}, \spire\ \citep{ambilduke2025spire}, and \voxtral\ \citep{liu2025voxtral} as SpeechLLMs. 
% A summary of the analyzed models is presented in \cref{tab:model-details}, and detailed descriptions are reported in Appendix \ref{app:model-v}.
Whisper, SeamlessM4T, Canary, and OWSM as SFMs; Aya Expanse, Gemma, and Tower+ as LLMs; and Phi-4-Multimodal, Qwen2-Audio, Qwen3-Omni, DeSTA2, Voxtral, and Spire as SpeechLLMs. Specific model descriptions, and details are reported in Appendix \ref{app:model-v}.

\subsection{Languages and Inference}

% Given the wide language coverage of available LLMs and SFMs, we select the set of languages based on those supported by most of the SpeechLLMs selected for our analysis. The language pairs are English-centric, where we either have \{German, French, Italian, Spanish, Portuguese, Chinese\}$\rightarrow$English, or English$\rightarrow$\{German, Dutch, French, Italian, Spanish, Portuguese, Chinese\}. 
Given the broad language coverage of current LLMs and SFMs, we select languages based on those most commonly supported across the SpeechLLMs analyzed in our study. The evaluation focuses on 
% English-centric pairs, including 
\{de, fr, it, es, pt, zh\}$\rightarrow$en and en$\rightarrow$\{de, nl, fr, it, es, pt, zh\}.
For LLMs, we follow the official translation prompt from the WMT 2025 General MT Shared Task \citep{kocmi-etal-2025-findings}, which we adapt for SpeechLLMs to accommodate spoken inputs (see Appendix \ref{app:prompts}). For SFMs, which do not support prompting, we specify either the target language or both the source and target languages, depending on the specific model. Default decoding parameters are used for all models,\footnote{The only exception is Spire, which produced unusable outputs under default settings and was therefore run with beam search (beam size 5).} reflecting real-world, out-of-the-box performance.
% \footnote{Although such tuning could improve output quality, a systematic exploration of parameter sensitivity is beyond the scope of this work, as it would require extensive model- and language-specific optimization.}
%
All inferences are performed using the Hugging Face Transformers library (see Appendix \ref{app:model-v}), except for OWSM, available only via ESPnet \citep{watanabe2018espnet}, and Canary, available via NVIDIA NeMo \citep{kuchaiev2019nemo}. 
% To promote reproducibility, we will release all system outputs and the complete evaluation codebase upon paper acceptance.

% \subsection{Evaluation Settings}
% \label{subsec:eval-set}
% All evaluations were conducted using Python 3.9.16. For \xcomet, we report scores with \texttt{unbabel/xcomet-xxl}. Scores were computed using the \texttt{comet} library (v2.2.2) with \texttt{fp32} precision. For \metricx, we report scores using {\small \texttt{google/metricx-24-hybrid-xxl-v2p6-bfloat16}}.

\section{Results}

We first present the overall results of the 22 systems analyzed in the paper, highlighting key trends (\cref{subsec:overall}). Then, we delve into two main aspects of ST evaluation, gender bias and accents (\cref{subsec:analysis}), and provide human evaluation results with automatic metrics correlation (\cref{subsec:human-eval}).

\subsection{Overall Results}
\label{subsec:overall}

\begin{table*}[!ht]
\centering
\scriptsize 
\setlength{\tabcolsep}{3.25pt}
\renewcommand{\arraystretch}{1}

\scalebox{1.015}{%
\begin{tabular}{lcccccccccccccccc}
%\toprule
\multirow{3}{*}{} &
&
\multicolumn{6}{c}{\scriptsize \GenericCat{} } &
\multicolumn{3}{c}{\scriptsize \GenderCat{} } &
\multicolumn{3}{c}{\scriptsize \AccentCat{} } &
\multicolumn{2}{c}{\scriptsize \CSCat{} } \\
\cmidrule(lr){2-8}
\cmidrule(lr){9-11}
\cmidrule(lr){12-14}
\cmidrule(lr){15-16}
\cmidrule(lr){15-16}
&
\multicolumn{2}{c}{\tiny FLEURS} &
\multicolumn{2}{c}{\tiny CoVoST2} &
\multicolumn{2}{c}{\tiny EuroParl-ST} &
\multicolumn{1}{c}{\tiny WMT} &
\multicolumn{2}{c}{\tiny FLEURS} &
\multicolumn{1}{c}{\tiny WinoST} &
\multicolumn{2}{c}{\tiny CommonAccent} &
\multicolumn{1}{c}{\tiny ManDi} &
\multicolumn{1}{c}{\tiny CS-Dialogue} &
\multicolumn{1}{c}{\tiny CS-FLEURS} \\
\cmidrule(lr){2-3}
\cmidrule(lr){4-5}
\cmidrule(lr){6-7}
\cmidrule(lr){8-8}
\cmidrule(lr){9-10}
\cmidrule(lr){11-11}
\cmidrule(lr){12-13}
\cmidrule(lr){14-14}
\cmidrule(lr){15-15}
\cmidrule(lr){16-16}
&  \multicolumn{7}{c}{ {{\tiny \textsc{\textbf{xCOMET}$^\text{QE}_\text{\textit{S}}$}}} }  &  
\multicolumn{2}{c}{\tiny \gendergap} 
& {\tiny \genderfone} &  
\multicolumn{2}{c}{ {{\tiny \textsc{\textbf{xCOMET}$^\text{QE}_\text{\textit{S}}$}}} } & \accentgap &  \multicolumn{2}{c}{ {{\tiny \textsc{\textbf{xCOMET}$^\text{QE}_\text{\textit{S}}$}}} }   \\

& en-x & x-en & en-x & x-en & en-x & x-en & en-x & en-x & x-en & en-x & en-x & x-en & zh-en & zh-en & x-en \\
% \midrule
\specialrule{0.5pt}{0pt}{0pt}

\sfmbox{\whisper} & \cellcolor{GenericColor!3} - & \cellcolor{GenericColor!44} 84.8 & \cellcolor{GenericColor!3} - & \cellcolor{GenericColor!37} 73.3 & \cellcolor{GenericColor!3} - & \cellcolor{GenericColor!42} 79.2 & \cellcolor{GenericColor!3} - & \cellcolor{GenderColor!3} - & \cellcolor{GenderColor!49} 0.7 & \cellcolor{GenderColor!3} - & \cellcolor{AccentColor!3} - & \cellcolor{AccentColor!40} 78.2 & \cellcolor{AccentColor!95} 4.6 & \cellcolor{CSColor!50} 69.7 & \cellcolor{CSColor!35} 76.0 \\
\sfmbox{\seamless} & \cellcolor{GenericColor!45} 88.6 & \cellcolor{GenericColor!48} 88.3 & \cellcolor{GenericColor!83} 87.4 & \cellcolor{GenericColor!80} 83.9 & \cellcolor{GenericColor!31} 77.1 & \cellcolor{GenericColor!47} 83.4 & \cellcolor{GenericColor!2} 26.6 & \cellcolor{GenderColor!57} -1.3 & \cellcolor{GenderColor!90} 0.1 & \cellcolor{GenderColor!36} 30.9 & \cellcolor{AccentColor!85} 90.1 & \cellcolor{AccentColor!55} 85.2 & \cellcolor{AccentColor!52} 31.1 & \cellcolor{CSColor!44} 65.0 & \cellcolor{CSColor!50} 85.5 \\
\sfmbox{\canary} & \cellcolor{GenericColor!3} - & \cellcolor{GenericColor!3} - & \cellcolor{GenericColor!3} - & \cellcolor{GenericColor!27} 66.0 & \cellcolor{GenericColor!3} - & \cellcolor{GenericColor!57} 86.4 & \cellcolor{GenericColor!3} - & \cellcolor{GenderColor!3} - & \cellcolor{GenderColor!3} - & \cellcolor{GenderColor!81} 8.6 & \cellcolor{AccentColor!3} - & \cellcolor{AccentColor!49} 84.1 & \cellcolor{AccentColor!3} - & \cellcolor{CSColor!3} - & \cellcolor{CSColor!3} - \\
\sfmbox{\owsm} & \cellcolor{GenericColor!0} 51.7 & \cellcolor{GenericColor!0} 44.4 & \cellcolor{GenericColor!0} 53.1 & \cellcolor{GenericColor!0} 48.2 & \cellcolor{GenericColor!0} 55.1 & \cellcolor{GenericColor!0} 42.6 & \cellcolor{GenericColor!0} 25.3 & \cellcolor{GenderColor!0} 8.5 & \cellcolor{GenderColor!0} 9.6 & \cellcolor{GenderColor!15} 51.6 & \cellcolor{AccentColor!0} 53.5 & \cellcolor{AccentColor!0} 52.7 & \cellcolor{AccentColor!100} 1.8 & \cellcolor{CSColor!0} 30.4 & \cellcolor{CSColor!0} 53.6 \\
\cascadebox{\whisper\ + \aya} & \cellcolor{GenericColor!57} 93.2 & \cellcolor{GenericColor!87} 92.6 & \cellcolor{GenericColor!50} 84.5 & \cellcolor{GenericColor!61} 82.5 & \cellcolor{GenericColor!72} 91.4 & \cellcolor{GenericColor!55} 86.3 & \cellcolor{GenericColor!100} 66.2 & \cellcolor{GenderColor!65} -1.1 & \cellcolor{GenderColor!60} -0.4 & \cellcolor{GenderColor!50} 17.8 & \cellcolor{AccentColor!53} 86.6 & \cellcolor{AccentColor!55} 85.2 & \cellcolor{AccentColor!30} 38.9 & \cellcolor{CSColor!98} 78.8 & \cellcolor{CSColor!78} 90.2 \\
\cascadebox{\hspace{24px} + \gemma} & \cellcolor{GenericColor!50} 92.9 & \cellcolor{GenericColor!74} 91.7 & \cellcolor{GenericColor!49} 83.8 & \cellcolor{GenericColor!50} 81.5 & \cellcolor{GenericColor!54} 90.7 & \cellcolor{GenericColor!49} 85.3 & \cellcolor{GenericColor!95} 64.9 & \cellcolor{GenderColor!46} -2.0 & \cellcolor{GenderColor!70} 0.3 & \cellcolor{GenderColor!41} 26.1 & \cellcolor{AccentColor!49} 85.5 & \cellcolor{AccentColor!49} 84.0 & \cellcolor{AccentColor!23} 41.3 & \cellcolor{CSColor!88} 76.8 & \cellcolor{CSColor!71} 89.0 \\
\cascadebox{\hspace{24px} + \tower} & \cellcolor{GenericColor!57} 93.2 & \cellcolor{GenericColor!90} 92.8 & \cellcolor{GenericColor!50} 84.4 & \cellcolor{GenericColor!58} 82.3 & \cellcolor{GenericColor!72} 91.4 & \cellcolor{GenericColor!50} 86.1 & \cellcolor{GenericColor!91} 63.9 & \cellcolor{GenderColor!50} -1.5 & \cellcolor{GenderColor!49} 0.7 & \cellcolor{GenderColor!97} -3.9 & \cellcolor{AccentColor!50} 86.0 & \cellcolor{AccentColor!50} 84.9 & \cellcolor{AccentColor!19} 42.6 & \cellcolor{CSColor!89} 77.0 & \cellcolor{CSColor!78} 90.2 \\
\cascadebox{\seamless\ + \aya} & \cellcolor{GenericColor!57} 93.2 & \cellcolor{GenericColor!66} 91.1 & \cellcolor{GenericColor!100} 88.9 & \cellcolor{GenericColor!100} 85.4 & \cellcolor{GenericColor!62} 91.0 & \cellcolor{GenericColor!80} 87.4 & \cellcolor{GenericColor!21} 36.6 & \cellcolor{GenderColor!48} -1.7 & \cellcolor{GenderColor!70} -0.3 & \cellcolor{GenderColor!49} 19.0 & \cellcolor{AccentColor!100} 91.7 & \cellcolor{AccentColor!73} 86.3 & \cellcolor{AccentColor!50} 32.2 & \cellcolor{CSColor!80} 75.4 & \cellcolor{CSColor!56} 86.6 \\
\cascadebox{\hspace{26px} + \gemma} & \cellcolor{GenericColor!51} 93.0 & \cellcolor{GenericColor!53} 90.2 & \cellcolor{GenericColor!91} 88.1 & \cellcolor{GenericColor!86} 84.4 & \cellcolor{GenericColor!50} 90.4 & \cellcolor{GenericColor!55} 86.3 & \cellcolor{GenericColor!20} 36.0 & \cellcolor{GenderColor!45} -2.2 & \cellcolor{GenderColor!50} 0.5 & \cellcolor{GenderColor!41} 26.5 & \cellcolor{AccentColor!94} 91.1 & \cellcolor{AccentColor!60} 85.5 & \cellcolor{AccentColor!43} 34.7 & \cellcolor{CSColor!60} 71.5 & \cellcolor{CSColor!48} 84.5 \\
\cascadebox{\hspace{26px} + \tower} & \cellcolor{GenericColor!60} 93.3 & \cellcolor{GenericColor!63} 90.9 & \cellcolor{GenericColor!98} 88.7 & \cellcolor{GenericColor!97} 85.2 & \cellcolor{GenericColor!64} 91.1 & \cellcolor{GenericColor!70} 87.0 & \cellcolor{GenericColor!20} 36.2 & \cellcolor{GenderColor!43} -2.4 & \cellcolor{GenderColor!48} 0.8 & \cellcolor{GenderColor!100} -3.1 & \cellcolor{AccentColor!97} 91.4 & \cellcolor{AccentColor!66} 85.9 & \cellcolor{AccentColor!48} 32.8 & \cellcolor{CSColor!61} 71.7 & \cellcolor{CSColor!52} 85.9 \\
\cascadebox{Canary + \aya} & \cellcolor{GenericColor!69} 93.6 & \cellcolor{GenericColor!3} - & \cellcolor{GenericColor!72} 86.4 & \cellcolor{GenericColor!3} - & \cellcolor{GenericColor!95} 92.3 & \cellcolor{GenericColor!100} 88.3 & \cellcolor{GenericColor!99} 66.1 & \cellcolor{GenderColor!52} -1.4 & \cellcolor{GenderColor!3} - & \cellcolor{GenderColor!50} 17.7 & \cellcolor{AccentColor!73} 88.8 & \cellcolor{AccentColor!74} 86.4 & \cellcolor{AccentColor!3} - & \cellcolor{CSColor!3} - & \cellcolor{CSColor!3} - \\
\cascadebox{\hspace{20px} + \gemma} & \cellcolor{GenericColor!60} 93.3 & \cellcolor{GenericColor!3} - & \cellcolor{GenericColor!60} 85.4 & \cellcolor{GenericColor!3} - & \cellcolor{GenericColor!79} 91.7 & \cellcolor{GenericColor!75} 87.2 & \cellcolor{GenericColor!95} 64.9 & \cellcolor{GenderColor!74} -0.9 & \cellcolor{GenderColor!3} - & \cellcolor{GenderColor!42} 25.7 & \cellcolor{AccentColor!67} 88.1 & \cellcolor{AccentColor!53} 85.1 & \cellcolor{AccentColor!3} - & \cellcolor{CSColor!3} - & \cellcolor{CSColor!3} - \\
\cascadebox{\hspace{20px} + \tower} & \cellcolor{GenericColor!69} 93.6 & \cellcolor{GenericColor!3} - & \cellcolor{GenericColor!68} 86.1 & \cellcolor{GenericColor!3} - & \cellcolor{GenericColor!100} 92.5 & \cellcolor{GenericColor!89} 87.8 & \cellcolor{GenericColor!91} 63.9 & \cellcolor{GenderColor!74} -0.9 & \cellcolor{GenderColor!3} - & \cellcolor{GenderColor!97} -4.0 & \cellcolor{AccentColor!71} 88.6 & \cellcolor{AccentColor!71} 86.2 & \cellcolor{AccentColor!3} - & \cellcolor{CSColor!3} - & \cellcolor{CSColor!3} - \\
\cascadebox{OWSM + \aya} & \cellcolor{GenericColor!49} 91.8 & \cellcolor{GenericColor!50} 90.0 & \cellcolor{GenericColor!50} 84.5 & \cellcolor{GenericColor!54} 82.0 & \cellcolor{GenericColor!50} 90.2 & \cellcolor{GenericColor!48} 84.1 & \cellcolor{GenericColor!55} 53.7 & \cellcolor{GenderColor!45} -2.1 & \cellcolor{GenderColor!49} -0.6 & \cellcolor{GenderColor!49} 18.9 & \cellcolor{AccentColor!49} 85.6 & \cellcolor{AccentColor!48} 83.8 & \cellcolor{AccentColor!3} 48.3 & \cellcolor{CSColor!47} 67.6 & \cellcolor{CSColor!47} 83.6 \\
\cascadebox{\hspace{22px} + \gemma} & \cellcolor{GenericColor!48} 91.7 & \cellcolor{GenericColor!48} 88.5 & \cellcolor{GenericColor!48} 83.5 & \cellcolor{GenericColor!49} 80.7 & \cellcolor{GenericColor!48} 89.4 & \cellcolor{GenericColor!46} 82.6 & \cellcolor{GenericColor!50} 52.4 & \cellcolor{GenderColor!100} 0.3 & \cellcolor{GenderColor!100} 0.0 & \cellcolor{GenderColor!42} 25.2 & \cellcolor{AccentColor!48} 85.1 & \cellcolor{AccentColor!46} 82.4 & \cellcolor{AccentColor!15} 44.0 & \cellcolor{CSColor!42} 63.2 & \cellcolor{CSColor!44} 81.5 \\
\cascadebox{\hspace{22px} + \tower} & \cellcolor{GenericColor!49} 91.9 & \cellcolor{GenericColor!50} 89.9 & \cellcolor{GenericColor!50} 84.2 & \cellcolor{GenericColor!50} 81.5 & \cellcolor{GenericColor!50} 90.3 & \cellcolor{GenericColor!47} 83.6 & \cellcolor{GenericColor!50} 52.3 & \cellcolor{GenderColor!48} -1.7 & \cellcolor{GenderColor!70} 0.3 & \cellcolor{GenderColor!96} -4.2 & \cellcolor{AccentColor!48} 85.3 & \cellcolor{AccentColor!47} 82.7 & \cellcolor{AccentColor!0} 49.2 & \cellcolor{CSColor!43} 64.2 & \cellcolor{CSColor!46} 83.1 \\
\speechllmbox{\desta} & \cellcolor{GenericColor!32} 78.3 & \cellcolor{GenericColor!37} 77.9 & \cellcolor{GenericColor!19} 65.2 & \cellcolor{GenericColor!17} 59.4 & \cellcolor{GenericColor!4} 58.2 & \cellcolor{GenericColor!26} 65.2 & \cellcolor{GenericColor!39} 46.3 & \cellcolor{GenderColor!100} -0.3 & \cellcolor{GenderColor!44} -1.6 & \cellcolor{GenderColor!63} 14.0 & \cellcolor{AccentColor!20} 66.4 & \cellcolor{AccentColor!16} 62.8 & \cellcolor{AccentColor!57} 28.0 & \cellcolor{CSColor!48} 68.4 & \cellcolor{CSColor!32} 74.2 \\
\speechllmbox{\qwenaudio} & \cellcolor{GenericColor!37} 82.2 & \cellcolor{GenericColor!40} 80.6 & \cellcolor{GenericColor!39} 77.9 & \cellcolor{GenericColor!39} 74.1 & \cellcolor{GenericColor!41} 84.1 & \cellcolor{GenericColor!41} 77.9 & \cellcolor{GenericColor!23} 38.0 & \cellcolor{GenderColor!49} -1.6 & \cellcolor{GenderColor!50} 0.5 & \cellcolor{GenderColor!18} 48.1 & \cellcolor{AccentColor!42} 80.9 & \cellcolor{AccentColor!33} 73.9 & \cellcolor{AccentColor!80} 14.2 & \cellcolor{CSColor!50} 69.7 & \cellcolor{CSColor!46} 82.9 \\
\speechllmbox{\phimultimodal} & \cellcolor{GenericColor!23} 71.0 & \cellcolor{GenericColor!48} 88.1 & \cellcolor{GenericColor!13} 61.0 & \cellcolor{GenericColor!27} 66.0 & \cellcolor{GenericColor!19} 68.3 & \cellcolor{GenericColor!40} 77.1 & \cellcolor{GenericColor!27} 39.8 & \cellcolor{GenderColor!44} -2.3 & \cellcolor{GenderColor!48} 0.9 & \cellcolor{GenderColor!0} 65.8 & \cellcolor{AccentColor!33} 75.1 & \cellcolor{AccentColor!43} 80.5 & \cellcolor{AccentColor!64} 23.7 & \cellcolor{CSColor!40} 61.7 & \cellcolor{CSColor!56} 86.5 \\
\speechllmbox{\voxtral} & \cellcolor{GenericColor!100} 94.7 & \cellcolor{GenericColor!76} 91.8 & \cellcolor{GenericColor!56} 85.0 & \cellcolor{GenericColor!53} 81.9 & \cellcolor{GenericColor!72} 91.4 & \cellcolor{GenericColor!55} 86.3 & \cellcolor{GenericColor!96} 65.2 & \cellcolor{GenderColor!70} -1.0 & \cellcolor{GenderColor!70} -0.3 & \cellcolor{GenderColor!81} 8.6 & \cellcolor{AccentColor!64} 87.8 & \cellcolor{AccentColor!61} 85.6 & \cellcolor{AccentColor!74} 17.8 & \cellcolor{CSColor!100} 79.1 & \cellcolor{CSColor!88} 91.9 \\
\speechllmbox{\spire} & \cellcolor{GenericColor!36} 81.4 & \cellcolor{GenericColor!3} - & \cellcolor{GenericColor!22} 66.8 & \cellcolor{GenericColor!3} - & \cellcolor{GenericColor!37} 81.2 & \cellcolor{GenericColor!3} - & \cellcolor{GenericColor!25} 38.7 & \cellcolor{GenderColor!78} -0.8 & \cellcolor{GenderColor!3} - & \cellcolor{GenderColor!61} 14.5 & \cellcolor{AccentColor!31} 73.7 & \cellcolor{AccentColor!3} - & \cellcolor{AccentColor!3} - & \cellcolor{CSColor!3} - & \cellcolor{CSColor!3} - \\
\speechllmbox{\qwenthreeomni} & \cellcolor{GenericColor!91} 94.4 & \cellcolor{GenericColor!100} 93.5 & \cellcolor{GenericColor!86} 87.7 & \cellcolor{GenericColor!97} 85.2 & \cellcolor{GenericColor!82} 91.8 & \cellcolor{GenericColor!80} 87.4 & \cellcolor{GenericColor!100} 66.3 & \cellcolor{GenderColor!91} -0.5 & \cellcolor{GenderColor!80} 0.2 & \cellcolor{GenderColor!69} 12.2 & \cellcolor{AccentColor!70} 88.5 & \cellcolor{AccentColor!100} 88.0 & \cellcolor{AccentColor!70} 20.2 & \cellcolor{CSColor!53} 70.3 & \cellcolor{CSColor!100} 94.0 \\

% \bottomrule
\specialrule{1pt}{0pt}{0pt}
\end{tabular}
}

% ==========================
% SPACER
% ==========================
\vspace{0.1cm} % Adjust this space as needed

% ==========================
% SECOND TABLE (BOTTOM)
% ==========================
\scalebox{1.015}{%
\begin{tabular}{lcccccccccccccccc}

\multirow{3}{*}{} &

\multicolumn{1}{c}{\scriptsize \DisfluentCat{} } &
\multicolumn{2}{c}{\scriptsize \NECat{} } &
\multicolumn{4}{c}{\scriptsize  \NoiseCat{} } &
\multicolumn{2}{c}{\scriptsize \EmotionCat{} } &
\multicolumn{2}{c}{\scriptsize \LongCat{} } \\
\cmidrule(lr){2-2}
\cmidrule(lr){3-4}
\cmidrule(lr){5-8}
\cmidrule(lr){9-10}
\cmidrule(lr){11-12}
&
\multicolumn{1}{c}{\tiny LibriStutter} &
\multicolumn{2}{c}{\tiny NEuRoparl-ST} &
\multicolumn{2}{c}{\tiny NoisyFLEURS\textsubscript{B}	} &
\multicolumn{2}{c}{\tiny NoisyFLEURS\textsubscript{A}} &
\multicolumn{1}{c}{\tiny mExpresso} &
\multicolumn{1}{c}{\tiny EmotionTalk} &
\multicolumn{1}{c}{\tiny ACL6060} &
\multicolumn{1}{c}{\tiny MCIF} \\
\cmidrule(lr){2-2}
\cmidrule(lr){3-4}
\cmidrule(lr){5-6}
\cmidrule(lr){7-8}
\cmidrule(lr){9-9}
\cmidrule(lr){10-10}
\cmidrule(lr){11-11}
\cmidrule(lr){12-12}
& {\tiny \difluentgap} & \neacc & \termacc & \multicolumn{4}{c}{\noisegap} & \multicolumn{2}{c}{ {{\tiny \textsc{\textbf{xCOMET}$^\text{QE}_\text{\textit{S}}$}}} } & \multicolumn{2}{c}{{\tiny \lengthgap} } \\
& en-x & en-x & en-x & en-x & x-en & en-x & x-en & en-x & zh-en & en-x & en-x \\
% \midrule
\specialrule{0.5pt}{0pt}{0pt}

\sfmboxl{\whisper} & \cellcolor{DisfluentColor!3} - & \cellcolor{NEColor!3} - & \cellcolor{NEColor!3} - & \cellcolor{NoiseColor!3} - & \cellcolor{NoiseColor!57} 54.4 & \cellcolor{NoiseColor!3} - & \cellcolor{NoiseColor!49} 13.1 & \cellcolor{EmotionColor!3} - & \cellcolor{EmotionColor!45} 68.3 & \cellcolor{LongColor!3} - & \cellcolor{LongColor!3} - \\
\sfmboxl{\seamless} & \cellcolor{DisfluentColor!0} 44.7 & \cellcolor{NEColor!41} 61.3 & \cellcolor{NEColor!35} 71.2 & \cellcolor{NoiseColor!47} 58.9 & \cellcolor{NoiseColor!50} 57.4 & \cellcolor{NoiseColor!48} 11.9 & \cellcolor{NoiseColor!50} 11.9 & \cellcolor{EmotionColor!45} 79.2 & \cellcolor{EmotionColor!41} 64.3 & \cellcolor{LongColor!3} - & \cellcolor{LongColor!3} - \\
\sfmboxl{\canary} & \cellcolor{DisfluentColor!3} - & \cellcolor{NEColor!56} 66.3 & \cellcolor{NEColor!52} 79.3 & \cellcolor{NoiseColor!3} - & \cellcolor{NoiseColor!3} - & \cellcolor{NoiseColor!3} - & \cellcolor{NoiseColor!3} - & \cellcolor{EmotionColor!3} - & \cellcolor{EmotionColor!3} - & \cellcolor{LongColor!3} - & \cellcolor{LongColor!3} - \\
\sfmboxl{\owsm} & \cellcolor{DisfluentColor!26} 30.4 & \cellcolor{NEColor!0} 43.1 & \cellcolor{NEColor!23} 64.7 & \cellcolor{NoiseColor!29} 66.6 & \cellcolor{NoiseColor!27} 63.9 & \cellcolor{NoiseColor!43} 19.5 & \cellcolor{NoiseColor!45} 19.8 & \cellcolor{EmotionColor!28} 62.6 & \cellcolor{EmotionColor!0} 26.0 & \cellcolor{LongColor!37} 26.9 & \cellcolor{LongColor!45} 11.0 \\
\cascadeboxl{\whisper\ + \aya} & \cellcolor{DisfluentColor!92} 5.9 & \cellcolor{NEColor!50} 65.1 & \cellcolor{NEColor!50} 79.2 & \cellcolor{NoiseColor!65} 51.3 & \cellcolor{NoiseColor!60} 53.1 & \cellcolor{NoiseColor!60} 8.1 & \cellcolor{NoiseColor!50} 11.8 & \cellcolor{EmotionColor!100} 87.4 & \cellcolor{EmotionColor!97} 78.1 & \cellcolor{LongColor!49} 5.3 & \cellcolor{LongColor!49} 4.6 \\
\cascadeboxl{\hspace{24px} + \gemma} & \cellcolor{DisfluentColor!92} 6.0 & \cellcolor{NEColor!48} 64.2 & \cellcolor{NEColor!45} 76.6 & \cellcolor{NoiseColor!66} 50.8 & \cellcolor{NoiseColor!58} 54.1 & \cellcolor{NoiseColor!61} 8.0 & \cellcolor{NoiseColor!54} 11.4 & \cellcolor{EmotionColor!72} 85.9 & \cellcolor{EmotionColor!86} 76.9 & \cellcolor{LongColor!50} 4.4 & \cellcolor{LongColor!50} 3.3 \\
\cascadeboxl{\hspace{24px} + \tower} & \cellcolor{DisfluentColor!89} 6.7 & \cellcolor{NEColor!60} 66.9 & \cellcolor{NEColor!80} 80.4 & \cellcolor{NoiseColor!68} 50.1 & \cellcolor{NoiseColor!66} 51.0 & \cellcolor{NoiseColor!63} 7.8 & \cellcolor{NoiseColor!58} 11.1 & \cellcolor{EmotionColor!83} 86.5 & \cellcolor{EmotionColor!86} 76.9 & \cellcolor{LongColor!50} 4.8 & \cellcolor{LongColor!49} 5.1 \\
\cascadeboxl{\seamless\ + \aya} & \cellcolor{DisfluentColor!59} 14.5 & \cellcolor{NEColor!55} 66.0 & \cellcolor{NEColor!68} 79.9 & \cellcolor{NoiseColor!56} 55.0 & \cellcolor{NoiseColor!46} 58.4 & \cellcolor{NoiseColor!50} 9.2 & \cellcolor{NoiseColor!58} 11.1 & \cellcolor{EmotionColor!49} 83.4 & \cellcolor{EmotionColor!95} 77.9 & \cellcolor{LongColor!3} - & \cellcolor{LongColor!3} - \\
\cascadeboxl{\hspace{26px} + \gemma} & \cellcolor{DisfluentColor!37} 23.8 & \cellcolor{NEColor!50} 65.0 & \cellcolor{NEColor!46} 77.3 & \cellcolor{NoiseColor!56} 55.1 & \cellcolor{NoiseColor!42} 59.7 & \cellcolor{NoiseColor!50} 9.3 & \cellcolor{NoiseColor!52} 11.6 & \cellcolor{EmotionColor!48} 83.0 & \cellcolor{EmotionColor!77} 75.8 & \cellcolor{LongColor!3} - & \cellcolor{LongColor!3} - \\
\cascadeboxl{\hspace{26px} + \tower} & \cellcolor{DisfluentColor!47} 18.7 & \cellcolor{NEColor!63} 67.6 & \cellcolor{NEColor!100} 81.2 & \cellcolor{NoiseColor!55} 55.3 & \cellcolor{NoiseColor!44} 59.2 & \cellcolor{NoiseColor!50} 9.5 & \cellcolor{NoiseColor!56} 11.3 & \cellcolor{EmotionColor!48} 82.4 & \cellcolor{EmotionColor!78} 75.9 & \cellcolor{LongColor!3} - & \cellcolor{LongColor!3} - \\
\cascadeboxl{Canary + \aya} & \cellcolor{DisfluentColor!61} 14.0 & \cellcolor{NEColor!54} 65.8 & \cellcolor{NEColor!75} 80.2 & \cellcolor{NoiseColor!47} 58.8 & \cellcolor{NoiseColor!3} - & \cellcolor{NoiseColor!59} 8.2 & \cellcolor{NoiseColor!3} - & \cellcolor{EmotionColor!96} 87.2 & \cellcolor{EmotionColor!3} - & \cellcolor{LongColor!95} -0.5 & \cellcolor{LongColor!100} -0.2 \\
\cascadeboxl{\hspace{20px} + \gemma} & \cellcolor{DisfluentColor!44} 19.9 & \cellcolor{NEColor!49} 64.6 & \cellcolor{NEColor!47} 77.8 & \cellcolor{NoiseColor!46} 59.3 & \cellcolor{NoiseColor!3} - & \cellcolor{NoiseColor!58} 8.4 & \cellcolor{NoiseColor!3} - & \cellcolor{EmotionColor!66} 85.6 & \cellcolor{EmotionColor!3} - & \cellcolor{LongColor!80} -1.8 & \cellcolor{LongColor!96} 0.4 \\
\cascadeboxl{\hspace{20px} + \tower} & \cellcolor{DisfluentColor!52} 16.3 & \cellcolor{NEColor!64} 67.7 & \cellcolor{NEColor!97} 81.1 & \cellcolor{NoiseColor!45} 59.6 & \cellcolor{NoiseColor!3} - & \cellcolor{NoiseColor!58} 8.4 & \cellcolor{NoiseColor!3} - & \cellcolor{EmotionColor!79} 86.3 & \cellcolor{EmotionColor!3} - & \cellcolor{LongColor!92} -0.8 & \cellcolor{LongColor!92} 0.6 \\
\cascadeboxl{OWSM + \aya} & \cellcolor{DisfluentColor!59} 14.5 & \cellcolor{NEColor!45} 62.8 & \cellcolor{NEColor!57} 79.5 & \cellcolor{NoiseColor!27} 67.5 & \cellcolor{NoiseColor!12} 68.3 & \cellcolor{NoiseColor!47} 14.5 & \cellcolor{NoiseColor!47} 16.9 & \cellcolor{EmotionColor!70} 85.8 & \cellcolor{EmotionColor!57} 73.5 & \cellcolor{LongColor!79} 1.9 & \cellcolor{LongColor!77} -1.4 \\
\cascadeboxl{\hspace{22px} + \gemma} & \cellcolor{DisfluentColor!39} 22.8 & \cellcolor{NEColor!42} 61.7 & \cellcolor{NEColor!46} 77.0 & \cellcolor{NoiseColor!23} 69.3 & \cellcolor{NoiseColor!4} 70.4 & \cellcolor{NoiseColor!46} 14.9 & \cellcolor{NoiseColor!45} 18.5 & \cellcolor{EmotionColor!50} 84.4 & \cellcolor{EmotionColor!49} 71.6 & \cellcolor{LongColor!100} -0.1 & \cellcolor{LongColor!50} -2.8 \\
\cascadeboxl{\hspace{22px} + \tower} & \cellcolor{DisfluentColor!49} 17.2 & \cellcolor{NEColor!50} 65.1 & \cellcolor{NEColor!88} 80.7 & \cellcolor{NoiseColor!10} 75.3 & \cellcolor{NoiseColor!0} 71.6 & \cellcolor{NoiseColor!0} 84.4 & \cellcolor{NoiseColor!0} 84.9 & \cellcolor{EmotionColor!57} 85.1 & \cellcolor{EmotionColor!50} 72.7 & \cellcolor{LongColor!97} -0.4 & \cellcolor{LongColor!71} -1.7 \\
\speechllmboxl{\desta} & \cellcolor{DisfluentColor!74} 10.6 & \cellcolor{NEColor!1} 43.7 & \cellcolor{NEColor!0} 52.1 & \cellcolor{NoiseColor!27} 67.7 & \cellcolor{NoiseColor!1} 71.3 & \cellcolor{NoiseColor!43} 19.9 & \cellcolor{NoiseColor!41} 24.4 & \cellcolor{EmotionColor!34} 68.2 & \cellcolor{EmotionColor!36} 59.7 & \cellcolor{LongColor!0} 93.8 & \cellcolor{LongColor!0} 92.3 \\
\speechllmboxl{\qwenaudio} & \cellcolor{DisfluentColor!41} 21.6 & \cellcolor{NEColor!45} 62.7 & \cellcolor{NEColor!39} 73.3 & \cellcolor{NoiseColor!81} 44.4 & \cellcolor{NoiseColor!50} 57.3 & \cellcolor{NoiseColor!50} 10.0 & \cellcolor{NoiseColor!46} 17.7 & \cellcolor{EmotionColor!39} 73.3 & \cellcolor{EmotionColor!47} 70.0 & \cellcolor{LongColor!0} 94.2 & \cellcolor{LongColor!0} 91.7 \\
\speechllmboxl{\phimultimodal} & \cellcolor{DisfluentColor!33} 26.5 & \cellcolor{NEColor!25} 54.3 & \cellcolor{NEColor!25} 65.5 & \cellcolor{NoiseColor!53} 56.1 & \cellcolor{NoiseColor!100} 36.8 & \cellcolor{NoiseColor!83} 5.6 & \cellcolor{NoiseColor!100} 7.3 & \cellcolor{EmotionColor!0} 34.1 & \cellcolor{EmotionColor!45} 67.7 & \cellcolor{LongColor!49} -6.1 & \cellcolor{LongColor!40} 21.5 \\
\speechllmboxl{\voxtral} & \cellcolor{DisfluentColor!100} 3.9 & \cellcolor{NEColor!60} 66.9 & \cellcolor{NEColor!60} 79.6 & \cellcolor{NoiseColor!96} 38.0 & \cellcolor{NoiseColor!78} 45.7 & \cellcolor{NoiseColor!85} 5.4 & \cellcolor{NoiseColor!94} 7.8 & \cellcolor{EmotionColor!75} 86.1 & \cellcolor{EmotionColor!50} 72.3 & \cellcolor{LongColor!98} 0.3 & \cellcolor{LongColor!94} 0.5 \\
\speechllmboxl{\spire} & \cellcolor{DisfluentColor!38} 23.2 & \cellcolor{NEColor!57} 66.5 & \cellcolor{NEColor!44} 75.9 & \cellcolor{NoiseColor!0} 79.5 & \cellcolor{NoiseColor!3} - & \cellcolor{NoiseColor!24} 47.6 & \cellcolor{NoiseColor!3} - & \cellcolor{EmotionColor!38} 73.1 & \cellcolor{EmotionColor!3} - & \cellcolor{LongColor!3} - & \cellcolor{LongColor!3} - \\
\speechllmboxl{\qwenthreeomni} & \cellcolor{DisfluentColor!78} 9.5 & \cellcolor{NEColor!100} 74.5 & \cellcolor{NEColor!93} 80.9 & \cellcolor{NoiseColor!100} 36.5 & \cellcolor{NoiseColor!75} 47.3 & \cellcolor{NoiseColor!100} 3.7 & \cellcolor{NoiseColor!98} 7.5 & \cellcolor{EmotionColor!89} 86.8 & \cellcolor{EmotionColor!100} 78.5 & \cellcolor{LongColor!50} 5.1 & \cellcolor{LongColor!88} -0.8 \\

% \bottomrule
\specialrule{1pt}{0pt}{0pt}
\end{tabular}%
}

\caption{Overall performance of the 22 evaluated systems. en-x denotes averages across all target languages, except where each benchmark covers a specific subset (e.g., WinoST: de/es/fr/it/pt; NEuRoparl-ST: es/fr/it; ACL 60/60: de/fr/zh/pt; MCIF: de/it/zh). }

%x-en denotes averages across all source languages for each benchmark, as per Table \ref{tab:bench-sum}.
\label{tab:overall-res}
\end{table*}

Aggregated \cometstrict\ are presented in \cref{tab:overall-res}, while aggregated \metricxstrict\ 
% are presented 
in Appendix \ref{app:overall-other-metrics}. 
Across the {\small\GenericCat{}} benchmarks, a consistent picture emerges: cascaded systems remain difficult to outperform but some SpeechLLMs are closing the gap. Cascades 
% typically 
outperform\footnote{According to \citet{kocmi-etal-2024-navigating}, a difference of 2 xCOMET corresponds to 90\% agreement by humans.} most SpeechLLMs and SFMs, with \voxtral\ and \qwenomni\ standing out as the only SpeechLLMs that reliably close---and sometimes overturn---the gap with best-performing cascades. 
% built on Gemma3 or Tower+. 
SFMs generally lag behind, and most SpeechLLMs struggle to match strong SFMs such as Whisper, and Seamless. OWSM performs worst as a standalone SFM, while, in combination with LLMs, it is able to recover most of its gap, indicating a poor language model ability. Overall, the strongest average results come from Canary and Whisper paired with Aya, \qwenomni\ and \voxtral, which are also the largest cascades and SpeechLLMs in our evaluation.

% Looking at {\small\GenderCat{}}, we notice that the gender gap (\gendergap) in FLEURS spans 0.9 to -2.4 across models, with the sole exception of OWSM, being biased towards the masculine form. The gap is slightly more pronounced in all models when translating from English than into English, with no specific paradigm winning over the others, as the lowest (hence, better) gaps are achieved, in order, by OWSM+Gemma3, \voxtral, and \seamless, and Whisper+Aya. In contrast, WinoST exhibits significantly larger F1 gaps. While SpeechLLMs like \qwenaudio\ and \phimultimodal\ show high disparities, bias in cascade systems is contingent on the choice of LLM, indicating that gender bias stems primarily from the text-generation module rather than the speech encoder: pairing ASR modules with \gemma\ results in substantial gaps, whereas using a specialized translation model like \tower\ significantly mitigates this disparity.

In the {\small\GenderCat{}} category, most models exhibit relatively small gender gaps (\gendergap from 0.9 to –2.4 on FLEURS), except OWSM, which is skewed toward male speakers. Gaps tend to be slightly larger when translating from English than into English. No single paradigm dominates: the smallest gaps are reached by OWSM+Gemma3, \qwenomni, \voxtral, \seamless, and Whisper+Aya.
By contrast, WinoST exposes substantially larger F1 gaps. While SpeechLLMs like \qwenaudio\ and \phimultimodal\ show high disparities, bias in cascades is contingent on the choice of LLM, indicating that gender bias stems primarily from the text-generation module rather than the speech module: pairing ASR modules with \gemma\ results in substantial gaps, whereas using a specialized translation model like \tower\ significantly mitigates this disparity.

% Results on {\small\AccentCat{}} show that Seamless, both used as a direct model or employed in a cascaded architecture, is the best performing on CommonAccent, outperforming the other cascades and Voxtral by at least 1.5 \cometstrict\ on en-x. Instead, OWSM and the other SpeechLLMs struggle with handling accented speech. When looking at ManDi, instead, we observe that the performance gap between standard and non-standard Chinese is better handled by SFMs (with Whisper and OWSM showing \accentgap<5) and SpeechLLMs (with Qwen2-Audio and Voxtral scoring 14-18\accentgap), while cascade models show a significant degradation. Overall, as expected, the speech encoder is the factor driving the performance on accented speech, indicating a major robustness of some SFMs over the others.

For {\small\AccentCat{}}, \seamless---used either directly or inside a cascade---achieves the strongest performance on CommonAccent, outperforming both best cascades and SpeechLLMs by at least 1.5 \cometstrict{} on en-x. OWSM and most SpeechLLMs (except \qwenomni\ and \voxtral) struggle to generalize across accents.
%Can't directly compare since for mandi we have a gap measure so i chnaged slightly
The ManDi results reveal that SFMs (Whisper, OWSM) and SpeechLLMs (Qwen2-Audio, Voxtral) are less biased toward standard Chinese, while cascades exhibit substantial bias toward the standard variety. These findings confirm that accent robustness is driven primarily by the speech encoder, with some SFMs displaying superiority depending on the languages.

% Regarding {\small\CSCat{}}, we observe that both cascaded Whisper and, especially, Voxtral achieve the best results compared to the rest of the systems. Despite both the cascaded Whisper and Voxtral leveraging Whisper as speech encoder, Whisper alone lags behind the other two paradigms, indicating that both encoder and decoder matter for code-switching, as proved by the lower results obtained by SFMs compared to their cascaded counterparts. 

In {\small\CSCat{}}, cascaded Whisper, and especially \voxtral, achieve top performance. Despite both the cascaded Whisper and Voxtral leveraging Whisper as speech encoder, Whisper alone lags behind the other two paradigms, indicating that both encoder and decoder matter for code-switching, as proved by the lower results obtained by SFMs compared to their cascaded counterparts. 

% In {\small\DisfluentCat{}}, some SpeechLLMs (Voxtral and DeSTA2) and the Whisper-based cascades are the most robust systems to stuttered speech, while Seamless and OWSM SFMs, and Phi-4-Multimodal SpeechLLM show high sensitivity to disfluencies, with severe degradations ranging 43-75\difluentgap. The improved capability of handling stuttered speech of DeSTA2 and, especially, Voxtral is not only driven by a more robust speech encoder, but also by how the speech modality integration has been performed, as also Qwen2-Audio leverages Whisper in its backbone (as both DeSTA2 and Voxtral), but its gap is more than double that of DeSTA2 and almost five times that of Voxtral. 

For {\small\DisfluentCat{}}, Voxtral, DeSTA2, and Whisper cascades are the most robust to stuttered speech. Seamless, OWSM, and Phi-4-Multimodal show large degradations (43-75\difluentgap). Interestingly, DeSTA2 and Voxtral outperform Qwen2-Audio (having double and five times, respectively, \difluentgap), even though all three rely on the Whisper encoder, suggesting that robustness to disfluencies is not driven solely by the speech encoder, but also by how it is integrated into the LLM.

In {\small\NECat{}}, trends in NE and terminology accuracy largely align: systems that handle NEs well also handle terminology well. 
NE accuracy and overall translation quality are also correlated, but not perfectly aligned. For instance, the highest translation quality on EuroParl-ST en–x is achieved by Canary-based cascades, whereas the best NE accuracy (on the same test set) is observed in systems combining Tower+ (with the strongest results when paired with Seamless), followed by \qwenomni. This suggests that the choice of LLM plays a central role in NE and terminology accuracy, only partially reflected in overall quality scores, highlighting the value of targeted metrics.
%However, accuracy and translation quality do not always correlate. For example, Canary+Gemma3 reaches 91.7 \cometstrict{} on 
% {\small\GenericCat{}} 
%EuroParl-ST but scores only 11.8\neacc\ and 13.1\termacc, underscoring the importance of targeted metrics. 
% Among top systems, cascades based on Tower+ (excluding OWSM) lead together with \qwenomni, 
% % followed by Canary and Voxtral, 
% but no paradigm dominates universally. It is interesting to note that while performance is similar for \termacc between Tower+ cascades and \qwenomni, the latter is especially strong in \neacc, achieving 74.5 \neacc.

% In noisy conditions ({\small\NoiseCat{}}), all systems show a degradation in performance with both noise types, which becomes very severe when babble noise is present in the speech (with a minimum of 38\noisegap). Interestingly, except for Spire, all SpeechLLMs proved to be comparable or even more robust than both SFMs and, especially, cascaded systems. A manual inspection revealed that SFMs that are involved as the ASR component in cascades tend to hallucinate in the presence of noise, and this behavior is not recovered by the subsequent LLMs, which, without access to the original speech input, can only perpetrate or even exacerbate the produced hallucinations. In this category, SpeechLLMs prove to be the overall best choice.

In noisy conditions ({\small\NoiseCat{}}), all models degrade under both noise types, with babble noise causing extreme degradation (minimum 38\noisegap). Interestingly, all SpeechLLMs except Spire show equal or greater robustness than both SFMs and cascades. A manual inspection revealed that SFMs used as ASR components in cascades often hallucinate under noise, and LLMs, lacking access to the original audio, propagate or amplify these errors. In this category, SpeechLLMs are the most reliable choice.

% Regarding {\small\EmotionCat{}}, cascade models prove to be more robust than direct systems (i.e., SFMs and SpeechLLMs), with the only exception of Voxtral, even without directly accessing audio information. Overall, cascaded systems are the best option, especially when equipped with Whisper or Canary encoders. This trend is in contrast with previous findings in ST \citep{tsiamas-etal-2024-speech}, where direct systems were shown to better capture prosodic information than cascade systems. Instead, our results show that current direct systems are not robust to emotions, whereas cascades are more accurate in translating these nuances.
% In {\small\EmotionCat{}}, cascades prove more robust than direct systems (SFMs and SpeechLLMs), except for Voxtral. This holds even though cascades do not directly access audio cues at the LLM stage.
In {\small\EmotionCat{}}, cascades are more robust than both SFMs and SpeechLLMs (with the exception of \qwenomni) despite lacking direct access to audio cues at the LLM stage. 
Contrary to prior work \citep{tsiamas-etal-2024-speech}, which found direct systems better at capturing prosody, our results show that direct models are not better at handling emotional speech, where cascades remain more stable.

In {\small\LongCat{}}, DeSTA2 and Qwen2-Audio show extreme length degradation (\lengthgap{}$\approx$91–94), suggesting poor suitability for document-level ST despite strong sentence-level results.
SFMs\footnote{\seamless{} and \spire\ were excluded as they lack native support for long-form inference.} achieve mid-range scores but still degrade due to their sentence-level optimization. In contrast, cascades with OWSM, Canary, and Voxtral achieve near-zero \lengthgap, indicating far superior long-form robustness. Interestingly, these cascaded systems degrade slightly on short-form inputs (negative \lengthgap), suggesting a higher level of LLM research maturity in handling long context (as also shown by \citealp{pang-etal-2025-salute}) compared to SpeechLLMs and SFMs. Among SpeechLLMs, Voxtral is again particularly notable: although it uses chunking for acoustic encoding (see Appendix \ref{app:model-v}), it re-concatenates all chunk representations before feeding them into the LLM, enabling real long-context ST. This architectural design makes Voxtral the strongest option for real long-form scenarios, which, in contrast to Canary, OWSM, and Whisper, can actually exploit contextual information.

All in all, we summarize the main findings as:
\begin{tcolorbox}[colback=sfmcolor_v2!10,colframe=sfmcolor_v2!60,title=\textbf{Takeaways}]
\fontsize{10}{10}\selectfont
\begin{itemize}[leftmargin=1.2em]
    \item[\coloredsquare{cascadecolor_v2}] \textbf{Cascade systems remain the most reliable overall}, delivering the strongest and most consistent translation quality across languages, benchmarks, and acoustic conditions.
    \vspace{0.5em}
    \item[\coloredsquare{speechllmcolor_v2}] \textbf{SpeechLLMs show growing potential}: the best models match or even surpass cascades in several settings, particularly when speech and language components are tightly integrated.
    \vspace{0.5em}
    \item[\coloredsquare{sfmcolor_v2}] \textbf{Standalone SFMs lag behind} both cascades and SpeechLLMs, indicating that the improved linguistic abilities achieved by current LLMs are crucial for accurate translation.
    \vspace{0.5em}
    \item[\diagonalsquare{cascadecolor_v2}{speechllmcolor_v2}] \textbf{No paradigm dominates universally, and robustness is phenomenon-dependent}: Cascades excel on emotional and long-form speech, SpeechLLMs are more resilient to noise and code switching, accent/dialect performance is primarily encoder-driven across paradigms, and gender bias disparity and named entities accuracy are tied to the LLM decoder.
\end{itemize}
\end{tcolorbox}

% We confirm some of these results based on automatic metrics with a small-scale human evaluation study in Appendix \ref{app:humeval}.

% Voxtral is the only model capable of approaching the best cascaded solutions

% To be added to the discussion: translation quality has been shown to correlate with gender bias \cite{kocmi-etal-2020-gender}

%\begin{table*}[]
%    \centering
%    \begin{tabular}{c|c}
%         &  \\
%         & 
%    \end{tabular}
%    \caption{Results.. \textbf{en-x} means averaged among all the 7 target languages except for WinoST (covering de,es,fr,it,pt), NEuRoparl-ST, ACL 60/60 (covering de,fr,zh,pt), and MCIF (covering de,it,zh). \textbf{x-en} means averaged among all source languages for each benchmark as described in \cref{tab:bench-sum}.}
%    \label{tab:overall-res}
%\end{table*}

\subsection{Analysis}
\label{subsec:analysis}

\paragraph{Gender Bias.}

%\noindent
%\textbf{Gender Coreference Gap.} 

\begin{figure}[!ht]
\centering
  \includegraphics[width=\linewidth]{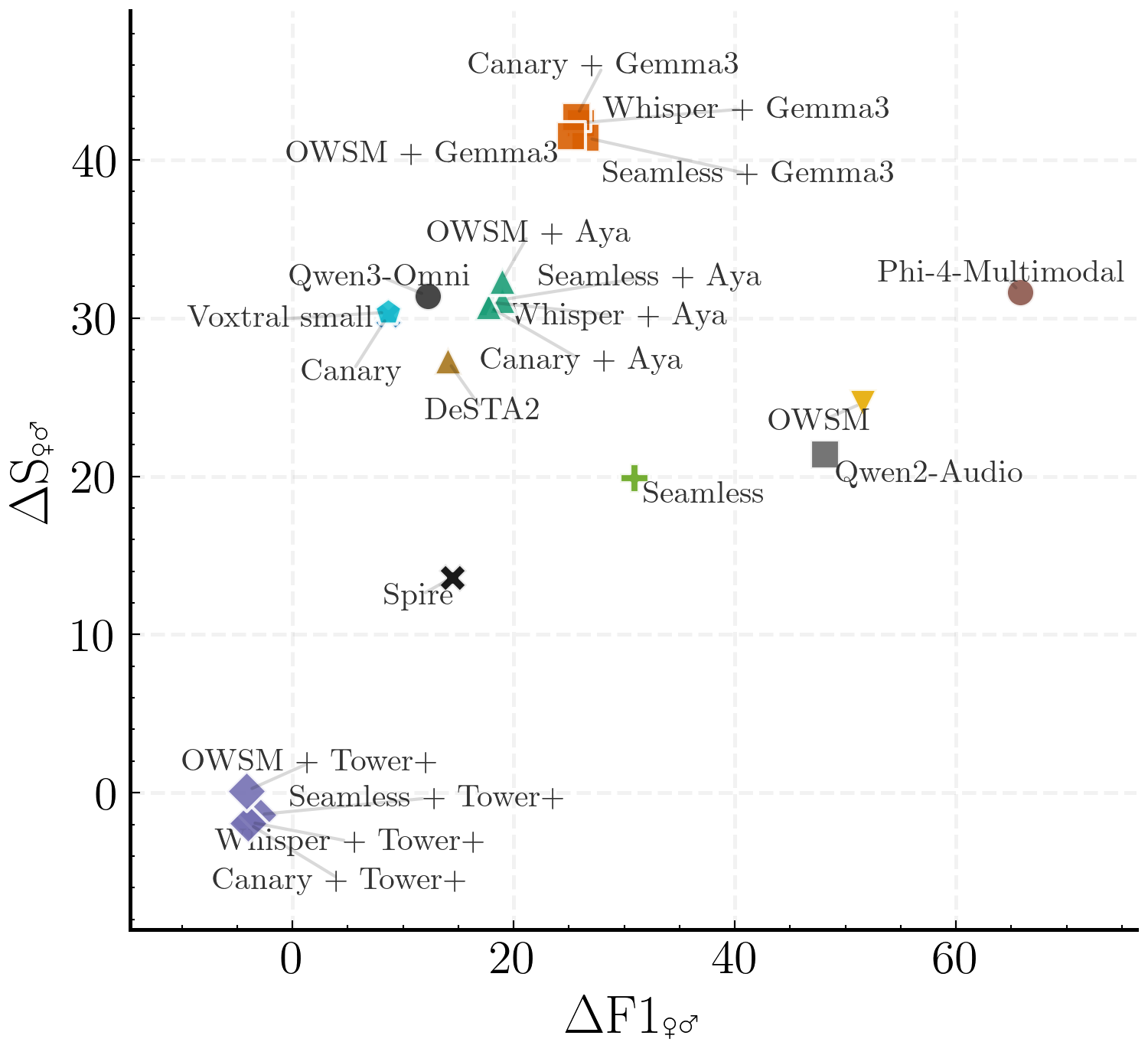}
  \vspace{-1em}
  \caption{Plot showing the relationship between Gender Coreference Gap (\genderfone) and Stereotypical Gap (\genderstereotypical) across all evaluated systems. }\label{fig:scatterplot_gender_bias}
\end{figure}

Beyond gender-term disparity, WinoST enables assessing whether models favour gender-stereotypical translations. A pro-stereotypical set contains occupations aligned with common societal biases (e.g., developer tagged as male, hairdresser as female), while an anti-stereotypical set inverts these assignments (e.g., developer as female, hairdresser as male). 
% We compute the Stereotypical Gap (\genderstereotypical) as follows:
% \begin{equation*}
%     \text{\genderstereotypical} = 100 \cdot (\text{\textbf{\%}}_\text{\text{pro}} - \text{\textbf{\%}}_\text{anti}) \; / \; \text{\textbf{\%}}_\text{pro}
% \end{equation*}
% \noindent
% where $\text{\textbf{\%}}_\text{\text{pro}}$ and $\text{\textbf{\%}}_\text{\text{anti}}$ are the accuracy of the set of sentences with pro-stereotypical entities and the set with anti-stereotypical entities respectively. Ideally, \genderstereotypical\ should be close to 0.
We compute the Stereotypical Gap (\genderstereotypical) as a performance gap (\cref{subsec:metrics}) where $Q_A=\text{\textbf{\%}}_\text{\text{pro}}$ and $Q_B=\text{\textbf{\%}}_\text{\text{anti}}$ are the accuracy of the set of sentences with pro-stereotypical entities and the set with anti-stereotypical entities respectively.
Figure~\ref{fig:scatterplot_gender_bias} shows the relationship between \genderfone\ and \genderstereotypical. Cascades using \tower\ demonstrate the most equitable performance, clustering near 0 with negligible bias across both metrics. In contrast, other systems show higher \genderstereotypical\ scores, indicating significant degradation when translating anti-stereotypical roles. 
% This suggests these models over-rely on training distribution priors rather than resolving gender based on context cues. These findings align with previous works on textual translation \cite{DBLP:journals/patterns/SavoldiBBV25}, as well as LLM generation \cite{DBLP:conf/ci2/KotekDS23}. Furthermore, there is a positive correlation between \genderfone\ and \genderstereotypical\ of 0.54, suggesting that models which struggle with gender co-reference also tend to exhibit a stronger pro-stereotypical bias.
This suggests that models over-rely on training priors rather than contextual cues for gender resolution, consistent with prior findings in MT \cite{DBLP:journals/patterns/SavoldiBBV25} and LLM generation \cite{DBLP:conf/ci2/KotekDS23}. Moreover, \genderfone\ and \genderstereotypical\ are positively correlated ($r=0.54$), indicating that models struggling with gender co-reference also exhibit stronger pro-stereotypical bias.

%\noindent
%\textbf{Gender Speaker Gap.} The Gender Speaker Gap (\gendergap{}) captures whether a system systematically prefers male or female speakers when generating translations. In the en $\to$ X directions, since the source speech is identical for all target languages, this gap would ideally remain stable across languages. However, our results show that for some SpeechLLMs (\desta, \phimultimodal\ and \spire) and \owsm\ SFM the Gender Speaker Gap is highly sensitive to the target language. For instance, \desta\ shows an extreme divergence, ranging from $-7.80$ (en-fr) to $+13.87$ (en-zh). These disparities suggest that gender bias in ST is heavily modulated by the generation process, likely driven by the training data distributions specific to each target language. \TODO{add table/plot in appendices and improve final sentence}

\begin{figure}[!ht]
    \centering
    \includegraphics[width=1.0\linewidth]{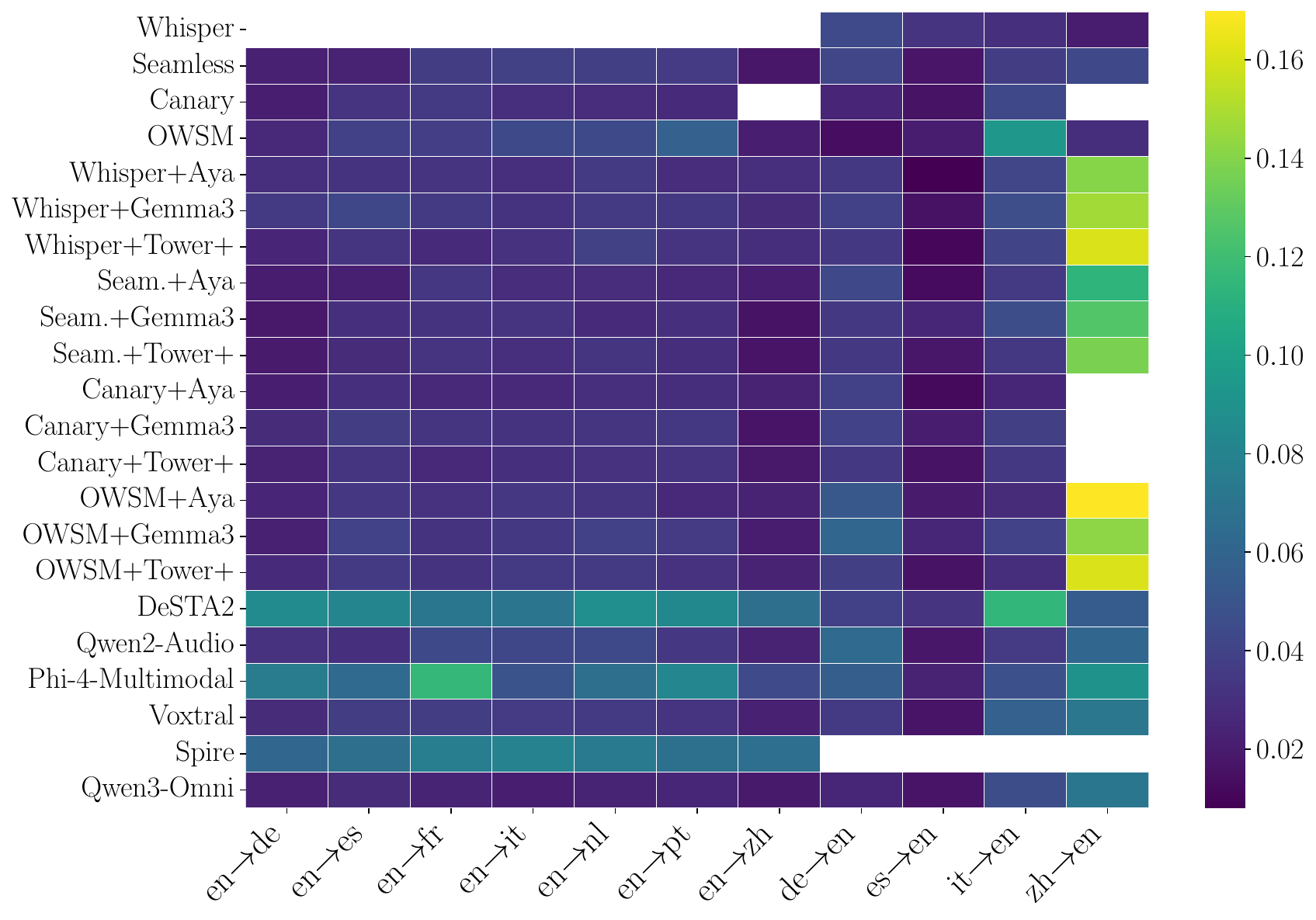}
    \vspace{-1em}
    \caption{Standard deviation of \cometstrict\ scores for ManDi (zh-en) and CommonAccent (all other directions) across source-language accent. Numerical values 
    %for all cells 
    can be found in Table \ref{tab:stdev}.}
    \label{fig:std_viz}
\end{figure}

\paragraph{Accents.}
On the accents benchmarks, x-en generally underperforms en-x, with \phimultimodal\ as a notable exception (\cometstrict\ 80.5 vs. 75.1). In ManDi, zh-en scores are markedly lower than CommonAccent x-en.
While averages provide a coarse view of accent robustness, they obscure patterns in models' weaknesses and strengths with respect to performance on specific accents, which we report in Appendix~\ref{app:accents}. 
To summarize this variability, Figure~\ref{fig:std_viz} presents the standard deviation of \cometstrict\ across source accents, revealing pronounced instability for several SpeechLLMs (\desta{}, \phimultimodal{}, and \spire) on CommonAccent, and for cascaded systems on ManDi, driven by large gaps between standard Mandarin and other dialects.
Across datasets, the most challenging accents include South Asian English, Austrian German, Rioplatense Spanish, and Basilicata-Trentino Italian. In ManDi, standard Mandarin yields the highest scores, while Taiyuan performs worst, likely reflecting both training data biases toward the standard variety and linguistic divergence, such as the Taiyuan tone merger \citep{zhao-chodroff-2022-mandi}. Overall, these results show that strong ST performance on a standard variety does not reliably transfer to other accents/dialects, underscoring the need for more diverse and accent-aware training strategies \citep{lonergran-etal-23-irish,hopton-chodroff-2025-impact,hosseinSameti-accent-invariant}.

\subsection{Human Evaluation}
\label{subsec:human-eval}

% So far, we have relied on automated tools to quantify the quality of translations, either by automated metrics or by detecting the output language.
% However, these tools are not perfect \citep{lavie-etal-2025-findings} and also do not reveal the nature of the errors.

% To evaluate the reliability of automated metrics, we ran a small-scale human evaluation of three models, resulting in the best (on average) for each paradigm in the {\small\GenericCat{}} category: \seamless\ (SFM), \voxtral\ (SpeechLLM), and Canary+Aya (cascade), where humans annotated the translation outputs of the CoVoST2 dataset.

So far, translation quality has been assessed using automatic metrics, which are known to be imperfect and offer limited insight into error types \citep{lavie-etal-2025-findings}. To ensure their reliability, we conducted a small-scale human evaluation on CoVoST2, comparing one of the top-performing models of each paradigm in the {\small\GenericCat{}} category: \seamless\ (SFM), \voxtral\ (SpeechLLM), and Canary+Aya (cascade).
The annotations were done by 
five native speakers of the respective non-English languages.%%%%%% CAMERA READY
\footnote{Human annotations are released under the CC-BY 4.0 license at \url{https://huggingface.co/datasets/zouharvi/hearing2translate-humeval}.}
%%%%% SUBMISSION
% \footnote{We will release the human annotations data under the CC BY 4.0 license upon paper acceptance.}
We used a combination of ESA and MQM protocols \citep{kocmi-etal-2024-error,freitag-etal-2021-experts} with three model outputs next to each other (an extension of side-by-side by \citealp{song2025enhancing}).
We used Pearmut \citep{zouhar2026pearmut} as an annotation interface (see Appendix \ref{app:humeval}) and collect the scores (e.g., 80/100) as well as marked error types (e.g., Accuracy/Omission).
%
% We show the averaged scores in \cref{tab:humeval-scores}, which mimic the automated results, 
% % (e.g., Canary + Aya>Seamless, see \cref{tab:covost}), 
% especially for x-en translation where Canary+Aya outperforms Voxtral and Seamless. 
The human scores are reported in \cref{tab:humeval-scores} and closely mirror the automated results, particularly for x-en translation, where Canary+Aya consistently outperforms Voxtral and Seamless.

\begin{table}[h]
\footnotesize
\setlength{\tabcolsep}{6pt}
\centering
\begin{tabular}{lm{1.75cm}m{1.5cm}m{1.5cm}}
% \toprule
\specialrule{1pt}{0pt}{0pt}
 & \cellcolor{cascadecolor}{Cascade} & \cellcolor{speechllmcolor}{SpeechLLM} & \cellcolor{sfmcolor} SFM \\
% \midrule
\specialrule{0.5pt}{0pt}{0pt}
Average & \cellcolor{GenericColor!54} 81.66 & \cellcolor{GenericColor!51} 80.41 & \cellcolor{GenericColor!46} 78.33 \\[-0.8em]\\en-de & \cellcolor{GenericColor!63} 85.21 & \cellcolor{GenericColor!57} 82.69 & \cellcolor{GenericColor!61} 84.52 \\
en-es & \cellcolor{GenericColor!51} 80.28 & \cellcolor{GenericColor!60} 84.17 & \cellcolor{GenericColor!74} 89.60 \\
en-it & \cellcolor{GenericColor!47} 78.61 & \cellcolor{GenericColor!48} 79.23 & \cellcolor{GenericColor!47} 78.93 \\
en-zh & \cellcolor{GenericColor!40} 76.00 & \cellcolor{GenericColor!32} 72.78 & \cellcolor{GenericColor!0} 54.87 \\
en-nl & \cellcolor{GenericColor!9} 63.47 & \cellcolor{GenericColor!20} 68.03 & \cellcolor{GenericColor!19} 67.78 \\
de-en & \cellcolor{GenericColor!60} 84.03 & \cellcolor{GenericColor!51} 80.60 & \cellcolor{GenericColor!43} 77.08 \\
es-en & \cellcolor{GenericColor!87} 94.79 & \cellcolor{GenericColor!86} 94.44 & \cellcolor{GenericColor!75} 89.92 \\
it-en & \cellcolor{GenericColor!76} 90.43 & \cellcolor{GenericColor!53} 81.02 & \cellcolor{GenericColor!59} 83.41 \\
% \bottomrule
\specialrule{1pt}{0pt}{0pt}
\end{tabular}

\caption{Average scores for human evaluation. Each language pair had 60 items annotated.}
\label{tab:humeval-scores}
\end{table}

% The distribution of error types in \cref{tab:humeval-errors} suggests that there is little to no qualitative difference in the types of errors even if the models are principally different (e.g., cascade vs. SpeechLLM).
\cref{tab:humeval-errors} shows that error type distributions are largely similar across paradigms.
Similar to textual MT, simple mistranslations are the most common errors \citep{freitag-etal-2021-experts}.
Omissions are two times more frequent in the SpeechLLM than they are in cascade and SFM, and SpeechLLM and direct models suffer more from undertranslation than cascades (as previously demonstrated by \citealp{bentivogli-etal-2021-cascade}). As expected, models employing LLMs are more affected by overtranslation \citep{bawden-yvon-2023-investigating}, doubling this error compared to the SFM. Lastly, wrong terminology represents the second most frequent error type, attesting to 11.5-12.5\% of the identified errors, and underscoring the importance of measuring NE and term accuracy, as discussed in \cref{subsec:overall}.

\begin{table}[ht]
\footnotesize
\setlength{\tabcolsep}{5pt}
\centering
\resizebox{0.4825\textwidth}{!}{
\begin{tabular}{@{}l@{\hspace{2mm}}m{1.2cm}m{1.4cm}m{1.2cm}}
% \toprule
\specialrule{1pt}{0pt}{0pt}
 & \cellcolor{cascadecolor} Cascade & \cellcolor{speechllmcolor}{SpeechLLM} & \cellcolor{sfmcolor} \hspace{-1mm}SFM\hspace{-4mm} \\
% \midrule
\specialrule{0.5pt}{0pt}{0pt}
Accuracy/Mistranslation & 71 {\fontsize{7}{6}\selectfont (45.2\%)} & 64 {\fontsize{7}{6}\selectfont (41.3\%)} & 77 {\fontsize{7}{6}\selectfont (46.1\%)} \\
Terminology/Wrong term & 18 {\fontsize{7}{6}\selectfont (11.5\%)} & 21 {\fontsize{7}{6}\selectfont (13.5\%)} & 21 {\fontsize{7}{6}\selectfont (12.6\%)} \\
Accuracy/Overtranslation & 14 {\fontsize{7}{6}\selectfont (8.9\%)} & 12 {\fontsize{7}{6}\selectfont (7.7\%)} & 7 {\fontsize{7}{6}\selectfont (4.2\%)} \\
Style/Unidiomatic style & 9 {\fontsize{7}{6}\selectfont (5.7\%)} & 8 {\fontsize{7}{6}\selectfont (5.2\%)} & 8 {\fontsize{7}{6}\selectfont (4.8\%)} \\
Linguistic/Grammar & 7 {\fontsize{7}{6}\selectfont (4.5\%)} & 5 {\fontsize{7}{6}\selectfont (3.2\%)} & 12 {\fontsize{7}{6}\selectfont (7.2\%)} \\
Accuracy/Omission & 6 {\fontsize{7}{6}\selectfont (3.8\%)} & 10 {\fontsize{7}{6}\selectfont (6.5\%)} & 5 {\fontsize{7}{6}\selectfont (3.0\%)} \\
Accuracy/Undertransl. & 4 {\fontsize{7}{6}\selectfont (2.5\%)} & 9 {\fontsize{7}{6}\selectfont (5.8\%)} & 7 {\fontsize{7}{6}\selectfont (4.2\%)} \\
\bottomrule
\end{tabular}
}
% \caption{Distribution of errors (from \href{https://themqm.org/the-mqm-typology/}{MQM error typology}) per model (summed across all languages).}
\caption{Top seven error types (from \href{https://themqm.org/the-mqm-typology/}{MQM error typology}) per model (summed across all languages).}
\label{tab:humeval-errors}
\end{table}

\begin{table}[!htb]
\footnotesize
\setlength{\tabcolsep}{5pt}
\centering
\begin{tabular}{lp{0.9cm}p{0.9cm}p{0.9cm}p{0.9cm}}
% \toprule
\specialrule{1pt}{0pt}{1pt}
 & \multicolumn{2}{c}{{{\scriptsize \textsc{\textbf{xCOMET}$^\text{QE}_\text{\textit{S}}$}}}} & \multicolumn{2}{c}{{{\scriptsize \textsc{\textbf{MetricX}$^\text{QE}_\text{\textit{S}}$}}}} \\
 & global & item & global & item \\
% \midrule
\specialrule{0.5pt}{0pt}{0pt}
Average & \cellcolor{GenericColor!65} \phantom{-}0.460 & \cellcolor{GenericColor!76} \phantom{-}0.152 & \cellcolor{GenericColor!93} \phantom{-}0.574 & \cellcolor{GenericColor!67} \phantom{-}0.134 \\
\\[-0.8em]
en-de & \cellcolor{GenericColor!35} \phantom{-}0.341 & \cellcolor{GenericColor!49} \phantom{-}0.098 & \cellcolor{GenericColor!53} \phantom{-}0.412 & \cellcolor{GenericColor!27} \phantom{-}0.054 \\
en-es & \cellcolor{GenericColor!100} \phantom{-}0.613 & \cellcolor{GenericColor!100} \phantom{-}0.237 & \cellcolor{GenericColor!100} \phantom{-}0.807 & \cellcolor{GenericColor!91} \phantom{-}0.181 \\
en-it & \cellcolor{GenericColor!63} \phantom{-}0.453 & \cellcolor{GenericColor!0} -0.002 & \cellcolor{GenericColor!89} \phantom{-}0.556 & \cellcolor{GenericColor!52} \phantom{-}0.103 \\
en-zh & \cellcolor{GenericColor!81} \phantom{-}0.523 & \cellcolor{GenericColor!100} \phantom{-}0.250 & \cellcolor{GenericColor!86} \phantom{-}0.546 & \cellcolor{GenericColor!100} \phantom{-}0.239 \\
en-nl & \cellcolor{GenericColor!28} \phantom{-}0.312 & \cellcolor{GenericColor!100} \phantom{-}0.202 & \cellcolor{GenericColor!83} \phantom{-}0.531 & \cellcolor{GenericColor!75} \phantom{-}0.149 \\
de-en & \cellcolor{GenericColor!100} \phantom{-}0.630 & \cellcolor{GenericColor!94} \phantom{-}0.189 & \cellcolor{GenericColor!100} \phantom{-}0.676 & \cellcolor{GenericColor!21} \phantom{-}0.042 \\
es-en & \cellcolor{GenericColor!54} \phantom{-}0.416 & \cellcolor{GenericColor!56} \phantom{-}0.112 & \cellcolor{GenericColor!98} \phantom{-}0.592 & \cellcolor{GenericColor!41} \phantom{-}0.081 \\
it-en & \cellcolor{GenericColor!47} \phantom{-}0.390 & \cellcolor{GenericColor!64} \phantom{-}0.128 & \cellcolor{GenericColor!67} \phantom{-}0.470 & \cellcolor{GenericColor!100} \phantom{-}0.224 \\
% \bottomrule
\specialrule{1pt}{0pt}{0pt}
\end{tabular}

\caption{Correlations (item=group-by-item Spearman, global=micro-Pearson) between human scores and strict versions of automated metrics.
}
\label{tab:humeval-correlations}
\end{table}

\paragraph{Correlation with Automatic Metrics.}
% Finally, we measure the agreement between human scores and automated metrics.
% We focus on two abilities of automated metrics: 
% (1) general scoring, and (2) ranking model outputs of the same source.
% We measure the first by global Pearson and the second by group-by-item Spearman \citep{lavie-etal-2025-findings}.
% The results in \cref{tab:humeval-correlations} indicate that, at the item level, automatic metrics are not good at distinguishing between models, likely also due to the presence of close ties and fine-grained quality differences.
% However, at the global level across most languages, both metrics achieve a micro-Pearson correlation of around 0.5, a level comparable to that reported for reference-based metrics \citep{machacek-etal-2023-mt,han-etal-2024-speechqe}.
% This suggests that automatic QE metrics produce reliable assessments for comparing ST systems in all our settings, supporting their usage in our study.
Finally, we assess the agreement between human judgments and automated metrics along two dimensions: \textit{overall scoring} and \textit{ranking} models' outputs for the same source. We quantify the former using global Pearson correlation and the latter using group-by-item Spearman correlation \citep{lavie-etal-2025-findings}.
As shown in \cref{tab:humeval-correlations}, automated metrics struggle to distinguish models at the item level, likely due to close ties and subtle quality differences. In contrast, at the global level across most languages, both metrics reach a micro-Pearson correlation of around 0.5, comparable to reference-based metrics \citep{machacek-etal-2023-mt,han-etal-2024-speechqe}. This indicates that automatic QE metrics provide sufficiently reliable system-level comparisons for ST, justifying their use throughout this study.

\section{Discussion}

\begin{table*}[!ht]
\centering
\scriptsize 
\setlength{\tabcolsep}{5pt}
\renewcommand{\arraystretch}{1.1}

\begin{tabular}{l ccccccc cc cc}
 & \multicolumn{7}{c}{\GenericCat{}} & \multicolumn{2}{c}{\GenderCat{}} & \multicolumn{2}{c}{\NECat{}} \\
\cmidrule(lr){2-8} \cmidrule(lr){9-10} \cmidrule(lr){11-12}
 & \multicolumn{2}{c}{FLEURS} & \multicolumn{2}{c}{CoVoST2} & \multicolumn{2}{c}{EuroParl-ST} & \multicolumn{1}{c}{WMT} & \multicolumn{2}{c}{FLEURS} & \multicolumn{2}{c}{NEuRoparl-ST} \\
\cmidrule(lr){2-3} \cmidrule(lr){4-5} \cmidrule(lr){6-7} \cmidrule(lr){8-8} \cmidrule(lr){9-10} \cmidrule(lr){11-12}
&  \multicolumn{7}{c}{ {{\tiny \textsc{\textbf{xCOMET}$^\text{QE}_\text{\textit{S}}$}}} }  &  
\multicolumn{2}{c}{\tiny \gendergap} & \neacc & \termacc \\
 & en-x & x-en & en-x & x-en & en-x & x-en & en-x & en-x & x-en & en-x & en-x \\

% \midrule

\specialrule{0.5pt}{0pt}{0pt}

\sfmbox{Best SFM} & \cellcolor{GenericColor!45} 88.6 & \cellcolor{GenericColor!48} 88.3 & \cellcolor{GenericColor!83} 87.4 & \cellcolor{GenericColor!79} 83.9 & \cellcolor{GenericColor!31} 77.1 & \cellcolor{GenericColor!47} 83.4 & \cellcolor{GenericColor!2} 26.6 & \cellcolor{GenderColor!58} -1.3 & \cellcolor{GenderColor!90} 0.1 & \cellcolor{NEColor!41} 61.3 & \cellcolor{NEColor!35} 71.2  \\

\cascadebox{Best Cascade} & \cellcolor{GenericColor!56} 93.2 & \cellcolor{GenericColor!87} 92.6 & \cellcolor{GenericColor!50} 84.5 & \cellcolor{GenericColor!59} 82.5 & \cellcolor{GenericColor!74} 91.4 & \cellcolor{GenericColor!62} 86.3 & \cellcolor{GenericColor!90} 66.2 & \cellcolor{GenderColor!67} -1.1 & \cellcolor{GenderColor!60} -0.4 & \cellcolor{NEColor!50} 65.1 & \cellcolor{NEColor!50} 79.2 \\

\speechllmbox{Best SpeechLLM} & \cellcolor{GenericColor!91} 94.4 & \cellcolor{GenericColor!100} 93.5 & \cellcolor{GenericColor!86} 87.7 & \cellcolor{GenericColor!97} 85.2 & \cellcolor{GenericColor!83} 91.8 & \cellcolor{GenericColor!83} 87.4 & \cellcolor{GenericColor!91} 66.3 & \cellcolor{GenderColor!92} -0.5 & \cellcolor{GenderColor!80} 0.2 & \cellcolor{NEColor!100} 74.5 & \cellcolor{NEColor!87} 80.9   \\

\closedllmbox{ Gemini-2.5-flash } & \cellcolor{GenericColor!88} 94.3 & \cellcolor{GenericColor!97} 93.3 & \cellcolor{GenericColor!50} 84.3 & \cellcolor{GenericColor!59} 82.5 & \cellcolor{GenericColor!49} 90.0 & \cellcolor{GenericColor!49} 84.7 & \cellcolor{GenericColor!91} 66.3 & \cellcolor{GenderColor!48} -1.8 & \cellcolor{GenderColor!49} -0.6 & \cellcolor{NEColor!54} 65.8 & \cellcolor{NEColor!100} 81.5  \\
\specialrule{1pt}{0pt}{0pt}

% \bottomrule
\end{tabular}
\caption{Comparative results across the Hearing-to-Translate suite. Performance of the best-in-class SFM, Cascade, and SpeechLLM architectures of Table \ref{tab:overall-res} compared against Gemini-2.5-flash.}
\label{tab:geminicomparison}
\end{table*}

\paragraph{Comparison with Proprietary Models.} 
%\TODO{--> Include closed-source models, like GPT or Gemini.  If the results are different, then your conclusions could change (for example, perhaps large closed-source models are just as good as, or better than, cascades, while open-weight models still lag behind).}
To assess how open-weight systems compare against proprietary models, we evaluate Gemini-2.5-flash on a subset of benchmarks. As shown in Table~\ref{tab:geminicomparison}, Gemini-2.5-flash is competitive on {\small\GenericCat{}}, often matching the strongest SpeechLLM. On the {\small\GenderCat{}} FLEURS subset, it shows small gender gaps, achieving parity levels comparable to the best open-weight systems. In {\small\NECat{}}, the proprietary model outperforms the best SFM and cascade systems in both NE and terminology accuracy, while it lags behind the best SpeechLLM in terms of NE and performs comparably in terminology. 
Overall, these results indicate that top open-weight models can match--and in some settings surpass--proprietary systems.
%In contrast, we observe a significant performance gap in the {\small\NECat{}} benchmark, where Gemini-2.5-flash reaches only 6.8\% \neacc, contrasting sharply with the 74.5\% achieved by the top open-weight SpeechLLM. This result aligns with our earlier observations in Section X, where translation quality was not correlated with accuracy.

\paragraph{Computational Efficiency and Latency.}

To offer an additional perspective on the comparison between the three paradigms, we benchmark the memory efficiency and latency of the top two performing systems implemented within the same environment (HuggingFace).\footnote{This excludes models such as Canary and OWSM.} Measurements are end-to-end, including feature extraction, and are computed on audio inputs of [10, 30, 60, 300, 600] seconds at 16 kHz, with batch size 1 and greedy decoding capped at 4096 output tokens on an NVIDIA A100 (64GB VRAM). Each experiment is repeated for 20 runs, and the average latency (in seconds) and peak memory usage (in GB) are reported in \Cref{fig:latency_memory}. As expected, the largest (and best performing) models are also the most computationally intensive: the cascade with Aya and the \qwenomni\ SpeechLLM are the least efficient in terms of both inference latency and memory usage, requiring $\times$27-34 the latency and $\times$6-14 the memory of the benchmarked SFMs. \voxtral\ lies in between, being substantially faster than Aya-based cascades and \qwenomni\ (up to $\times$33 lower latency), while still requiring more memory than SFMs and Tower+-based cascades (up to +52GB). Overall, these results highlight a clear trade-off between translation quality and computational efficiency across the evaluated paradigms.

% \TODO{--> Include a comparison of the computational efficiency and latency of the various models.}

% Aya + Whisper
% 10s: 28.8569 | Peak Mem: 62.9543 
% 30s: 166.8528  | Peak Mem: 62.9707
% RUNNING 
% 300s: 984.815 | Peak Mem: 63.0309
% 600s: 1792.4703 | Peak Mem: 63.053

\pgfplotstableread[row sep=\\]{
LENGTH	LATENCY   MEMORY \\
1      1.1363      9.5905 \\  
2      2.6965      10.8844 \\  
3      4.9132      14.7395 \\ 
% 4   OOM \\
% 5   OOM \\
}\seamlessplot

\pgfplotstableread[row sep=\\]{
LENGTH	LATENCY   MEMORY \\
1       0.9317       4.4012 \\
2       1.4513      4.4176 \\   
3       5.257       4.4533 \\
4       27.5938      4.4982 \\ 
5       59.4373       4.5041 \\ 
}\whisperplot

\pgfplotstableread[row sep=\\]{
LENGTH	LATENCY   MEMORY \\
1       5.6315       27.9935 \\  
2       6.7442       29.1646 \\ 
3       8.2231       32.8821 \\
% 4       OOM \\ 
% 5       OOM \\
}\seamlesstowerplot

\pgfplotstableread[row sep=\\]{
LENGTH	LATENCY   MEMORY \\
1       154.9187       59.2400 \\  
2       179.0849       60.1746 \\ 
3       181.3582       63.8923 \\
% 4       OOM \\ 
% 5       OOM \\
}\seamlessayaplot

\pgfplotstableread[row sep=\\]{
LENGTH	LATENCY   MEMORY \\
1       2.7452       22.8847 \\ 
2       4.0097       22.901 \\   
3       14.6557      22.9367 \\ 
4       67.0905      22.9613 \\
5       133.7394     22.9846 \\ 
}\whispertowerplot

\pgfplotstableread[row sep=\\]{
LENGTH	LATENCY   MEMORY \\
1       28.8569       62.9543  \\ 
2       166.8528       62.9707 \\   
3       181.3582     63.8923 \\ 
4       984.815      63.0309 \\
5       1792.4703     63.053 \\ 
}\whisperayaplot

\pgfplotstableread[row sep=\\]{
LENGTH	LATENCY   MEMORY \\
1       2.0492       48.7269 \\ 
2       5.0405       48.7269 \\
3       9.7417       48.8842 \\ 
4       53.7652      50.1272 \\
5       111.687      51.6794 \\ 
}\voxtralplot

\pgfplotstableread[row sep=\\]{
LENGTH	LATENCY   MEMORY \\
1       31.9861   60.4058 \\
2       85.5546   60.4763 \\ 
3       177.6559  60.6186 \\ 
4       798.7779      61.7541 \\ 
5       1611.7927       63.1763 \\ 
}\qwenomniplot

\begin{figure}[!ht]
\centering
\small
\begin{minipage}{0.24\textwidth}
\centering
\begin{tikzpicture}
    \begin{axis}[
            ymajorgrids=true,
            xtick pos=left,
            ytick pos=left,
            %minor y tick num=10,
            minor x tick num=0,
            ymin=0.8,
            ymax=2000,
            xmin=0.85,
            xmax=5.15,
            xtick={1,2,3,4,5},
            xticklabels={10,30,60,300,600},
            ytick={1,10,100,1000,2000},
            yticklabels={1,10,100,1000,2000},
            ylabel=Latency(s),
            xlabel=Input Duration (s), 
            xlabel shift={-4pt},
            ylabel shift={-12pt},
            ymode=log,
            log basis y=10,
            width=4.4cm,
            height=6cm,
            compat=newest,
            every axis plot/.append style={thick},
            legend style={at={(1.05,1.22)},    
                    anchor=north,legend columns=4,
                    font=\tiny, line width=0.1pt,
                    nodes={inner sep=0.5pt},
                    },
        ]
        \legend{Seamless, Seamless+Tower+, Seamless+Aya, Voxtral, Whisper, Whisper+Tower+, Whisper+Aya, Qwen3Omni}     
        \addplot[color=lightblue, mark=*] table[x=LENGTH,y=LATENCY]{\seamlessplot};
        \addplot[color=lightblue, dashed, mark=square*] table[x=LENGTH,y=LATENCY]{\seamlesstowerplot};
        \addplot[color=lightblue, dotted, mark=triangle*] table[x=LENGTH,y=LATENCY]{\seamlessayaplot};
        \addplot[color=magenta, mark=diamond*] table[x=LENGTH,y=LATENCY]{\voxtralplot};
        \addplot[color=darkgreen, mark=*] table[x=LENGTH,y=LATENCY]{\whisperplot};
        \addplot[color=darkgreen, dashed, mark=square*] table[x=LENGTH,y=LATENCY]{\whispertowerplot};
        \addplot[color=darkgreen, dotted, mark=triangle*] table[x=LENGTH,y=LATENCY]{\whisperayaplot};
        \addplot[color=orange, mark=diamond*] table[x=LENGTH,y=LATENCY]{\qwenomniplot};
        % \addplot[color=blue, mark=*] table[x=LAAL,y=BLEU]{\topseven};
    \end{axis}
\end{tikzpicture}
\end{minipage}
\hspace{-0.4em}
\begin{minipage}{0.235\textwidth}
\vspace{2.2em}
\begin{tikzpicture}
    \begin{axis}[
            ymajorgrids=true,
            xtick pos=left,
            ytick pos=left,
            minor y tick num=1,
            minor x tick num=0,
            xmin=0.85,
            xmax=5.15,
            ymax=65,
            xtick={1,2,3,4,5},
            xticklabels={10,30,60,300,600},
            ylabel=Memory Peak (GB), 
            xlabel=Input Duration (s),
            ylabel shift={-5pt},
            xlabel shift={-4pt},
            width=4.4cm,
            height=6cm,
            compat=newest,
            ymin=1,
            ytick={5,15,25,35,45,55,65},
            every axis plot/.append style={thick}
        ]   
        \addplot[color=lightblue, mark=*] table[x=LENGTH,y=MEMORY]{\seamlessplot};
        \addplot[color=lightblue, dashed, mark=square*] table[x=LENGTH,y=MEMORY]{\seamlesstowerplot};
        \addplot[color=lightblue, dotted, mark=diamond*] table[x=LENGTH,y=MEMORY]{\seamlessayaplot};
        \addplot[color=magenta, mark=diamond*] table[x=LENGTH,y=MEMORY]{\voxtralplot};
        \addplot[color=darkgreen, mark=*] table[x=LENGTH,y=MEMORY]{\whisperplot};
        \addplot[color=darkgreen, dashed, mark=square*] table[x=LENGTH,y=MEMORY]{\whispertowerplot};
        \addplot[color=darkgreen, dotted, mark=triangle*] table[x=LENGTH,y=MEMORY]{\whisperayaplot};
        \addplot[color=orange, mark=diamond*] table[x=LENGTH,y=MEMORY]{\qwenomniplot};
        % \addplot[color=darkyellow, mark=*] table[y=LENGTH,x=MEMORY]{\seamlessplot};
        % \addplot[color=darkgreen, mark=*] table[y=LENGTH,x=MEMORY]{\whisperplot};
        % \addplot[color=lightblue, mark=*] table[y=LENGTH,x=MEMORY]{\seamlesstowerplot};
        % \addplot[color=orange, mark=*] table[y=LENGTH,x=MEMORY]{\whispertowerplot};
        % \addplot[color=magenta, mark=*] table[y=LENGTH,x=MEMORY]{\voxtralplot};
        % \addplot[color=acqua, mark=*] table[y=LENGTH,x=MEMORY]{\qwenomniplot};
    \end{axis}
\end{tikzpicture}
\end{minipage}
\caption{Latency (s) and Memory Peak (GB) for the top-2 models for each paradigm. Seamless values not reported are due to out of memory. Note that the Whisper VRAM is almost constant due to  chunking mechanisms described in Appendix \ref{app:model-v}.}
\label{fig:latency_memory}
\end{figure}

\paragraph{Effect of Models' Training Mixture.} 
% The models considered in this study are trained on substantially different data mixtures and training strategies (see Appendix \ref{app:model-v}), which may directly impact their performance on the Hearing-to-Translate suite. On the one hand, speech-native models (including SFMs and several SpeechLLMs) are explicitly trained with speech supervision for ASR and/or speech translation. These models are directly optimized to map acoustic representations to text, often through end-to-end objectives or tightly aligned speech-text pretraining, leading to strong inductive biases toward transcription fidelity and acoustic robustness. On the other hand, text-native LLMs are trained purely on multilingual text data and later applied in a cascaded setting, relying on an external transcription stage. Their performance therefore depends primarily on multilingual text modeling capacity, which is typically stronger than speech models as they are usually trained on huge text corpora in a wide set of languages. Notably, among SpeechLLMs, models that incorporate speech already during pretraining rather than only during adaptation (such as \voxtral\ and \qwenomni) consistently outperform other speech-native systems, suggesting that early multimodal alignment may play a key role. 
The models considered in this study are trained on substantially different data mixtures and training strategies (see Appendix \ref{app:model-v}), which may directly impact their performance on the Hearing-to-Translate suite. On the one hand, speech-native models (including SFMs and several SpeechLLMs) are explicitly trained with speech supervision for ASR and/or ST \citep{barrault2023seamlessm4t,raokoluguri25_interspeech}. These systems are directly optimized to map acoustic representations to text, often through end-to-end objectives or tightly aligned speech-text pretraining, leading to strong inductive biases toward transcription fidelity and acoustic robustness. On the other hand, text-native models (LLMs) are trained purely on multilingual text corpora and later applied in a cascaded setting that relies on an external transcription stage. Their performance therefore depends primarily on multilingual text modeling capacity, which is typically stronger than that of speech models due to training on massive text collections spanning many languages \citep{liu2025datasets}. Notably, among SpeechLLMs, models that incorporate speech already during pretraining rather than only during adaptation (e.g., \voxtral, \qwenomni) tend to outperform other speech-native systems, suggesting that early multimodal alignment may play an important role in effectively integrating the speech modality into LLMs.

\paragraph{Cascade vs. SpeechLLM with the same LLM Backbone.}
Comparing a SpeechLLM with cascaded models that leverage the same LLM backbone better isolates the effect of speech modality integration, partially disentangling it from differences in LLM training, quality, and robustness. \cref{tab:same-llm} reports the performance of \qwenomni\ and cascaded systems based on the same LLM, Qwen3,\footnote{\href{https://huggingface.co/Qwen/Qwen3-30B-A3B-Instruct-2507}{\texttt{Qwen/Qwen3-30B-A3B-Instruct-2507}}} on {\small\GenericCat{}}. 
The results show that \qwenomni\ 
% generally outperforms the corresponding cascades across all datasets and language directions, highlighting the promise of the SpeechLLM paradigm. 
generally outperforms the corresponding cascades, while remaining competitive in the few cases where cascades perform slightly better, highlighting the promise of the SpeechLLM paradigm. 
% This advantage comes 
These gains come at the cost of additional training to integrate speech encoders into the LLM, rather than combining off-the-shelf components, but yields clear performance improvements.
% These gains come at the cost of additional training to integrate speech encoders into the LLM, but yield overall performance improvements.

\begin{table}[!ht]
\centering
\scriptsize 
\setlength{\tabcolsep}{2.75pt}
\renewcommand{\arraystretch}{1.1}

\begin{tabular}{lccccccc}
% & \multicolumn{7}{c}{\GenericCat{}} \\
\cmidrule(lr){2-8}
 & \multicolumn{2}{c}{\scriptsize FLEURS} & \multicolumn{2}{c}{\scriptsize CoVoST2} & \multicolumn{2}{c}{\scriptsize EuroParl-ST} & \multicolumn{1}{c}{\scriptsize WMT} \\
\cmidrule(lr){2-3} \cmidrule(lr){4-5} \cmidrule(lr){6-7} \cmidrule(lr){8-8}
 & {\scriptsize en-x} & {\scriptsize x-en} & {\scriptsize en-x} & {\scriptsize x-en} & {\scriptsize en-x} & {\scriptsize x-en} & {\scriptsize en-x} \\
 
% \midrule
\specialrule{0.5pt}{0pt}{0pt}

\speechllmboxss{\fontsize{7pt}{7pt}\selectfont Qwen3-Omni } & \cellcolor{GenericColor!92} 94.4 & \cellcolor{GenericColor!100} 93.5  & \cellcolor{GenericColor!86} 87.7 & \cellcolor{GenericColor!97} 85.2 & \cellcolor{GenericColor!82} 91.8 & \cellcolor{GenericColor!79} 87.4 & \cellcolor{GenericColor!91} 66.3 \\

\cascadeboxss{\fontsize{7pt}{7pt}\selectfont Whisper{\fontsize{6pt}{5pt}\selectfont\;+ Qwen3} } & \cellcolor{GenericColor!50} 92.7 & \cellcolor{GenericColor!79} 92.1  & \cellcolor{GenericColor!49} 84.1 & \cellcolor{GenericColor!51} 82.0 & \cellcolor{GenericColor!57} 90.8 & \cellcolor{GenericColor!51} 86.2 & \cellcolor{GenericColor!87} 65.0 \\
\cascadeboxss{\fontsize{7pt}{7pt}\selectfont Seamless{\fontsize{6pt}{5pt}\selectfont\;+ Qwen3} } & \cellcolor{GenericColor!50} 92.9 & \cellcolor{GenericColor!62} 90.9  & \cellcolor{GenericColor!93} 88.3 & \cellcolor{GenericColor!91} 84.8  & \cellcolor{GenericColor!50} 90.5 & \cellcolor{GenericColor!72} 87.1 & \cellcolor{GenericColor!19} 35.8 \\
\cascadeboxss{\fontsize{7pt}{7pt}\selectfont Canary{\fontsize{6pt}{5pt}\selectfont\;+ Qwen3} } & \cellcolor{GenericColor!56} 93.1 & - & \cellcolor{GenericColor!65} 85.8 & - & \cellcolor{GenericColor!80} 91.7 & \cellcolor{GenericColor!91} 87.9 & \cellcolor{GenericColor!86} 64.8 \\
\cascadeboxss{\fontsize{7pt}{7pt}\selectfont OWSM{\fontsize{6pt}{5pt}\selectfont\;+ Qwen3} } & \cellcolor{GenericColor!48} 91.3 & \cellcolor{GenericColor!49} 89.4 & \cellcolor{GenericColor!49} 83.9 & \cellcolor{GenericColor!49} 81.3 & \cellcolor{GenericColor!49} 89.6 & \cellcolor{GenericColor!47} 83.6 & \cellcolor{GenericColor!49} 52.8 \\

% \bottomrule
\specialrule{1pt}{0pt}{0pt}
\end{tabular}
\caption{Comparison of SpeechLLM and cascade models with the same LLM backbone (Qwen3).}
\label{tab:same-llm}
\end{table}
\vspace{-1em}

% \TODO{--> Add more details about the training data, supported languages, and tasks supported by the various models, and discuss any implications of these differences for interpreting your results.}

\paragraph{Prompt and Parameter Sensitivity.}
Prompt formulation and decoding parameters can influence the outputs of large language models, although the magnitude of this effect varies across tasks and languages \citep{mondshine-etal-2025-beyond-english}. In translation, prior work suggests that prompt variations generally have a smaller impact compared to tasks such as reasoning or summarization, particularly for high-resource language pairs \citep{kocmi-etal-2025-findings-wmt25,aly2025evaluationlargelanguagemodels,schmidtova-etal-2026-important}. Standardized prompts, such as those used in WMT Shared Tasks \citep{kocmi-etal-2025-findings}, are commonly adopted to improve reproducibility and comparability across MT studies \citep{deutsch-etal-2025-wmt24}. In our experiments, we evaluate all models using the same prompt template and suggested decoding configuration, ensuring that performance differences reflect off-the-shelf models' usage. While reformulations and prompt engineering \citep{NEURIPS2020_1457c0d6} may affect absolute scores \citep{garces-arias-etal-2025-decoding}, prior analyses in related translation settings \citep{papi2025mcifmultimodalcrosslingualinstructionfollowing} suggest that such effects are unlikely to alter comparative conclusions substantially.

% \TODO{--> Discuss to what extent prompt sensitivity could affect your results.  Ideally a small study of prompt sensitivity with quantitative results would be included.  If this is not practical, a discussion of any anecdotal or preliminary results, and any potential implications of prompt senstivity, is sufficient.}

\paragraph{Validity of QE Metrics.}
Our evaluation relies on quality estimation metrics due to the scarcity of high-quality reference translations in speech benchmarks.
We validate this approach by assessing their alignment with reference-based metrics and human judgments.
\cref{tab:humeval-correlations-ref} indicates that QE metrics are extremely strong proxies for reference-based evaluation: on benchmarks with available references, global correlations between QE and standard variants of {\small \textsc{\textbf{xCOMET}}} and {\small \textsc{\textbf{MetricX}}} approach 1.0.
Regarding human alignment, while segment-level differentiation remains challenging, \cref{tab:humeval-correlations} shows system-level correlations reaching 0.57 for {\small \textsc{\textbf{MetricX}}}.
This performance is comparable to reference-based baselines reported in recent metrics campaigns \citep{lavie-etal-2025-findings}.
Despite the limited scale of our human evaluation, the consistency across metric-to-metric and metric-to-human comparisons supports the validity of the architectural rankings presented in this work.

% \TODO{
% - small human eval but still better than nothing as most similar works do \\
% - in Table 6 we show correlation with reference-based metrics, which is approx 1.0 for global correlation and between 0.8 and 0.9 on item level. \\
% - based on this we conclude the validity of QE

% Vilem
% \TODO{--> Your study relies heavily on QE metrics, and the human evaluation included is quite small.  Discuss the implications of this for interpreting your results and for future work.}
% }

\begin{table}[!htb]
\footnotesize
\setlength{\tabcolsep}{5pt}
\centering
\begin{tabular}{lp{0.9cm}p{0.9cm}p{0.9cm}p{0.9cm}}
% \toprule
\specialrule{1pt}{0pt}{0pt}
 & \multicolumn{2}{c}{{\scriptsize \textsc{\textbf{xCOMET}$_\text{\textit{S}}$}}} & \multicolumn{2}{c}{{\scriptsize \textsc{\textbf{MetricX}$_\text{\textit{S}}$}}} \\
 & global & item & global & item \\
% \midrule
\specialrule{0.5pt}{0pt}{0pt}
{ACL6060-long} & \cellcolor{GenericColor!100} \phantom{-}1.000 & \cellcolor{GenericColor!82} \phantom{-}0.912 & \cellcolor{GenericColor!100} \phantom{-}1.000 & \cellcolor{GenericColor!84} \phantom{-}0.918 \\
{ACL6060-short} & \cellcolor{GenericColor!100} \phantom{-}0.999 & \cellcolor{GenericColor!72} \phantom{-}0.861 & \cellcolor{GenericColor!100} \phantom{-}0.999 & \cellcolor{GenericColor!76} \phantom{-}0.878 \\
{CoVoST2} & \cellcolor{GenericColor!100} \phantom{-}0.998 & \cellcolor{GenericColor!61} \phantom{-}0.804 & \cellcolor{GenericColor!100} \phantom{-}0.999 & \cellcolor{GenericColor!56} \phantom{-}0.781 \\
{CS-FLEURS} & \cellcolor{GenericColor!100} \phantom{-}0.998 & \cellcolor{GenericColor!76} \phantom{-}0.879 & \cellcolor{GenericColor!100} \phantom{-}0.998 & \cellcolor{GenericColor!73} \phantom{-}0.867 \\
{EuroParl-ST} & \cellcolor{GenericColor!100} \phantom{-}0.998 & \cellcolor{GenericColor!61} \phantom{-}0.807 & \cellcolor{GenericColor!100} \phantom{-}1.000 & \cellcolor{GenericColor!72} \phantom{-}0.859 \\
{FLEURS} & \cellcolor{GenericColor!100} \phantom{-}0.999 & \cellcolor{GenericColor!66} \phantom{-}0.831 & \cellcolor{GenericColor!100} \phantom{-}0.999 & \cellcolor{GenericColor!76} \phantom{-}0.882 \\
{MCIF-long} & \cellcolor{GenericColor!100} \phantom{-}1.000 & \cellcolor{GenericColor!81} \phantom{-}0.907 & \cellcolor{GenericColor!100} \phantom{-}1.000 & \cellcolor{GenericColor!82} \phantom{-}0.912 \\
{MCIF-short} & \cellcolor{GenericColor!100} \phantom{-}0.999 & \cellcolor{GenericColor!78} \phantom{-}0.891 & \cellcolor{GenericColor!100} \phantom{-}0.999 & \cellcolor{GenericColor!79} \phantom{-}0.893 \\
{mExpresso} & \cellcolor{GenericColor!100} \phantom{-}0.998 & \cellcolor{GenericColor!62} \phantom{-}0.809 & \cellcolor{GenericColor!100} \phantom{-}0.999 & \cellcolor{GenericColor!59} \phantom{-}0.794 \\
% \bottomrule
\specialrule{1pt}{0pt}{0pt}
\end{tabular}
\caption{Correlations (item=group-by-item Spearman, global=micro-Pearson) between strict quality-estimation and reference-based variants of {\small \textsc{\textbf{xCOMET}}} and {\small \textsc{\textbf{MetricX}}}, averaged over reference-based language pairs.}
\label{tab:humeval-correlations-ref}
\end{table}
\vspace{-0.5em}

\section{Conclusions}
We introduced \textit{Hearing to Translate}, a comprehensive test suite for evaluating 22 ST systems across 13 language pairs, 9 phenomena, and 16 benchmarks. Our results show that cascaded architectures remain the most reliable, 
but recent SpeechLLMs, which are rapidly evolving, are able to match or even outperform them in various settings, such as noise, code-switching, and disfluencies.
% consistently outperforming both SpeechLLMs and standalone SFMs. While SpeechLLMs are improving rapidly, they only match or slightly surpass cascades in limited scenarios such as noise, code-switching, and disfluencies. 
Standalone SFMs lag behind, highlighting the crucial role of LLMs (either integrated or as part of a cascade) for high-quality ST. 
% Analyses of gender bias and accent variation reveal that all paradigms struggle with contextual cues for gender assignment and are sensitive to accented speech, with SpeechLLMs being particularly affected. 
Targeted analyses of gender bias and accent variation further reveal that all three paradigms struggle to leverage contextual cues for gender assignment, often defaulting to masculine forms, with bias mostly driven by the LLM component. Models, particularly SpeechLLMs, exhibit high sensitivity to accents, showing substantial performance variations.
Human evaluation highlights recurring ST error patterns, with mistranslations, terminology errors, and overtranslation emerging as the dominant failures---with the latter being especially prevalent in LLM-based systems---and their alignment with automatic metrics validates our evaluation framework. 
% Overall, our study suggests that cascades remain the most dependable approach, while SpeechLLMs hold promise for future development with careful attention to robustness and bias.

%%%%%%
\section*{Acknowledgments}
This work has received funding from the European Union’s Horizon research and innovation programme under grant agreement No 101135798, project Meetween (My Personal AI Mediator for Virtual MEETings BetWEEN People). % << Sara + Maike (1:2)
This research was also supported by the G-LAMP Program of the National Research Foundation of Korea (NRF) grant funded by the Ministry of Education (No. RS-2025-25441317). % << ahrii (1:1)
% Dominik+Patrícia:
The authors acknowledge the support of the National Recovery Plan funded project MPO 60273/24/21300/21000 CEDMO 2.0 NPO.
This paper has received funding from the Project OP JAK Mezisektorová spolupráce Nr.~CZ.02.01.01/00/23\_020/0008518 named ``Jazykověda, umělá inteligence a jazykové a řečové technologie: od výzkumu k aplikacím.''
This research was also co-funded by the European Union (ERC, NG-NLG, 101039303) and by Charles University projects GAUK 252986. %and SVV project 260 821. (4:2)
% << Dominik+Patrícia
This work was also supported by MLLM4TRA (PID2024-158157OB-C32) funded by
MCIN/AEI/10.13039/501100011033/FEDER, UE. This work was also supported by the ELOQUENCE project (Horizon Europe Grant Number 101135916). % << Javi (2:1)
This work is also funded by the Ministerio para la Transformación Digital y de la Función Pública and Plan de Recuperación, Transformación y Resiliencia – Funded by EU – NextGenerationEU within the framework of the project Modelos del Lenguaje. % << Gerard (1:1)
The research leading to these results has also received funding from
EU4Health Programme 2021--2027 as part of Europe's Beating Cancer Plan
under Grant Agreements no. 101129375; and from the
Government of Spain's grant PID2021-122443OB-I00 funded by
MICIU/AEI/\allowbreak10.13039/\allowbreak5011\-00011033 and by
``ERDF/EU'', in addition to the financial support of
Generalitat Valenciana under project
IDIFEDER/\allowbreak2021/\allowbreak059. % << Jorge Iranzo (3:1)
Zachary Hopton was supported by the Swiss National Science Foundation (Grant No. 10003607).
Vilém Zouhar gratefully acknowledges the support of the Google PhD Fellowship.

We extend our appreciation to Nuo Xu and David Kaczér for their human annotation effort.
%%%%%%

\bibliography{anthology.min,custom}
\bibliographystyle{acl_natbib}

\clearpage
% \onecolumn

\appendix

%%%% up to 5 pages
% \begin{figure*}[!ht]
%     \centering
%     \includegraphics[width=1\linewidth]{figs/pearmut_screenshot_1.png}
%     \caption{Screenshot of the Pearmut \citep{zouhar2026pearmut} annotation interface together with annotation guidelines. The annotator first listens to the source audio, then scans the three model outputs where they mark error spans with severities and categories. Lastly, the annotator assigns the final scores and proceeds to the next item.}
%     \label{fig:screenshot-pearmut}
% \end{figure*}

\twocolumn[{%
\centering
\includegraphics[width=1\linewidth]{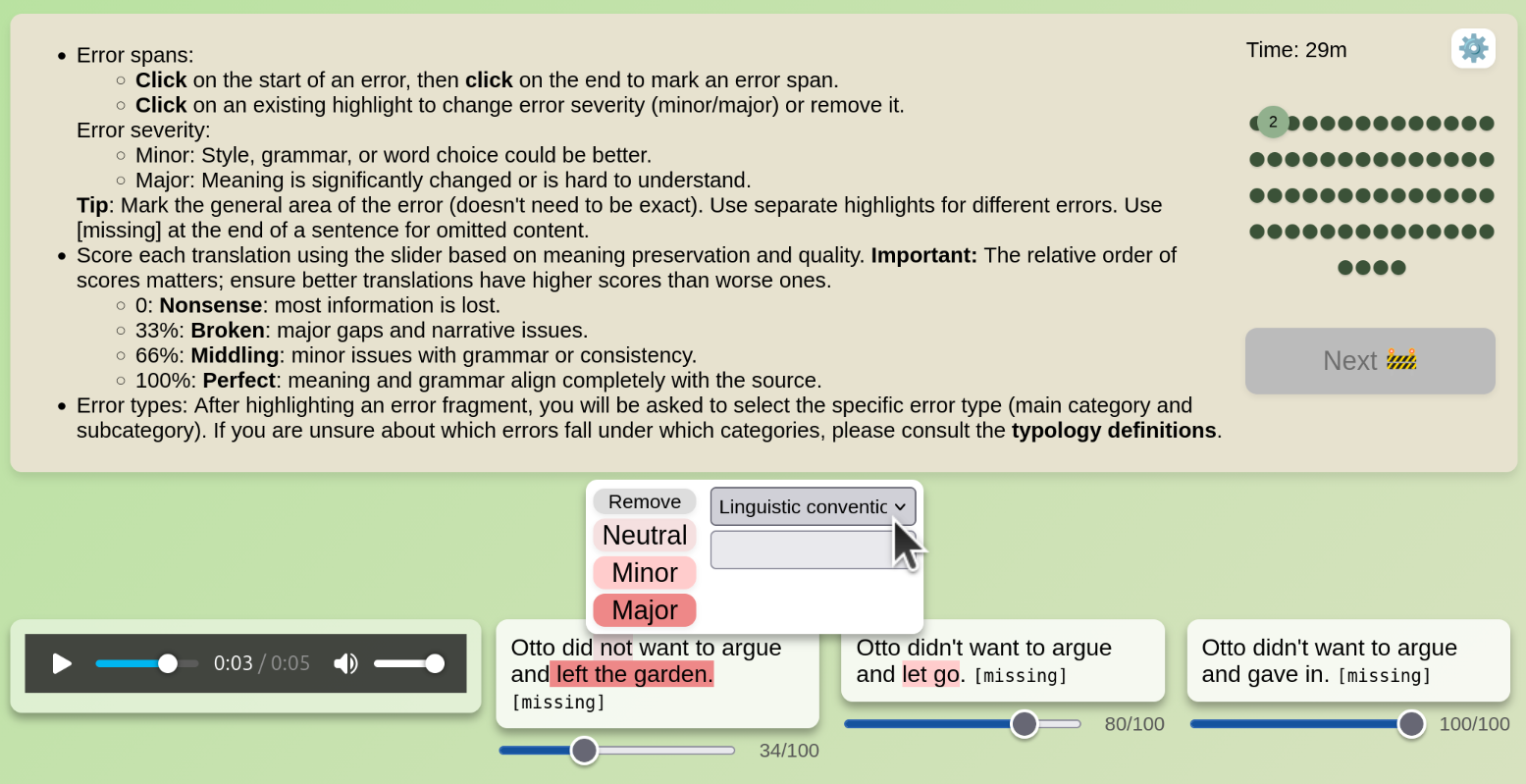}
\vspace{-2mm}
\captionof{figure}{Screenshot of the Pearmut \citep{zouhar2026pearmut} annotation interface together with annotation guidelines. The annotator first listens to the source audio, then scans the three model outputs where they mark error spans with severities and categories. Lastly, the annotator assigns the final scores and proceeds to the next item.}
\vspace{4mm}
\label{fig:screenshot-pearmut}
}]

\section{Human Evaluation Interface}
\label{app:humeval}

Annotation guidelines and interface are shown in \Cref{fig:screenshot-pearmut}.

\section{Evaluation Settings}
\label{app:eval-set}
All evaluations are conducted using Python 3.9.16. For \xcomet, we report scores with \texttt{unbabel/xcomet-xxl}. Scores were computed using the \texttt{comet} library (v2.2.2) with \texttt{fp32} precision. For \metricx, we report scores using {\small \texttt{google/metricx-24-hybrid-xxl-v2p6-bfloat16}}.

\section{Model Details}
\label{app:model-v}

The descriptions of the SFMs, LLMs, and SpeechLLMs used in our study are provided below. Models' weights, parameters, and library versions are reported \cref{tab:models-details}.

\begin{table*}[!ht]
\footnotesize
    \centering
    \setlength{\tabcolsep}{9pt}
    \begin{tabular}{llllllllllll}
    \toprule
    \textbf{Model} & \textbf{Param.} & \textbf{Weights} & \textbf{HFv} \\
    \midrule
    \makebox[0pt][l]{\hypertarget{aya}{}} \coloredsquare{cascadecolor_v2} \aya\ \citep{dang2024ayaexpanse} & 32B &  \href{https://hf.co/CohereLabs/aya-expanse-32b}{\texttt{CohereLabs/aya-expanse-32b}} & 4.51.0 \\
    \makebox[0pt][l]{\hypertarget{gemma3}{}} \coloredsquare{cascadecolor_v2} \gemma\ \citep{team2025gemma} & 12B & \href{https://hf.co/google/gemma-3-12b-it}{\texttt{google/gemma-3-12b-it}} & 4.51.0 \\
    \makebox[0pt][l]{\hypertarget{tower}{}} \coloredsquare{cascadecolor_v2} \tower\ \citep{rei2025tower+} & 9B &  \href{https://hf.co/Unbabel/Tower-Plus-9B}{\texttt{Unbabel/Tower-Plus-9B}} & 4.51.0 \\
    \makebox[0pt][l]{\hypertarget{whisper}{}} \coloredsquare{sfmcolor_v2} \whisper\ \citep{whisper-paper} & 1.6B & \href{https://hf.co/openai/whisper-large-v3}{\texttt{openai/whisper-large-v3}} & 4.51.3 \\
    \makebox[0pt][l]{\hypertarget{seamless}{}} \coloredsquare{sfmcolor_v2} \seamless\ \citep{barrault2023seamlessm4t} & 2.3B &  \href{https://hf.co/facebook/seamless-m4t-v2-large}{\texttt{facebook/seamless-m4t-v2-large}} & 4.51.3 \\
    \makebox[0pt][l]{\hypertarget{canary}{}} \coloredsquare{sfmcolor_v2} \canary\ \citep{canaryv2} & 1B &  \href{https://hf.co/nvidia/canary-1b-v2}{\texttt{nvidia/canary-1b-v2}}  & \crossmark{} \\
    \makebox[0pt][l]{\hypertarget{owsm}{}} \coloredsquare{sfmcolor_v2} \owsm\ \citep{peng25c_interspeech} & 1B & \href{https://hf.co/espnet/owsm_ctc_v4_1B}{\texttt{espnet/owsm\_ctc\_v4\_1B}} & \crossmark{} \\
    \makebox[0pt][l]{\hypertarget{desta2}{}} \coloredsquare{speechllmcolor_v2} \desta\ \citep{lu2024desta2} & 8B & \href{https://hf.co/DeSTA-ntu/DeSTA2-8B-beta}{\texttt{DeSTA-ntu/DeSTA2-8B-beta}} & 4.51.3 \\
    \makebox[0pt][l]{\hypertarget{phi4m}{}} \coloredsquare{speechllmcolor_v2} \phimultimodal\ \citep{abouelenin2025phi} & 5.6B &   \href{https://hf.co/microsoft/Phi-4-multimodal-instruct}{\texttt{microsoft/Phi-4-multimodal-instruct}} & 4.48.2 \\
    \makebox[0pt][l]{\hypertarget{qwen2a}{}} \coloredsquare{speechllmcolor_v2} \qwenaudio\ \citep{chu2024qwen2} & 7B & \href{https://hf.co/Qwen/Qwen2-Audio-7B-Instruct}{\texttt{Qwen/Qwen2-Audio-7B-Instruct}} & 4.51.3 \\
    \makebox[0pt][l]{\hypertarget{qwen3o}{}} \coloredsquare{speechllmcolor_v2} \qwenomni\ \citep{xu2025qwen3omnitechnicalreport} & 30B & \href{https://huggingface.co/Qwen/Qwen3-Omni-30B-A3B-Instruct}{\texttt{Qwen/Qwen3-Omni-30B-A3B-Instruct}} & 5.0.0 \\
    \makebox[0pt][l]{\hypertarget{spire}{}} \coloredsquare{speechllmcolor_v2} \spire\ \citep{ambilduke2025spire} & 7B &  \href{https://hf.co/utter-project/SpireFull}{\texttt{utter-project/SpireFull}} & 4.40.1 \\
    \makebox[0pt][l]{\hypertarget{voxtral}{}} \coloredsquare{speechllmcolor_v2} \voxtral\ \citep{liu2025voxtral} & 24B &  \href{https://hf.co/mistralai/Voxtral-Small-24B-2507}{\texttt{mistralai/Voxtral-Small-24B-2507}} & 4.54.0 \\
    \bottomrule
    \end{tabular}
    \caption{Details of the analyzed models, including the number of parameters, category (LLM\coloredsquare{cascadecolor_v2}, SFM\coloredsquare{sfmcolor_v2}, and SpeechLLM\coloredsquare{speechllmcolor_v2}), their public weights release, and the HuggingFace Transformer version (HFv).}
    \label{tab:models-details}
\end{table*}

\subsection{SFMs}

We select the most popular SFMs supporting translation tasks and covering a wide set of languages:

\paragraph{\whisper.} It is a Transformer encoder-decoder model trained on 
%680k hours of 
large-scale weakly and pseudo-labeled audio in many languages for ASR and direct many-to-English translation. We use the best-performing large-v3 model with 1.5B parameters that was trained on 5M hours for 99 languages, from which 58 achieve better than 50\% WER on the ASR task. % https://platform.openai.com/docs/guides/speech-to-text#supported-languages
%To process long-form audio, we use the pipeline in Transformers. 
% It enables the model to subsequently process 30-second chunks and shift them by timestamps generated by the model itself, along with other strategies such as temperature scaling and beam search, which aim to recover from timestamp errors. We do not use any voice activity detection tool with Whisper in this project. 
To process long-form audio, we adopt the chunked decoding pipeline provided in Transformers.\footnote{\href{https://huggingface.co/openai/whisper-large-v3\#chunked-long-form}{https://huggingface.co/openai/whisper-large-v3\#chunked-long-form}} This approach processes the input in 30-second segments, dynamically shifting the window based on timestamps predicted by the model itself. It also incorporates strategies such as temperature scaling and beam search to mitigate timestamp inaccuracies. We do not employ any external voice activity detection tool when using Whisper in this work.

\paragraph{\seamless.} SeamlessM4T is a foundational all-in-one \underline{M}assively \underline{M}ultilingual and \underline{M}ultimodal \underline{M}achine \underline{T}ranslation model covering multiple languages and modalities. We use the v2-large model, supporting 101-to-96 speech-to-text languages. For ST, it is composed of a Conformer encoder \cite{gulati20_interspeech} initialized from w2v-BERT 2.0 \citep{baevski2020wav2vec}, pretrained on over 4M hours, and a Transformer decoder initialized from NLLB \citep{costa2022no}. Since there is no standard implementation for processing long-form speech with this model, we do not process it in this work.

\paragraph{\canary.} It is FastConformer encoder \cite{Rekesh2023FastCW} and Transformer decoder model trained for ASR and English-to-X and X-to-English ST for 25 European languages. The model is trained on 1.7M hours contained in Granary \citep{raokoluguri25_interspeech}, covering various domains. We use the v2 version with 1B parameters, which is released under a permissive CC BY 4.0 license. Long-form audio is handled by the default implementation in the NeMo toolkit. It segments the audio into 30-40 second chunks, with a 1-second overlap between adjacent chunks. The overlapping transcripts are then merged with the longest common subsequence algorithm.

%Brief description. Language coverage. Specify the version + mention how it  handles long form speech

\paragraph{\owsm.} 
% The Open Whisper-style Speech Model (OWSM) 
% %has developed 
% is a series of fully open speech foundation models based on
% academic-scale resources and reproducible training pipelines, supporting over 151 languages. While originally closely following the Whisper architecture, newer versions have been released at a regular rate, improving its performance by extending the data, the prepossessing pipeline, or by adding newer, more powerful model architectures. At the time of writing, version 4.0 is the latest. Both an Encoder-Decoder and a CTC-based Encoder-only model exist. In this article, we use the latter. We make this decisions based on the reported robustness, especially for long-form input, faster inference and ST performance in contrast to its Encoder-Decoder counterpart. Additionally, to our knowledge, it is currently the only available large scale non-autoregressive ST and ASR model, making it really interesting for its consideration in this case study.
% Long-form inference is done with the batched parallel algorithm available on the Espnet~\citep{watanabe2018espnet}.
The \underline{O}pen \underline{W}hisper-style \underline{S}peech \underline{M}odel (OWSM) is a family of open speech foundation models trained on academic-scale resources with reproducible pipelines (about 166k hours), covering over 150 languages. Initially inspired by the Whisper architecture, successive releases have progressively improved performance through larger datasets, refined preprocessing, and more powerful architectures. We use the CTC-based encoder-only variant of OWSM 4.0 with 1B parameters (latest version at the time of writing) for its superior robustness on long-form input, faster inference, and stronger ST performance compared to its encoder-decoder counterpart. Moreover, it is currently the only large-scale non-autoregressive model supporting both ST and ASR, making it especially interesting for this study. Long-form inference is performed using the batched parallel decoding algorithm implemented in ESPnet~\citep{watanabe2018espnet}.

\subsection{LLMs}

% During the experiment design phase, we decided 
To maintain comparable sizes with existing SFMs and SpeechLLMs, and to allow easier reproducibility of the outputs, we choose to include one medium-sized model (20-40B parameters), one small model (<20B parameters), and one translation-specific LLM.
To select the actual models, we rely on the WMT25 General MT Findings \citep{kocmi-etal-2025-findings}, identifying the top-performing LLMs that met our size constraints for each language pair under consideration.

For the translation-specific and small-model categories, the choice was clear with Tower+ 9B and Gemma 3 12B standing out in their respective categories.
For the medium-sized category, we considered Aya Expanse 32B and Gemma 3 27B, and ultimately selected Aya Expanse due to its stronger performance across more language pairs as well as to promote model diversity in our selection.

\paragraph{\aya.}
Aya Expanse 32B is a decoder-only multilingual model built upon the Cohere Command architecture and optimized for 23 high-resource and mid-resource languages, covering all of the languages in our scope. It incorporates standard modern Transformer components such as SwiGLU activations \citep{shazeer2020gluvariantsimprovetransformer}, RoPE positional embeddings \citep{su2023roformerenhancedtransformerrotary}, and Grouped-Query Attention \citep{ainslie-etal-2023-gqa}. Its maximum context window is 128k tokens. The model is trained with a two-stage multilingual preference optimization pipeline: offline preference training followed by online preference training. It is further improved through weighted model merging across intermediate checkpoints. Aya Expanse combines strong general-purpose multilingual capabilities with competitive translation performance in a wide range of language pairs. The size of the training data is not disclosed publicly, but it was trained on multilingual LLM tasks, including translation, summarization, and question answering.

\paragraph{\gemma.}
Gemma 3 12B is a small multimodal model, supporting both image and text inputs and over 140 languages. Similarly to Aya, it offers a 128k-token context window and uses a decoder-only architecture with Grouped-Query Attention \citep{ainslie-etal-2023-gqa}. Training includes distillation from larger Gemma models and a post-training phase targeting multilingual and instruction-following performance. Gemma 3 12B offers a strong balance between model size, multilingual coverage, and general-purpose performance. The training data size has not been disclosed as well as the set of tasks, which, in general, include multimodal understanding: text, images, video, multimodal reasoning, and generation.

\paragraph{\tower.}
Tower+ 9B is a translation-focused model developed on top of the Gemma 2 9B foundation. Its training follows a four-stage recipe: Continued Pretraining (on 32 billion tokens) to strengthen multilingual representations, Instruction Tuning, Weighted Preference Optimization, and Reinforcement Learning with Verifiable Rewards. Tower+ surpasses larger general-purpose LLMs in translation quality in some of our selected language pairs, making it a competitive specialized option while being the smallest text LLM in our scope. It covers 47 language pairs, including 27 dialects.

\subsection{SpeechLLMs}

We select the SpeechLLMs available on HuggingFace and covering translation tasks (e.g., models covering English transcription only are discarded):

\paragraph{Phi-4-Multimodal.} Phi-4-Multimodal is a multimodal LLM that integrates text, image, and speech input modalities into a single model. The pre-trained speech encoder (consisting of 3 convolution layers--with a total subsampling rate of 8, and 80ms token rate--and 24 Conformer blocks; \citealp{gulati20_interspeech}) is connected with the Phi-4-Mini LLM through an audio adapter (a 2-layer MLP), and LoRA \citep{hu2022lora} is applied to the LLM. The training follows a 2-stage approach: a pre-training with large-scale ASR data (of approximately 2M hours) to align the speech encoder and the adapter with Phi-4-Mini in the semantic space (leaving only the LLM frozen), and a post-training with about 100M curated supervised samples (updating both the adapter and LoRA parameters only). The model covers 8 input languages: Chinese, English, French, German, Italian, Japanese, Portuguese, and Spanish. Given the 128k context length of the LLM, theoretically Phi-4-Multimodal can support a maximum of 2.8 hours of audio (as 750 tokens corresponds to 1-minute audio), but the model has not been finetuned on long audio data over 30 minutes.

\paragraph{Qwen2-Audio.} It is a large-scale SpeechLLM (Apache 2.0 license) featuring two distinct audio interaction modes for voice chat and audio analysis. In voice chat mode, users can engage in voice interactions without textual input. In the audio analysis mode, users can provide both audio and text instructions during the interaction. Qwen2-Audio is based on the Whisper large-v3 encoder with an additional pooling layer (performing a subsampling of 2) and Qwen-7B \citep{bai2023qwen} LLM. The model is first pre-trained on multiple tasks (including ASR) with natural language prompts, then it is fine-tuned with the two audio interaction modes, and, lastly, DPO \citep{rafailov2023direct} is applied. 

\paragraph{Qwen3-Omni.} It is the most recent Omni model from the Qwen family capable of processing text, speech, and video, and generating speech and text. It supports 119 languages for text, and 20 for speech understanding. Released under Apache 2.0 license, the model follows a Thinker-Talker Mixture of Experts architecture \citep{xu2025qwen25omnitechnicalreport} equipped with Qwen3-30B-A3B \citep{yang2025qwen3technicalreport} as LLM, Qwen3-VL \citep{bai2025qwen3vltechnicalreport} as video encoder, and an attention-based encoder-decoder speech encoder (with 0.6B parameters) trained from scratch on 20 million hours of supervised audio data, with 80\% Chinese and English pseudo-labeled ASR data, 10\% ASR data from other languages, and 10\% audio understanding data. The audio is transformed into filterbank features and then downsampled by a factor of 8 through Conv2D blocks. 
The model is pretrained following a three-stage approach: \textit{1)} encoder alignment, where the pretrained vision and speech encoders are loaded and the adapters are trained separately, \textit{2)} general stage during which the model is trained on all modalities (text, audio, image, video, and video-audio) on a large-scale dataset of 2 trillion tokens, and \textit{3)} long context, where the maximum token length is increased from 8,192 to 32,768 and longer audio and video are included in the training data. This is followed by a post-training phase, where instruction-following capabilities are introduced into the model by supervised fine-tuning, knowledge distillation from bigger models (Qwen3-32B or Qwen3-235B-A22B), and preference optimization (specifically, GSPO by \citealt{zheng2025groupsequencepolicyoptimization}).
% The specific training recipe has not fully published, but it has been trained on a wide range of tasks spanning text, image, audio, and video input processing, including ASR and translation.

\paragraph{DeSTA2.} It is a SpeechLLM built on Whisper-small and LLaMa 3 \citep{grattafiori2024llama}, augmented with a Q-Former adapter \citep{li2023blip}. It is trained on a mix of datasets totaling 155 hours (including speech with noise and reverberation) covering multiple tasks, with additional metadata such as speaker gender, age, accent, and emotion extracted from external models. Both the LLM and the Whisper components are kept frozen during training. Unlike the other SpeechLLMs considered in this study, DeSTA2 uses both the encoder and decoder of Whisper, providing the transcript alongside speech features to the LLM, implementing a hybrid between direct and cascaded architectures.

\paragraph{Voxtral.} Voxtral is a family of two open-weight SpeechLLMs (Apache 2.0 license) supporting 8 input languages (the ones used in this study plus Hindi), a context window of 32k tokens, and up to 40 minutes of speech input. The models are trained in three phases: pretraining (with speech-text interleaving; \citealt{nguyen-etal-2025-spirit}), supervised finetuning (with a mixture of synthetized data), and preference alignment (with standard and online DPO; \citealt{guo2024direct}). It was pretrained on large audio-text corpora (exact hours not disclosed) spanning tasks such as ASR, ST, audio question answering, audio summarization, and long-context audio understanding. We adopt the \texttt{small} version with 24B parameters that is made of the Whisper encoder, which processes the input in chunks of 30s, an MLP adapter, which maps the audio sequence in the LLM embedding space by also performing a downsampling of 4, and the Mistral Small 3.1 24B\footnote{\href{https://mistral.ai/news/mistral-small-3-1}{https://mistral.ai/news/mistral-small-3-1}} model as decoder.

\paragraph{Spire.} Spire is a speech-augmented LLM with 7B parameters, released under the CC-BY-NC 4.0 license. It builds on the multilingual LLM Tower \citep{alves2024tower} by introducing a discretized speech interface, where acoustic representations from HuBERT \citep{10.1109/TASLP.2021.3122291} are quantized with k-means clustering. Training follows a two-stage strategy: continued pretraining of TowerBase on mixed text-speech data (totaling 42k hours), and subsequent instruction tuning on text translation, ASR, and ST tasks. The main variant, SpireFull, preserves strong text-translation performance from Tower, while extending the model to English speech recognition and translation into 10 languages. It is important to note that the model is only instruction-tuned for speech recognition and translation tasks, and it relies on tightly defined instruction formats. As a result, its scope remains narrow, and Spire should be considered as a particular case of a SpeechLLM with no general-purpose capabilities.

\section{Prompts}
\label{app:prompts}

The prompts used for LLMs and SpeechLLMs\footnote{For \spire, we use the prompt template it was trained on:\newline\texttt{Speech: \textbf{\{DSUs\}}\textbackslash n\textbf{\{tgt\_lang\}}:}} are reported below. The \textbf{\texttt{\{src\_lang\}}} and \textbf{\texttt{\{tgt\_lang\}}} are replaced with the extended language name (e.g., \textbf{\texttt{English}} or \textbf{\texttt{Chinese (Simplified)}}).

\begin{promptbox}[colback=LLMcolor!20!white, colframe=LLMcolor!80]{LLMs Prompt}
You are a professional \textbf{\{src\_lang\}}-to-\textbf{\{tgt\_lang\}} translator. Your goal is to accurately convey
the meaning and nuances of the original \textbf{\{src\_lang\}} text while adhering to \textbf{\{tgt\_lang\}} grammar,
vocabulary, and cultural sensitivities. Preserve the line breaks. Use precise terminology
and a tone appropriate for academic or instructional materials. Produce only the \textbf{\{tgt\_lang\}}
translation, without any additional explanations or commentary. Please translate the
provided \textbf{\{src\_lang\}} text into \textbf{\{tgt\_lang\}}:
\end{promptbox}

\begin{promptbox}[colback=SpeechLLMcolor!20!white, colframe=SpeechLLMcolor!60]{SpeechLLMs Prompt}
You are a professional \textbf{\{src\_lang\}}-to-\textbf{\{tgt\_lang\}} translator. Your goal is to accurately convey
the meaning and nuances of the original \textbf{\{src\_lang\}} speech while adhering to \textbf{\{tgt\_lang\}}
grammar, vocabulary, and cultural sensitivities. Use precise terminology and a tone
appropriate for academic or instructional materials. Produce only the \textbf{\{tgt\_lang\}}
translation, without any additional explanations or commentary. Please translate the
provided \textbf{\{src\_lang\}} speech into \textbf{\{tgt\_lang\}}:
\end{promptbox}

\section{Limitations}
While our study provides a comprehensive evaluation of SpeechLLMs across multiple languages, benchmarks, and speech phenomena, it has a few inherent limitations. First, the analysis remains English-centric, reflecting the current language support of available SpeechLLMs. Expanding to a fully multilingual setup will require broader model coverage and additional resources. Second, we do not report results for traditional neural MT models, as our focus is on assessing the integration of speech within LLMs and the comparison with cascaded and direct speech-to-text translation pipelines. Third, we do not include toxicity or safety benchmarks, since no publicly available datasets currently target these aspects in speech-to-text translation. 
% Lastly, we do not report latency measurements, as we positioned our work in offline conditions, where real-time performance is not the main focus.
Despite these constraints, our work provides the first systematic, phenomenon-aware evaluation of SpeechLLMs, offering critical insights into their translation quality, robustness, and the practical trade-offs between integrated and modular architectures.

\clearpage

%%%% up to 3 pages
\onecolumn
\begin{multicols}{2}
\section{\metricxstrict{} Overall Results}
\label{app:overall-other-metrics}
We report the overall results using \metricxstrict\ in Table \ref{tab:st_metricX_color_stacked}. To ensure consistency with \cometstrict, where higher values indicate better performance, we transform the scores as $100 - 4 \cdot \metricxstrict$, mapping them to the $[0, 100]$ range.

\section{Accent-specific Results}
\label{app:accents}

\begin{figure*}[!h]
    \centering
    \includegraphics[width=1\linewidth]{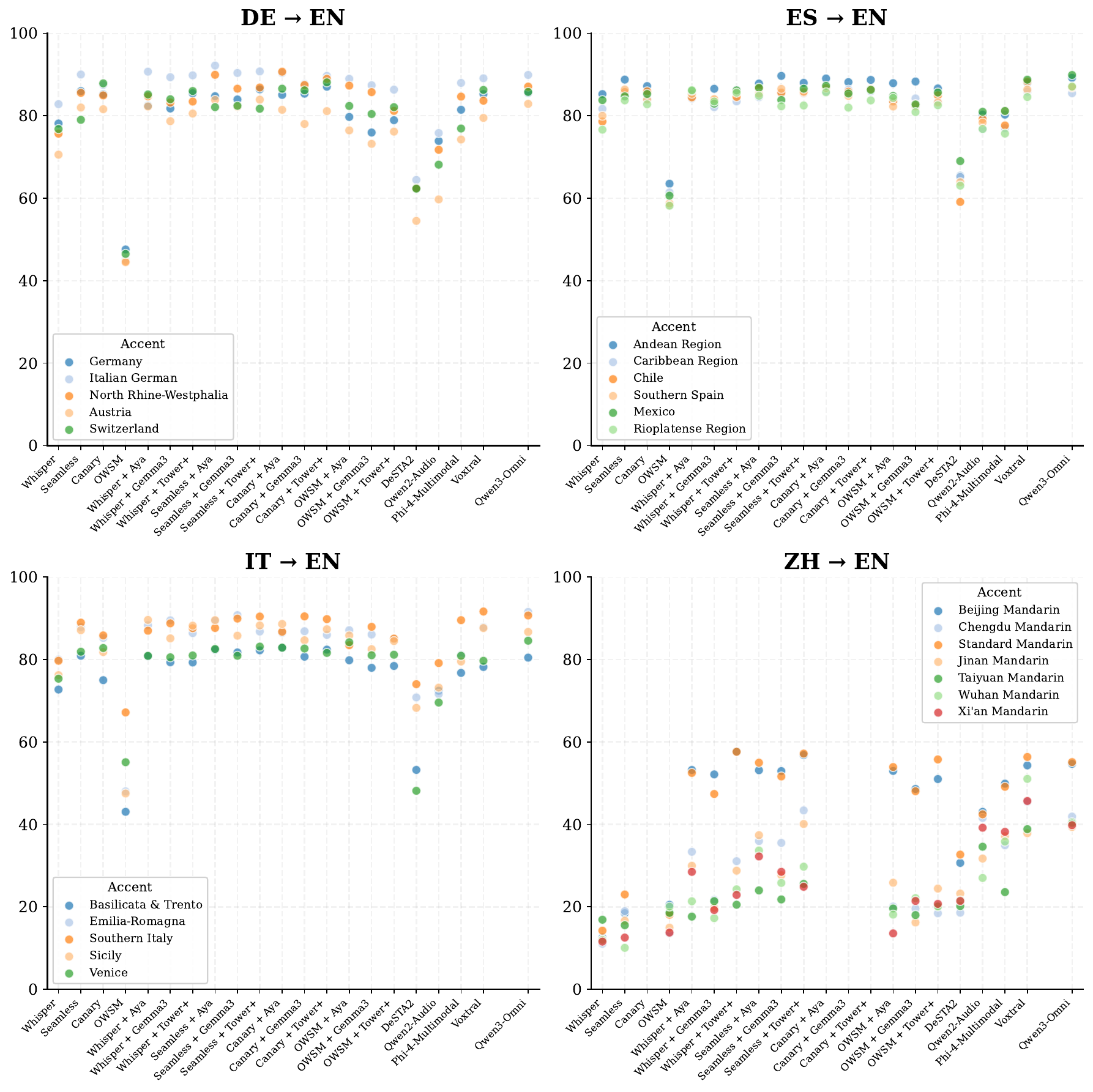}
    \caption{\cometstrict\ results for language pairs into English, broken down by source-language accent. zh-en results come from ManDi, while all other pairs represent CommonAccent results.}
    \label{fig:en_tgt_acc}
\end{figure*}

\begin{figure*}[h]
    \centering
    \includegraphics[width=1\linewidth]{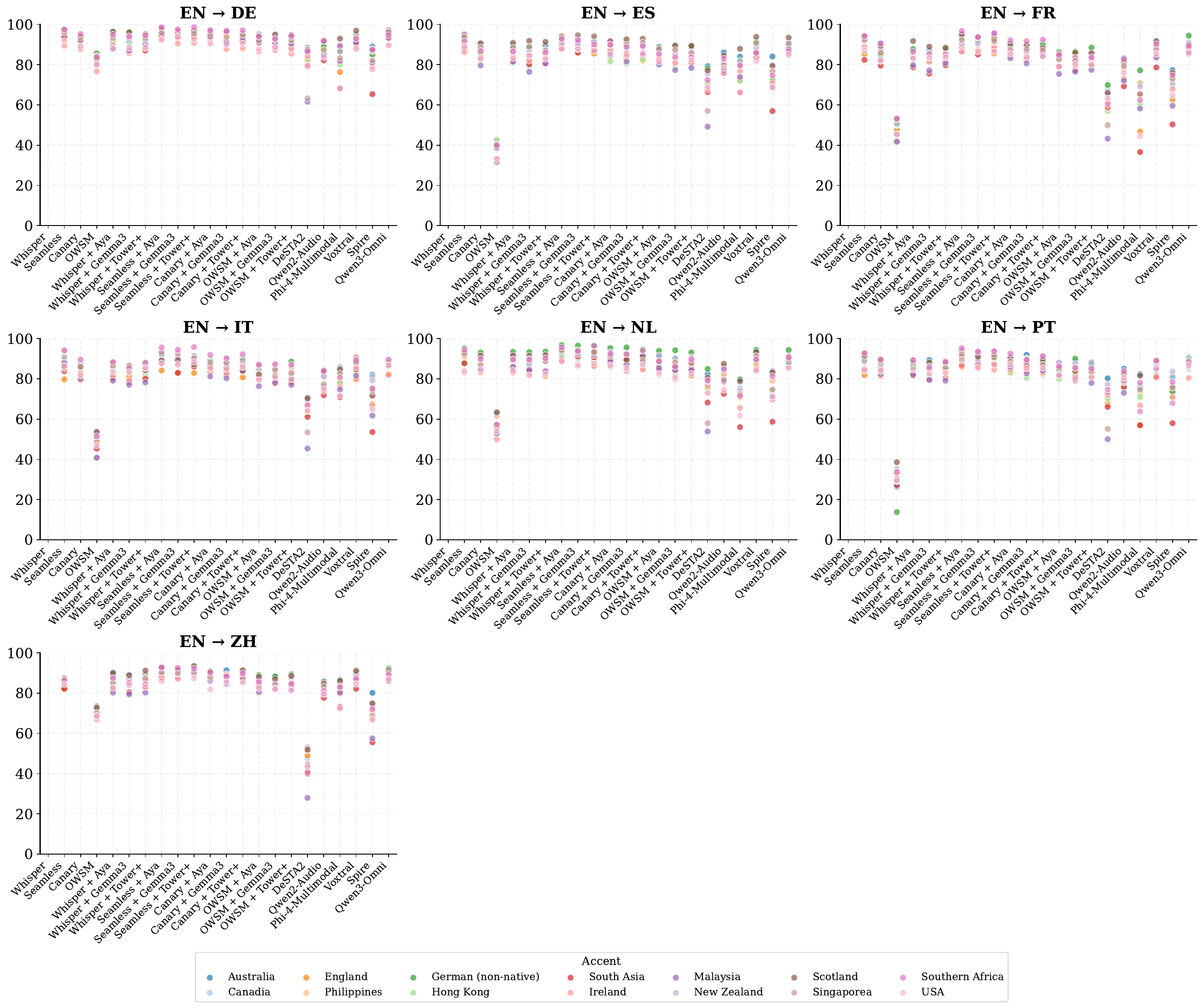}
    \caption{CommonAccent \cometstrict\ results for language pairs out of English, broken down by source speech accent.}
    \label{fig:en_src_acc}
\end{figure*}

Accent-specific results are presented in \cref{fig:en_src_acc,fig:en_tgt_acc}, and numeric values used to create \cref{fig:std_viz} are reported in \cref{tab:stdev}.
\end{multicols}

\begin{table}[H]
\centering
\scriptsize 
\setlength{\tabcolsep}{4pt}
\renewcommand{\arraystretch}{1.15}
\scalebox{0.8355}{%
\begin{tabular}{lccccccccccccccc}
%\toprule
\multirow{3}{*}{} &
&
\multicolumn{6}{c}{\scriptsize \GenericCat{} } &
\multicolumn{2}{c}{\scriptsize \GenderCat{} } &
\multicolumn{3}{c}{\scriptsize \AccentCat{} } &
\multicolumn{2}{c}{\scriptsize \CSCat{} } \\
\cmidrule(lr){2-8}
\cmidrule(lr){9-10}
\cmidrule(lr){11-13}
\cmidrule(lr){14-15}
\cmidrule(lr){14-15}
&
\multicolumn{2}{c}{\tiny FLEURS} &
\multicolumn{2}{c}{\tiny CoVoST2} &
\multicolumn{2}{c}{\tiny EuroParl-ST} &
\multicolumn{1}{c}{\tiny WMT} &
\multicolumn{2}{c}{\tiny FLEURS} &
\multicolumn{2}{c}{\tiny CommonAccent} &
\multicolumn{1}{c}{\tiny ManDi} &
\multicolumn{1}{c}{\tiny CS-Dialogue} &
\multicolumn{1}{c}{\tiny CS-FLEURS} \\
\cmidrule(lr){2-3}
\cmidrule(lr){4-5}
\cmidrule(lr){6-7}
\cmidrule(lr){8-8}
\cmidrule(lr){9-10}
\cmidrule(lr){11-12}
\cmidrule(lr){13-13}
\cmidrule(lr){14-14}
\cmidrule(lr){15-15}
&  \multicolumn{7}{c}{ {{\tiny \textsc{\textbf{MetricX}$^\text{QE}_\text{\textit{S}}$}}} }  &  
\multicolumn{2}{c}{\tiny \gendergap} &  
\multicolumn{2}{c}{ {{\tiny \textsc{\textbf{MetricX}$^\text{QE}_\text{\textit{S}}$}}} } & \accentgap & \multicolumn{2}{c}{ {{\tiny \textsc{\textbf{MetricX}$^\text{QE}_\text{\textit{S}}$}}} }    \\

& en-x & x-en & en-x & x-en & en-x & x-en & en-x & en-x & x-en & en-x & x-en & zh-en & zh-en & x-en \\
% \midrule
\specialrule{0.5pt}{0pt}{0pt}

\sfmbox{\whisper} & \cellcolor{GenericColor!3} - & \cellcolor{GenericColor!41} 83.0 & \cellcolor{GenericColor!3} - & \cellcolor{GenericColor!39} 72.8 & \cellcolor{GenericColor!3} - & \cellcolor{GenericColor!40} 78.7 & \cellcolor{GenericColor!3} - & \cellcolor{GenderColor!3} - & \cellcolor{GenderColor!45} 0.5 & \cellcolor{AccentColor!3} - & \cellcolor{AccentColor!38} 76.1 & \cellcolor{AccentColor!47} 32.5 & \cellcolor{CSColor!48} 69.8 & \cellcolor{CSColor!32} 70.9 \\
\sfmbox{\seamless} & \cellcolor{GenericColor!44} 86.2 & \cellcolor{GenericColor!46} 86.3 & \cellcolor{GenericColor!70} 86.1 & \cellcolor{GenericColor!50} 80.2 & \cellcolor{GenericColor!23} 70.9 & \cellcolor{GenericColor!46} 81.8 & \cellcolor{GenericColor!0} 26.4 & \cellcolor{GenderColor!48} -1.0 & \cellcolor{GenderColor!83} 0.1 & \cellcolor{AccentColor!77} 86.8 & \cellcolor{AccentColor!49} 82.1 & \cellcolor{AccentColor!35} 37.0 & \cellcolor{CSColor!43} 65.0 & \cellcolor{CSColor!48} 79.4 \\
\sfmbox{\canary} & \cellcolor{GenericColor!3} - & \cellcolor{GenericColor!3} - & \cellcolor{GenericColor!3} - & \cellcolor{GenericColor!26} 64.5 & \cellcolor{GenericColor!3} - & \cellcolor{GenericColor!49} 83.8 & \cellcolor{GenericColor!3} - & \cellcolor{GenderColor!3} - & \cellcolor{GenderColor!3} - & \cellcolor{AccentColor!3} - & \cellcolor{AccentColor!47} 80.8 & \cellcolor{AccentColor!3} - & \cellcolor{CSColor!3} - & \cellcolor{CSColor!3} - \\
\sfmbox{\owsm} & \cellcolor{GenericColor!0} 49.9 & \cellcolor{GenericColor!0} 51.9 & \cellcolor{GenericColor!0} 52.9 & \cellcolor{GenericColor!0} 48.0 & \cellcolor{GenericColor!0} 54.8 & \cellcolor{GenericColor!0} 55.0 & \cellcolor{GenericColor!4} 29.1 & \cellcolor{GenderColor!0} -7.1 & \cellcolor{GenderColor!0} 2.2 & \cellcolor{AccentColor!0} 50.2 & \cellcolor{AccentColor!0} 54.7 & \cellcolor{AccentColor!50} 31.6 & \cellcolor{CSColor!0} 27.4 & \cellcolor{CSColor!0} 53.4 \\
\cascadebox{\whisper\ + \aya} & \cellcolor{GenericColor!55} 91.3 & \cellcolor{GenericColor!88} 91.5 & \cellcolor{GenericColor!51} 84.8 & \cellcolor{GenericColor!79} 81.9 & \cellcolor{GenericColor!77} 90.3 & \cellcolor{GenericColor!75} 85.1 & \cellcolor{GenericColor!99} 80.4 & \cellcolor{GenderColor!64} -0.6 & \cellcolor{GenderColor!100} -0.0 & \cellcolor{AccentColor!53} 84.6 & \cellcolor{AccentColor!68} 83.7 & \cellcolor{AccentColor!72} 18.6 & \cellcolor{CSColor!100} 79.7 & \cellcolor{CSColor!84} 85.1 \\
\cascadebox{\hspace{24px} + \gemma} & \cellcolor{GenericColor!52} 91.2 & \cellcolor{GenericColor!75} 90.8 & \cellcolor{GenericColor!50} 84.7 & \cellcolor{GenericColor!60} 80.8 & \cellcolor{GenericColor!63} 89.9 & \cellcolor{GenericColor!50} 84.4 & \cellcolor{GenericColor!96} 78.6 & \cellcolor{GenderColor!64} -0.6 & \cellcolor{GenderColor!67} -0.2 & \cellcolor{AccentColor!50} 84.0 & \cellcolor{AccentColor!50} 82.6 & \cellcolor{AccentColor!48} 32.4 & \cellcolor{CSColor!88} 77.7 & \cellcolor{CSColor!75} 83.9 \\
\cascadebox{\hspace{24px} + \tower} & \cellcolor{GenericColor!50} 91.1 & \cellcolor{GenericColor!85} 91.3 & \cellcolor{GenericColor!50} 84.7 & \cellcolor{GenericColor!71} 81.4 & \cellcolor{GenericColor!73} 90.2 & \cellcolor{GenericColor!68} 84.9 & \cellcolor{GenericColor!95} 78.3 & \cellcolor{GenderColor!50} -0.8 & \cellcolor{GenderColor!45} 0.5 & \cellcolor{AccentColor!49} 83.9 & \cellcolor{AccentColor!63} 83.4 & \cellcolor{AccentColor!41} 34.6 & \cellcolor{CSColor!89} 77.8 & \cellcolor{CSColor!80} 84.6 \\
\cascadebox{\seamless\ + \aya} & \cellcolor{GenericColor!55} 91.3 & \cellcolor{GenericColor!67} 90.4 & \cellcolor{GenericColor!100} 88.1 & \cellcolor{GenericColor!90} 82.5 & \cellcolor{GenericColor!57} 89.7 & \cellcolor{GenericColor!86} 85.4 & \cellcolor{GenericColor!27} 42.9 & \cellcolor{GenderColor!48} -1.0 & \cellcolor{GenderColor!83} 0.1 & \cellcolor{AccentColor!100} 89.0 & \cellcolor{AccentColor!78} 84.3 & \cellcolor{AccentColor!86} 10.0 & \cellcolor{CSColor!80} 76.3 & \cellcolor{CSColor!65} 82.6 \\
\cascadebox{\hspace{26px} + \gemma} & \cellcolor{GenericColor!55} 91.3 & \cellcolor{GenericColor!56} 89.8 & \cellcolor{GenericColor!97} 87.9 & \cellcolor{GenericColor!72} 81.5 & \cellcolor{GenericColor!50} 89.4 & \cellcolor{GenericColor!61} 84.7 & \cellcolor{GenericColor!27} 42.9 & \cellcolor{GenderColor!47} -1.2 & \cellcolor{GenderColor!83} 0.1 & \cellcolor{AccentColor!96} 88.6 & \cellcolor{AccentColor!63} 83.4 & \cellcolor{AccentColor!65} 22.7 & \cellcolor{CSColor!56} 72.3 & \cellcolor{CSColor!55} 81.2 \\
\cascadebox{\hspace{26px} + \tower} & \cellcolor{GenericColor!55} 91.3 & \cellcolor{GenericColor!58} 89.9 & \cellcolor{GenericColor!97} 87.9 & \cellcolor{GenericColor!81} 82.0 & \cellcolor{GenericColor!53} 89.6 & \cellcolor{GenericColor!82} 85.3 & \cellcolor{GenericColor!25} 41.5 & \cellcolor{GenderColor!45} -1.4 & \cellcolor{GenderColor!67} 0.2 & \cellcolor{AccentColor!95} 88.5 & \cellcolor{AccentColor!70} 83.8 & \cellcolor{AccentColor!72} 18.8 & \cellcolor{CSColor!50} 71.2 & \cellcolor{CSColor!60} 81.9 \\
\cascadebox{Canary + \aya} & \cellcolor{GenericColor!61} 91.5 & \cellcolor{GenericColor!3} - & \cellcolor{GenericColor!72} 86.2 & \cellcolor{GenericColor!3} - & \cellcolor{GenericColor!100} 91.0 & \cellcolor{GenericColor!100} 85.8 & \cellcolor{GenericColor!99} 80.3 & \cellcolor{GenderColor!93} -0.2 & \cellcolor{GenderColor!3} - & \cellcolor{AccentColor!73} 86.5 & \cellcolor{AccentColor!80} 84.4 & \cellcolor{AccentColor!3} - & \cellcolor{CSColor!3} - & \cellcolor{CSColor!3} - \\
\cascadebox{\hspace{20px} + \gemma} & \cellcolor{GenericColor!58} 91.4 & \cellcolor{GenericColor!3} - & \cellcolor{GenericColor!67} 85.9 & \cellcolor{GenericColor!3} - & \cellcolor{GenericColor!87} 90.6 & \cellcolor{GenericColor!71} 85.0 & \cellcolor{GenericColor!98} 79.8 & \cellcolor{GenderColor!79} -0.4 & \cellcolor{GenderColor!3} - & \cellcolor{AccentColor!68} 86.0 & \cellcolor{AccentColor!60} 83.2 & \cellcolor{AccentColor!3} - & \cellcolor{CSColor!3} - & \cellcolor{CSColor!3} - \\
\cascadebox{\hspace{20px} + \tower} & \cellcolor{GenericColor!58} 91.4 & \cellcolor{GenericColor!3} - & \cellcolor{GenericColor!67} 85.9 & \cellcolor{GenericColor!3} - & \cellcolor{GenericColor!97} 90.9 & \cellcolor{GenericColor!93} 85.6 & \cellcolor{GenericColor!95} 78.2 & \cellcolor{GenderColor!100} 0.1 & \cellcolor{GenderColor!3} - & \cellcolor{AccentColor!68} 86.0 & \cellcolor{AccentColor!72} 83.9 & \cellcolor{AccentColor!3} - & \cellcolor{CSColor!3} - & \cellcolor{CSColor!3} - \\
\cascadebox{OWSM + \aya} & \cellcolor{GenericColor!48} 89.9 & \cellcolor{GenericColor!50} 89.5 & \cellcolor{GenericColor!49} 84.4 & \cellcolor{GenericColor!64} 81.0 & \cellcolor{GenericColor!49} 89.0 & \cellcolor{GenericColor!48} 83.5 & \cellcolor{GenericColor!54} 58.8 & \cellcolor{GenderColor!50} -0.8 & \cellcolor{GenderColor!67} 0.2 & \cellcolor{AccentColor!49} 83.6 & \cellcolor{AccentColor!50} 82.4 & \cellcolor{AccentColor!48} 32.4 & \cellcolor{CSColor!50} 70.9 & \cellcolor{CSColor!50} 80.3 \\
\cascadebox{\hspace{22px} + \gemma} & \cellcolor{GenericColor!48} 89.7 & \cellcolor{GenericColor!48} 88.2 & \cellcolor{GenericColor!49} 84.1 & \cellcolor{GenericColor!49} 79.5 & \cellcolor{GenericColor!49} 88.6 & \cellcolor{GenericColor!47} 82.4 & \cellcolor{GenericColor!52} 57.9 & \cellcolor{GenderColor!100} 0.1 & \cellcolor{GenderColor!83} 0.1 & \cellcolor{AccentColor!48} 83.2 & \cellcolor{AccentColor!47} 81.1 & \cellcolor{AccentColor!0} 49.2 & \cellcolor{CSColor!43} 65.3 & \cellcolor{CSColor!45} 77.6 \\
\cascadebox{\hspace{22px} + \tower} & \cellcolor{GenericColor!48} 89.5 & \cellcolor{GenericColor!49} 89.0 & \cellcolor{GenericColor!49} 84.0 & \cellcolor{GenericColor!50} 80.2 & \cellcolor{GenericColor!49} 88.8 & \cellcolor{GenericColor!48} 83.2 & \cellcolor{GenericColor!46} 54.5 & \cellcolor{GenderColor!79} -0.4 & \cellcolor{GenderColor!42} 0.6 & \cellcolor{AccentColor!48} 82.9 & \cellcolor{AccentColor!47} 81.0 & \cellcolor{AccentColor!5} 47.5 & \cellcolor{CSColor!44} 65.7 & \cellcolor{CSColor!46} 78.4 \\
\speechllmbox{\desta} & \cellcolor{GenericColor!36} 79.3 & \cellcolor{GenericColor!42} 83.4 & \cellcolor{GenericColor!25} 68.6 & \cellcolor{GenericColor!34} 70.2 & \cellcolor{GenericColor!13} 64.0 & \cellcolor{GenericColor!37} 76.9 & \cellcolor{GenericColor!48} 55.7 & \cellcolor{GenderColor!38} -2.3 & \cellcolor{GenderColor!13} -1.7 & \cellcolor{AccentColor!26} 68.0 & \cellcolor{AccentColor!31} 72.1 & \cellcolor{AccentColor!48} 32.3 & \cellcolor{CSColor!59} 72.8 & \cellcolor{CSColor!43} 76.5 \\
\speechllmbox{\qwenaudio} & \cellcolor{GenericColor!36} 79.9 & \cellcolor{GenericColor!41} 82.5 & \cellcolor{GenericColor!40} 78.4 & \cellcolor{GenericColor!41} 74.7 & \cellcolor{GenericColor!42} 83.9 & \cellcolor{GenericColor!41} 79.3 & \cellcolor{GenericColor!35} 47.9 & \cellcolor{GenderColor!43} -1.7 & \cellcolor{GenderColor!50} 0.3 & \cellcolor{AccentColor!41} 78.3 & \cellcolor{AccentColor!37} 75.4 & \cellcolor{AccentColor!100} 1.9 & \cellcolor{CSColor!58} 72.6 & \cellcolor{CSColor!49} 79.9 \\
\speechllmbox{\phimultimodal} & \cellcolor{GenericColor!22} 68.1 & \cellcolor{GenericColor!46} 86.6 & \cellcolor{GenericColor!10} 59.0 & \cellcolor{GenericColor!36} 71.0 & \cellcolor{GenericColor!16} 66.2 & \cellcolor{GenericColor!36} 76.4 & \cellcolor{GenericColor!38} 49.6 & \cellcolor{GenderColor!43} -1.7 & \cellcolor{GenderColor!39} 0.7 & \cellcolor{AccentColor!31} 71.6 & \cellcolor{AccentColor!41} 77.5 & \cellcolor{AccentColor!88} 8.9 & \cellcolor{CSColor!39} 61.8 & \cellcolor{CSColor!50} 80.5 \\
\speechllmbox{\voxtral} & \cellcolor{GenericColor!100} 92.7 & \cellcolor{GenericColor!79} 91.0 & \cellcolor{GenericColor!60} 85.4 & \cellcolor{GenericColor!50} 80.1 & \cellcolor{GenericColor!73} 90.2 & \cellcolor{GenericColor!61} 84.7 & \cellcolor{GenericColor!98} 79.8 & \cellcolor{GenderColor!49} -0.9 & \cellcolor{GenderColor!42} 0.6 & \cellcolor{AccentColor!65} 85.7 & \cellcolor{AccentColor!62} 83.3 & \cellcolor{AccentColor!84} 11.3 & \cellcolor{CSColor!99} 79.5 & \cellcolor{CSColor!90} 86.0 \\
\speechllmbox{\spire} & \cellcolor{GenericColor!38} 81.3 & \cellcolor{GenericColor!3} - & \cellcolor{GenericColor!27} 70.2 & \cellcolor{GenericColor!3} - & \cellcolor{GenericColor!41} 83.6 & \cellcolor{GenericColor!3} - & \cellcolor{GenericColor!42} 52.0 & \cellcolor{GenderColor!64} -0.6 & \cellcolor{GenderColor!3} - & \cellcolor{AccentColor!34} 73.4 & \cellcolor{AccentColor!3} - & \cellcolor{AccentColor!3} - & \cellcolor{CSColor!3} - & \cellcolor{CSColor!3} - \\
\speechllmbox{\qwenthreeomni} & \cellcolor{GenericColor!71} 91.8 & \cellcolor{GenericColor!100} 92.1 & \cellcolor{GenericColor!85} 87.1 & \cellcolor{GenericColor!100} 83.1 & \cellcolor{GenericColor!87} 90.6 & \cellcolor{GenericColor!100} 85.8 & \cellcolor{GenericColor!100} 80.7 & \cellcolor{GenderColor!79} -0.4 & \cellcolor{GenderColor!50} -0.3 & \cellcolor{AccentColor!70} 86.2 & \cellcolor{AccentColor!100} 85.6 & \cellcolor{AccentColor!88} -9.2 & \cellcolor{CSColor!50} 71.0 & \cellcolor{CSColor!100} 87.3 \\

\specialrule{1pt}{0pt}{0pt}
% \bottomrule
\end{tabular}%
}

% ==========================
% SPACER
% ==========================
\vspace{0.05cm} % Adjust this space as needed

% ==========================
% SECOND TABLE (BOTTOM)
% ==========================
\scalebox{0.8}{%
\begin{tabular}{lcccccccccccccc}

\multirow{3}{*}{} &

\multicolumn{1}{c}{\scriptsize \DisfluentCat{} } &
\multicolumn{4}{c}{\scriptsize  \NoiseCat{} } &
\multicolumn{2}{c}{\scriptsize \EmotionCat{} } &
\multicolumn{2}{c}{\scriptsize \LongCat{} } \\
\cmidrule(lr){2-2}
\cmidrule(lr){3-6}
\cmidrule(lr){7-8}
\cmidrule(lr){9-10}
\cmidrule(lr){11-12}
&
\multicolumn{1}{c}{\tiny LibriStutter} &
\multicolumn{2}{c}{\tiny NoisyFLEURS\textsubscript{B}	} &
\multicolumn{2}{c}{\tiny NoisyFLEURS\textsubscript{A}} &
\multicolumn{1}{c}{\tiny mExpresso} &
\multicolumn{1}{c}{\tiny EmotionTalk} &
\multicolumn{1}{c}{\tiny ACL6060} &
\multicolumn{1}{c}{\tiny MCIF} \\
\cmidrule(lr){2-2}
\cmidrule(lr){3-4}
\cmidrule(lr){5-6}
\cmidrule(lr){7-7}
\cmidrule(lr){8-8}
\cmidrule(lr){9-9}
\cmidrule(lr){10-10}
& {\tiny \difluentgap} & \multicolumn{4}{c}{\noisegap} & \multicolumn{2}{c}{ {{\tiny \textsc{\textbf{MetricX}$^\text{QE}_\text{\textit{S}}$}}} } & \multicolumn{2}{c}{{\tiny \lengthgap} } \\
& en-x & en-x & x-en & en-x & x-en & en-x & zh-en & en-x & en-x \\
% \midrule
\specialrule{0.5pt}{0pt}{0pt}

\sfmboxl{\whisper} & \cellcolor{DisfluentColor!3} - & \cellcolor{NoiseColor!3} - & \cellcolor{NoiseColor!55} 48.6 & \cellcolor{NoiseColor!3} - & \cellcolor{NoiseColor!49} 10.1 & \cellcolor{EmotionColor!3} - & \cellcolor{EmotionColor!44} 75.2 & \cellcolor{LongColor!3} - & \cellcolor{LongColor!3} - \\
\sfmboxl{\seamless} & \cellcolor{DisfluentColor!0} 36.6 & \cellcolor{NoiseColor!43} 59.2 & \cellcolor{NoiseColor!50} 51.0 & \cellcolor{NoiseColor!48} 11.4 & \cellcolor{NoiseColor!50} 9.0 & \cellcolor{EmotionColor!44} 80.1 & \cellcolor{EmotionColor!43} 74.0 & \cellcolor{LongColor!3} - & \cellcolor{LongColor!3} - \\
\sfmboxl{\canary} & \cellcolor{DisfluentColor!3} - & \cellcolor{NoiseColor!3} - & \cellcolor{NoiseColor!3} - & \cellcolor{NoiseColor!3} - & \cellcolor{NoiseColor!3} - & \cellcolor{EmotionColor!3} - & \cellcolor{EmotionColor!3} - & \cellcolor{LongColor!3} - & \cellcolor{LongColor!3} - \\
\sfmboxl{\owsm} & \cellcolor{DisfluentColor!19} 25.7 & \cellcolor{NoiseColor!8} 75.8 & \cellcolor{NoiseColor!0} 68.4 & \cellcolor{NoiseColor!44} 19.8 & \cellcolor{NoiseColor!45} 15.8 & \cellcolor{EmotionColor!29} 63.7 & \cellcolor{EmotionColor!0} 32.9 & \cellcolor{LongColor!38} 24.7 & \cellcolor{LongColor!46} 9.3 \\
\cascadeboxl{\whisper\ + \aya} & \cellcolor{DisfluentColor!100} 2.7 & \cellcolor{NoiseColor!61} 50.8 & \cellcolor{NoiseColor!57} 47.9 & \cellcolor{NoiseColor!60} 7.1 & \cellcolor{NoiseColor!51} 8.7 & \cellcolor{EmotionColor!100} 87.8 & \cellcolor{EmotionColor!100} 84.7 & \cellcolor{LongColor!50} 3.8 & \cellcolor{LongColor!49} 4.3 \\
\cascadeboxl{\hspace{24px} + \gemma} & \cellcolor{DisfluentColor!94} 3.4 & \cellcolor{NoiseColor!60} 51.3 & \cellcolor{NoiseColor!56} 48.3 & \cellcolor{NoiseColor!57} 7.4 & \cellcolor{NoiseColor!53} 8.6 & \cellcolor{EmotionColor!69} 86.7 & \cellcolor{EmotionColor!81} 83.1 & \cellcolor{LongColor!50} 3.7 & \cellcolor{LongColor!49} 4.4 \\
\cascadeboxl{\hspace{24px} + \tower} & \cellcolor{DisfluentColor!89} 3.9 & \cellcolor{NoiseColor!63} 50.1 & \cellcolor{NoiseColor!65} 44.2 & \cellcolor{NoiseColor!60} 7.1 & \cellcolor{NoiseColor!61} 8.0 & \cellcolor{EmotionColor!71} 86.8 & \cellcolor{EmotionColor!77} 82.8 & \cellcolor{LongColor!49} 4.8 & \cellcolor{LongColor!48} 5.2 \\
\cascadeboxl{\seamless\ + \aya} & \cellcolor{DisfluentColor!55} 7.6 & \cellcolor{NoiseColor!54} 53.9 & \cellcolor{NoiseColor!48} 51.7 & \cellcolor{NoiseColor!50} 8.4 & \cellcolor{NoiseColor!57} 8.3 & \cellcolor{EmotionColor!57} 86.3 & \cellcolor{EmotionColor!99} 84.6 & \cellcolor{LongColor!3} - & \cellcolor{LongColor!3} - \\
\cascadeboxl{\hspace{26px} + \gemma} & \cellcolor{DisfluentColor!45} 11.3 & \cellcolor{NoiseColor!55} 53.8 & \cellcolor{NoiseColor!43} 53.5 & \cellcolor{NoiseColor!50} 8.3 & \cellcolor{NoiseColor!50} 8.8 & \cellcolor{EmotionColor!49} 85.4 & \cellcolor{EmotionColor!73} 82.4 & \cellcolor{LongColor!3} - & \cellcolor{LongColor!3} - \\
\cascadeboxl{\hspace{26px} + \tower} & \cellcolor{DisfluentColor!47} 9.7 & \cellcolor{NoiseColor!53} 54.5 & \cellcolor{NoiseColor!44} 53.1 & \cellcolor{NoiseColor!50} 8.7 & \cellcolor{NoiseColor!53} 8.6 & \cellcolor{EmotionColor!49} 84.9 & \cellcolor{EmotionColor!75} 82.6 & \cellcolor{LongColor!3} - & \cellcolor{LongColor!3} - \\
\cascadeboxl{Canary + \aya} & \cellcolor{DisfluentColor!66} 6.4 & \cellcolor{NoiseColor!47} 57.4 & \cellcolor{NoiseColor!3} - & \cellcolor{NoiseColor!53} 7.8 & \cellcolor{NoiseColor!3} - & \cellcolor{EmotionColor!94} 87.6 & \cellcolor{EmotionColor!3} - & \cellcolor{LongColor!66} -2.7 & \cellcolor{LongColor!94} -0.3 \\
\cascadeboxl{\hspace{20px} + \gemma} & \cellcolor{DisfluentColor!48} 9.6 & \cellcolor{NoiseColor!46} 57.6 & \cellcolor{NoiseColor!3} - & \cellcolor{NoiseColor!56} 7.5 & \cellcolor{NoiseColor!3} - & \cellcolor{EmotionColor!63} 86.5 & \cellcolor{EmotionColor!3} - & \cellcolor{LongColor!80} -1.8 & \cellcolor{LongColor!75} 0.9 \\
\cascadeboxl{\hspace{20px} + \tower} & \cellcolor{DisfluentColor!50} 8.2 & \cellcolor{NoiseColor!45} 58.3 & \cellcolor{NoiseColor!3} - & \cellcolor{NoiseColor!52} 7.9 & \cellcolor{NoiseColor!3} - & \cellcolor{EmotionColor!69} 86.7 & \cellcolor{EmotionColor!3} - & \cellcolor{LongColor!83} -1.6 & \cellcolor{LongColor!81} 0.7 \\
\cascadeboxl{OWSM + \aya} & \cellcolor{DisfluentColor!59} 7.2 & \cellcolor{NoiseColor!29} 65.6 & \cellcolor{NoiseColor!19} 61.9 & \cellcolor{NoiseColor!47} 12.8 & \cellcolor{NoiseColor!48} 12.4 & \cellcolor{EmotionColor!69} 86.7 & \cellcolor{EmotionColor!70} 82.2 & \cellcolor{LongColor!70} -2.4 & \cellcolor{LongColor!69} -1.1 \\
\cascadeboxl{\hspace{22px} + \gemma} & \cellcolor{DisfluentColor!46} 10.6 & \cellcolor{NoiseColor!23} 68.5 & \cellcolor{NoiseColor!11} 64.7 & \cellcolor{NoiseColor!47} 13.1 & \cellcolor{NoiseColor!47} 13.6 & \cellcolor{EmotionColor!50} 85.6 & \cellcolor{EmotionColor!49} 79.5 & \cellcolor{LongColor!73} -2.2 & \cellcolor{LongColor!50} -1.7 \\
\cascadeboxl{\hspace{22px} + \tower} & \cellcolor{DisfluentColor!50} 8.2 & \cellcolor{NoiseColor!0} 79.4 & \cellcolor{NoiseColor!8} 65.7 & \cellcolor{NoiseColor!0} 98.8 & \cellcolor{NoiseColor!0} 84.6 & \cellcolor{EmotionColor!50} 85.8 & \cellcolor{EmotionColor!50} 80.5 & \cellcolor{LongColor!63} -2.9 & \cellcolor{LongColor!62} -1.3 \\
\speechllmboxl{\desta} & \cellcolor{DisfluentColor!61} 7.0 & \cellcolor{NoiseColor!21} 69.5 & \cellcolor{NoiseColor!6} 66.4 & \cellcolor{NoiseColor!45} 17.8 & \cellcolor{NoiseColor!44} 17.9 & \cellcolor{EmotionColor!39} 74.9 & \cellcolor{EmotionColor!46} 77.0 & \cellcolor{LongColor!3} 88.6 & \cellcolor{LongColor!1} 90.3 \\
\speechllmboxl{\qwenaudio} & \cellcolor{DisfluentColor!44} 11.5 & \cellcolor{NoiseColor!88} 38.4 & \cellcolor{NoiseColor!50} 50.9 & \cellcolor{NoiseColor!51} 8.0 & \cellcolor{NoiseColor!48} 12.4 & \cellcolor{EmotionColor!40} 76.1 & \cellcolor{EmotionColor!48} 78.7 & \cellcolor{LongColor!0} 94.9 & \cellcolor{LongColor!0} 92.0 \\
\speechllmboxl{\phimultimodal} & \cellcolor{DisfluentColor!43} 12.3 & \cellcolor{NoiseColor!54} 53.9 & \cellcolor{NoiseColor!100} 27.9 & \cellcolor{NoiseColor!84} 4.6 & \cellcolor{NoiseColor!95} 5.4 & \cellcolor{EmotionColor!0} 33.7 & \cellcolor{EmotionColor!44} 74.7 & \cellcolor{LongColor!47} -9.7 & \cellcolor{LongColor!40} 19.5 \\
\speechllmboxl{\voxtral} & \cellcolor{DisfluentColor!100} 2.7 & \cellcolor{NoiseColor!97} 34.0 & \cellcolor{NoiseColor!80} 37.3 & \cellcolor{NoiseColor!90} 4.0 & \cellcolor{NoiseColor!96} 5.3 & \cellcolor{EmotionColor!71} 86.8 & \cellcolor{EmotionColor!50} 80.2 & \cellcolor{LongColor!100} -0.5 & \cellcolor{LongColor!66} 1.2 \\
\speechllmboxl{\spire} & \cellcolor{DisfluentColor!45} 10.9 & \cellcolor{NoiseColor!4} 77.6 & \cellcolor{NoiseColor!3} - & \cellcolor{NoiseColor!30} 43.9 & \cellcolor{NoiseColor!3} - & \cellcolor{EmotionColor!42} 77.7 & \cellcolor{EmotionColor!3} - & \cellcolor{LongColor!3} - & \cellcolor{LongColor!3} - \\
\speechllmboxl{\qwenthreeomni} & \cellcolor{DisfluentColor!82} 4.7 & \cellcolor{NoiseColor!100} 32.6 & \cellcolor{NoiseColor!75} 39.4 & \cellcolor{NoiseColor!100} 3.0 & \cellcolor{NoiseColor!100} 5.0 & \cellcolor{EmotionColor!77} 87.0 & \cellcolor{EmotionColor!85} 83.4 & \cellcolor{LongColor!50} 4.1 & \cellcolor{LongColor!100} -0.1 \\

% \bottomrule
\specialrule{1pt}{0pt}{0pt}
\end{tabular}%
}

\caption{Overall performance
% of the 21 evaluated systems computed using
computed with
\metricxstrict. \genderfone, \neacc, and \termacc\ metrics are excluded as they are not computed via QE models. en-x denotes averages across all target languages, except where each benchmark covers a specific subset.
% (e.g., ACL 60/60: de/fr/zh/pt; MCIF: de/it/zh). 
x-en denotes averages across all source languages for each benchmark, as per Table \ref{tab:bench-sum}.}
\label{tab:st_metricX_color_stacked}
\end{table}

\twocolumn

% \section{\metricxstrict{} Overall Results}
% \label{app:overall-other-metrics}
% %\nopagebreak
% \input{tables/table_results_overall_2}

% We report the overall results using \metricxstrict\ in Table \ref{tab:st_metricX_color_stacked}. To ensure consistency with \cometstrict, where higher values indicate better performance, we transform the scores as $100 - 4 \cdot \metricxstrict$, mapping them to the $[0, 100]$ range.

% We report the overall results using \metricxstrict\ in Table \ref{tab:st_metricX_color_stacked} (on the next page). To ensure consistency with \cometstrict\ where higher values indicate better performance, we transform the scores using the formula $100 - 4 \cdot \metricxstrict$, mapping the values to the $[0, 100]$ range. 

\newpage

\clearpage

\clearpage

\begin{table*}[!ht]
    \centering
    \footnotesize
    \scalebox{0.94}{\begin{tabular}{lrrrrrrrrrrr}
% \toprule
\specialrule{1pt}{0pt}{0pt}
System & en$\rightarrow$de & en$\rightarrow$es & en$\rightarrow$fr & en$\rightarrow$it & en$\rightarrow$nl & en$\rightarrow$pt & en$\rightarrow$zh & de$\rightarrow$en & es$\rightarrow$en & it$\rightarrow$en & zh$\rightarrow$en \\ 
% \midrule
\specialrule{0.5pt}{0pt}{0pt}
\cellcolor{sfmcolor} \whisper & \cellcolor{white} - & \cellcolor{white} - & \cellcolor{white} - & \cellcolor{white} - & \cellcolor{white} - & \cellcolor{white} - & \cellcolor{white} - & \cellcolor[RGB]{223,242,230} 0.044 & \cellcolor[RGB]{233,246,238} 0.032 & \cellcolor[RGB]{235,247,239} 0.031 & \cellcolor[RGB]{244,250,246} 0.020 \\
\cellcolor{sfmcolor} \seamless & \cellcolor[RGB]{241,249,244} 0.023 & \cellcolor[RGB]{241,249,244} 0.024 & \cellcolor[RGB]{229,245,235} 0.037 & \cellcolor[RGB]{227,244,234} 0.039 & \cellcolor[RGB]{228,244,234} 0.038 & \cellcolor[RGB]{230,245,235} 0.036 & \cellcolor[RGB]{246,251,248} 0.018 & \cellcolor[RGB]{225,243,232} 0.042 & \cellcolor[RGB]{247,251,248} 0.017 & \cellcolor[RGB]{229,245,235} 0.037 & \cellcolor[RGB]{224,242,231} 0.043 \\
\cellcolor{sfmcolor} \canary & \cellcolor[RGB]{243,250,245} 0.022 & \cellcolor[RGB]{233,246,238} 0.033 & \cellcolor[RGB]{231,245,236} 0.035 & \cellcolor[RGB]{236,247,240} 0.030 & \cellcolor[RGB]{237,247,241} 0.029 & \cellcolor[RGB]{238,248,242} 0.027 & \cellcolor{white} - & \cellcolor[RGB]{240,249,243} 0.025 & \cellcolor[RGB]{248,252,249} 0.016 & \cellcolor[RGB]{224,242,231} 0.043 & \cellcolor{white} - \\
\cellcolor{sfmcolor} \owsm & \cellcolor[RGB]{239,248,242} 0.026 & \cellcolor[RGB]{227,244,234} 0.039 & \cellcolor[RGB]{229,244,235} 0.037 & \cellcolor[RGB]{223,242,230} 0.044 & \cellcolor[RGB]{223,242,230} 0.044 & \cellcolor[RGB]{211,237,221} 0.058 & \cellcolor[RGB]{243,250,246} 0.021 & \cellcolor[RGB]{250,253,251} 0.013 & \cellcolor[RGB]{243,250,246} 0.021 & \cellcolor[RGB]{179,225,196} 0.094 & \cellcolor[RGB]{236,247,240} 0.029 \\
\cellcolor{cascadecolor} \whisper\ + \aya & \cellcolor[RGB]{236,247,240} 0.030 & \cellcolor[RGB]{234,246,239} 0.032 & \cellcolor[RGB]{233,246,238} 0.032 & \cellcolor[RGB]{235,247,240} 0.030 & \cellcolor[RGB]{231,245,237} 0.035 & \cellcolor[RGB]{236,247,241} 0.029 & \cellcolor[RGB]{235,247,240} 0.030 & \cellcolor[RGB]{232,245,237} 0.034 & \cellcolor[RGB]{255,255,255} 0.008 & \cellcolor[RGB]{225,243,232} 0.042 & \cellcolor[RGB]{138,209,165} 0.141 \\
\cellcolor{cascadecolor} \nonefixed \,+ \gemma & \cellcolor[RGB]{231,245,236} 0.035 & \cellcolor[RGB]{224,243,231} 0.043 & \cellcolor[RGB]{230,245,236} 0.036 & \cellcolor[RGB]{234,246,239} 0.032 & \cellcolor[RGB]{230,245,236} 0.036 & \cellcolor[RGB]{231,245,237} 0.035 & \cellcolor[RGB]{237,248,241} 0.028 & \cellcolor[RGB]{227,244,234} 0.039 & \cellcolor[RGB]{248,252,250} 0.015 & \cellcolor[RGB]{221,241,229} 0.046 & \cellcolor[RGB]{132,206,161} 0.147 \\
\cellcolor{cascadecolor} \nonefixed \,+ \tower & \cellcolor[RGB]{240,249,243} 0.025 & \cellcolor[RGB]{232,246,238} 0.033 & \cellcolor[RGB]{238,248,242} 0.027 & \cellcolor[RGB]{234,247,239} 0.031 & \cellcolor[RGB]{227,244,234} 0.039 & \cellcolor[RGB]{233,246,238} 0.032 & \cellcolor[RGB]{236,247,240} 0.029 & \cellcolor[RGB]{232,245,237} 0.034 & \cellcolor[RGB]{252,254,253} 0.011 & \cellcolor[RGB]{226,243,232} 0.041 & \cellcolor[RGB]{120,202,152} 0.161 \\
\cellcolor{cascadecolor} \seamless\ + \aya & \cellcolor[RGB]{244,250,246} 0.020 & \cellcolor[RGB]{242,250,245} 0.022 & \cellcolor[RGB]{232,246,237} 0.034 & \cellcolor[RGB]{237,247,241} 0.028 & \cellcolor[RGB]{236,247,240} 0.029 & \cellcolor[RGB]{238,248,242} 0.027 & \cellcolor[RGB]{243,250,245} 0.021 & \cellcolor[RGB]{224,242,231} 0.043 & \cellcolor[RGB]{250,253,251} 0.013 & \cellcolor[RGB]{230,245,236} 0.036 & \cellcolor[RGB]{162,218,184} 0.113 \\
\cellcolor{cascadecolor} \nonefixed \,+ \gemma & \cellcolor[RGB]{245,251,248} 0.018 & \cellcolor[RGB]{235,247,239} 0.031 & \cellcolor[RGB]{234,246,239} 0.032 & \cellcolor[RGB]{233,246,238} 0.032 & \cellcolor[RGB]{238,248,242} 0.027 & \cellcolor[RGB]{235,247,239} 0.031 & \cellcolor[RGB]{248,252,249} 0.016 & \cellcolor[RGB]{231,245,237} 0.034 & \cellcolor[RGB]{240,249,243} 0.025 & \cellcolor[RGB]{222,242,229} 0.045 & \cellcolor[RGB]{151,214,175} 0.126 \\
\cellcolor{cascadecolor} \nonefixed \,+ \tower & \cellcolor[RGB]{244,250,247} 0.020 & \cellcolor[RGB]{237,248,241} 0.028 & \cellcolor[RGB]{233,246,238} 0.033 & \cellcolor[RGB]{236,247,240} 0.030 & \cellcolor[RGB]{232,246,237} 0.033 & \cellcolor[RGB]{236,247,240} 0.029 & \cellcolor[RGB]{247,251,249} 0.017 & \cellcolor[RGB]{232,245,237} 0.034 & \cellcolor[RGB]{246,251,248} 0.018 & \cellcolor[RGB]{231,245,237} 0.035 & \cellcolor[RGB]{141,210,167} 0.137 \\
\cellcolor{cascadecolor} \canary\ + \aya & \cellcolor[RGB]{243,250,245} 0.021 & \cellcolor[RGB]{235,247,239} 0.031 & \cellcolor[RGB]{239,248,242} 0.026 & \cellcolor[RGB]{238,248,242} 0.027 & \cellcolor[RGB]{236,247,240} 0.030 & \cellcolor[RGB]{236,247,240} 0.029 & \cellcolor[RGB]{241,249,244} 0.024 & \cellcolor[RGB]{228,244,234} 0.039 & \cellcolor[RGB]{251,253,252} 0.012 & \cellcolor[RGB]{239,248,243} 0.026 & \cellcolor{white} - \\
\cellcolor{cascadecolor} \nonefixed \,+ \gemma & \cellcolor[RGB]{237,248,241} 0.028 & \cellcolor[RGB]{229,245,235} 0.037 & \cellcolor[RGB]{233,246,238} 0.033 & \cellcolor[RGB]{234,246,239} 0.032 & \cellcolor[RGB]{233,246,238} 0.033 & \cellcolor[RGB]{232,245,237} 0.034 & \cellcolor[RGB]{247,252,249} 0.017 & \cellcolor[RGB]{227,243,233} 0.040 & \cellcolor[RGB]{244,250,246} 0.020 & \cellcolor[RGB]{228,244,234} 0.038 & \cellcolor{white} - \\
\cellcolor{cascadecolor} \nonefixed \,+ \tower & \cellcolor[RGB]{241,249,244} 0.023 & \cellcolor[RGB]{233,246,238} 0.033 & \cellcolor[RGB]{238,248,242} 0.026 & \cellcolor[RGB]{235,247,240} 0.030 & \cellcolor[RGB]{234,247,239} 0.031 & \cellcolor[RGB]{233,246,238} 0.033 & \cellcolor[RGB]{245,251,248} 0.018 & \cellcolor[RGB]{232,245,237} 0.034 & \cellcolor[RGB]{248,252,249} 0.016 & \cellcolor[RGB]{232,245,237} 0.034 & \cellcolor{white} - \\
\cellcolor{cascadecolor} \owsm\ + \aya & \cellcolor[RGB]{240,249,243} 0.025 & \cellcolor[RGB]{232,246,237} 0.034 & \cellcolor[RGB]{234,247,239} 0.031 & \cellcolor[RGB]{232,246,237} 0.034 & \cellcolor[RGB]{232,246,238} 0.033 & \cellcolor[RGB]{238,248,242} 0.027 & \cellcolor[RGB]{241,249,244} 0.023 & \cellcolor[RGB]{216,239,225} 0.052 & \cellcolor[RGB]{244,251,247} 0.020 & \cellcolor[RGB]{237,248,241} 0.028 & \cellcolor[RGB]{113,199,146} 0.170 \\
\cellcolor{cascadecolor} \nonefixed \,+ \gemma & \cellcolor[RGB]{241,249,244} 0.023 & \cellcolor[RGB]{226,243,233} 0.040 & \cellcolor[RGB]{234,246,239} 0.032 & \cellcolor[RGB]{232,245,237} 0.034 & \cellcolor[RGB]{228,244,234} 0.038 & \cellcolor[RGB]{230,245,236} 0.036 & \cellcolor[RGB]{243,250,246} 0.021 & \cellcolor[RGB]{208,236,219} 0.061 & \cellcolor[RGB]{239,248,243} 0.026 & \cellcolor[RGB]{227,244,233} 0.040 & \cellcolor[RGB]{137,208,164} 0.142 \\
\cellcolor{cascadecolor} \nonefixed \,+ \tower & \cellcolor[RGB]{238,248,242} 0.027 & \cellcolor[RGB]{231,245,236} 0.035 & \cellcolor[RGB]{234,246,239} 0.031 & \cellcolor[RGB]{230,245,236} 0.036 & \cellcolor[RGB]{230,245,236} 0.036 & \cellcolor[RGB]{234,247,239} 0.031 & \cellcolor[RGB]{241,249,244} 0.023 & \cellcolor[RGB]{228,244,234} 0.038 & \cellcolor[RGB]{248,252,249} 0.016 & \cellcolor[RGB]{236,247,240} 0.029 & \cellcolor[RGB]{121,202,152} 0.161 \\
\cellcolor{speechllmcolor}{\desta} & \cellcolor[RGB]{187,228,202} 0.085 & \cellcolor[RGB]{190,229,205} 0.081 & \cellcolor[RGB]{199,233,212} 0.071 & \cellcolor[RGB]{199,233,212} 0.071 & \cellcolor[RGB]{185,227,201} 0.088 & \cellcolor[RGB]{188,228,204} 0.084 & \cellcolor[RGB]{203,234,215} 0.066 & \cellcolor[RGB]{228,244,234} 0.039 & \cellcolor[RGB]{233,246,238} 0.033 & \cellcolor[RGB]{161,218,183} 0.115 & \cellcolor[RGB]{214,238,223} 0.054 \\
\cellcolor{speechllmcolor}{\qwenaudio} & \cellcolor[RGB]{234,246,239} 0.031 & \cellcolor[RGB]{235,247,239} 0.031 & \cellcolor[RGB]{223,242,230} 0.044 & \cellcolor[RGB]{224,243,231} 0.042 & \cellcolor[RGB]{223,242,231} 0.043 & \cellcolor[RGB]{232,246,237} 0.034 & \cellcolor[RGB]{241,249,244} 0.024 & \cellcolor[RGB]{206,235,217} 0.063 & \cellcolor[RGB]{246,251,248} 0.018 & \cellcolor[RGB]{230,245,236} 0.036 & \cellcolor[RGB]{208,236,219} 0.061 \\
\cellcolor{speechllmcolor}{\phimultimodal} & \cellcolor[RGB]{196,231,210} 0.075 & \cellcolor[RGB]{206,235,217} 0.063 & \cellcolor[RGB]{160,217,182} 0.115 & \cellcolor[RGB]{218,240,227} 0.049 & \cellcolor[RGB]{203,234,215} 0.066 & \cellcolor[RGB]{189,229,204} 0.083 & \cellcolor[RGB]{223,242,230} 0.044 & \cellcolor[RGB]{213,238,222} 0.056 & \cellcolor[RGB]{241,249,244} 0.024 & \cellcolor[RGB]{220,241,228} 0.048 & \cellcolor[RGB]{182,226,199} 0.090 \\
\cellcolor{speechllmcolor}{\voxtral} & \cellcolor[RGB]{237,248,241} 0.028 & \cellcolor[RGB]{229,244,235} 0.037 & \cellcolor[RGB]{229,245,235} 0.037 & \cellcolor[RGB]{230,245,235} 0.036 & \cellcolor[RGB]{231,245,236} 0.035 & \cellcolor[RGB]{233,246,238} 0.032 & \cellcolor[RGB]{242,249,245} 0.023 & \cellcolor[RGB]{230,245,236} 0.036 & \cellcolor[RGB]{247,252,249} 0.016 & \cellcolor[RGB]{211,237,221} 0.058 & \cellcolor[RGB]{198,232,211} 0.072 \\
\cellcolor{speechllmcolor}{\spire} & \cellcolor[RGB]{208,236,218} 0.062 & \cellcolor[RGB]{203,234,215} 0.067 & \cellcolor[RGB]{195,231,209} 0.076 & \cellcolor[RGB]{192,230,207} 0.079 & \cellcolor[RGB]{197,232,210} 0.074 & \cellcolor[RGB]{203,234,215} 0.067 & \cellcolor[RGB]{204,234,216} 0.066 & \cellcolor{white} - & \cellcolor{white} - & \cellcolor{white} - & \cellcolor{white} - \\
\cellcolor{speechllmcolor}{\qwenomni} & \cellcolor[RGB]{242,249,245} 0.023 & \cellcolor[RGB]{237,248,241} 0.028 & \cellcolor[RGB]{240,249,244} 0.024 & \cellcolor[RGB]{242,250,245} 0.022 & \cellcolor[RGB]{240,249,244} 0.024 & \cellcolor[RGB]{239,248,243} 0.026 & \cellcolor[RGB]{245,251,247} 0.019 & \cellcolor[RGB]{239,248,243} 0.025 & \cellcolor[RGB]{247,252,249} 0.016 & \cellcolor[RGB]{222,242,229} 0.045 & \cellcolor[RGB]{199,233,212} 0.071 \\
% \bottomrule
\specialrule{1pt}{0pt}{0pt}
\end{tabular}}
    \caption{St. dev. of \cometstrict\ scores for ManDi (zh-en) and CommonAccent (all other directions).
    % across source-language accent. Values correspond to those used to create Fig. \ref{fig:std_viz}.
    }
    \label{tab:stdev}
\end{table*}

\end{document}

Altough not on the final models, SLCP (*SoftLCP*) is a variant of LCP which we propose that relaxes LCP by looking at word "anchors" that stay stable between hyopthesys across time and accepting previous words to the anchor as valid prefixes even if they have changed. This helps avoid possible spikes in latency that are common of LCP due to situtatiosn where the mdoels oscillates on hypothesis. At the price of a small quality degradation, latency can substantially be improved compared to LCP, and avoids problems of other LCP variations such as RLCP (necessity of multiple hypothesis at the same timestamp, insertion of repetitions).